\documentclass[journal]{IEEEtran}
\usepackage{graphicx}
\usepackage{amsmath, amssymb}
\usepackage{caption, subcaption}
\usepackage{multirow}
\usepackage{soul,color}
\usepackage{cite}
\usepackage{url}
\usepackage{tikz}
\usetikzlibrary{shapes, arrows.meta}
\usetikzlibrary{shapes.geometric, arrows, positioning, calc}
\usepackage{pifont}
\newcommand{\cmark}{\ding{51}}
\newcommand{\xmark}{\ding{55}}
\usepackage{pgfplots}
\pgfplotsset{compat=newest}
\pgfplotsset{compat=1.18}
\usetikzlibrary{pgfplots.groupplots}
\begin{document}
\title{Hybrid State-Space and GRU-based Graph Tokenization Mamba for Hyperspectral Image Classification}
\author{Muhammad Ahmad, Muhammad Hassaan Farooq Butt, Muhammad Usama, Manuel Mazzara, Salvatore Distefano, Adil Mehmood Khan, Danfeng Hong
\thanks{M. Ahmad and S. Distefano are with the Dipartimento di Matematica e Informatica---MIFT, University of Messina, Messina 98121, Italy. (e-mail: mahmad00@gmail.com; sdistefano@unime.it).}
\thanks{Institute of Artificial Intelligence, School of Mechanical and Electrical Engineering, Shaoxing University, Shaoxing 312000, China. (e-mail: hassaanbutt67@gmail.com)}
\thanks{M.Usama is with the National University of Computer and Emerging Sciences (NUCES), CFD, Pakistan. (m.usama@nu.edu.pk)}
\thanks{M. Mazzara is with the Institute of Software Development and Engineering, Innopolis University, Innopolis, 420500, Russia. (e-mail: m.mazzara@innopolis.ru)}
\thanks{A.M. Khan is with the School of Computer Science, University of Hull, Hull HU6 7RX, UK. (e-mail: a.m.khan@hull.ac.uk)}
\thanks{D. Hong is with the Aerospace Information Research Institute, Chinese Academy of Sciences, Beijing, 100094, China, and also with the School of Electronic, Electrical and Communication Engineering, University of Chinese Academy of Sciences, 100049 Beijing, China. (e-mail: hongdf@aircas.ac.cn)}
}
\markboth{Journal of \LaTeX\ Class Files,}
{Ahmad \MakeLowercase{\textit{et al.}}}
\maketitle
\begin{abstract}
Hyperspectral image (HSI) classification plays a pivotal role in domains such as environmental monitoring, agriculture, and urban planning. However, it faces significant challenges due to the high-dimensional nature of the data and the complex spectral-spatial relationships inherent in HSI. Traditional methods, including conventional machine learning and convolutional neural networks (CNNs), often struggle to effectively capture these intricate spectral-spatial features and global contextual information. Transformer-based models, while powerful in capturing long-range dependencies, often demand substantial computational resources, posing challenges in scenarios where labeled datasets are limited, as is commonly seen in HSI applications. To overcome these challenges, this work proposes GraphMamba, a hybrid model that combines spectral-spatial token generation, graph-based token prioritization, and cross-attention mechanisms. The model introduces a novel hybridization of state-space modeling and Gated Recurrent Units (GRU), capturing both linear and nonlinear spatial-spectral dynamics. This approach enhances the ability to model complex spatial-spectral relationships while maintaining scalability and computational efficiency across diverse HSI datasets. Through comprehensive experiments, we demonstrate that GraphMamba outperforms existing state-of-the-art models, offering a scalable and robust solution for complex HSI classification tasks.
\end{abstract}
\IEEEpeerreviewmaketitle
\section{Introduction}

\IEEEPARstart{H}{yperspectral imaging (HSI)} plays a vital role in remote sensing, supporting applications such as environmental monitoring, urban planning, and agricultural management \cite{hong2024spectralgpt}. HSI classification aims to accurately classify materials or land cover types based on the rich spectral information provided by hyperspectral data \cite{8769901}. This process is crucial for crop monitoring, land use planning, and environmental change detection.

However, HSI classification faces several challenges. Firstly, HSI data typically contain hundreds of spectral bands, leading to high dimensionality. This high-dimensional space can cause issues such as the curse of dimensionality, making it difficult to extract meaningful features without losing important information \cite{10433668}. Secondly, HSIs often suffer from spectral redundancy, where neighboring pixels exhibit highly correlated spectral responses \cite{SABIN202491,yao2022semi}. This redundancy increases computational complexity and makes differentiating materials with similar spectral profiles harder. Thirdly, the scarcity of labeled data in remote sensing applications poses a significant limitation, as labeling large datasets is costly and time-consuming \cite{5238508, AHMAD2021166267}. Additionally, hyperspectral data may be affected by noise, sensor imperfections, and atmospheric interference, further complicating the classification process. These factors necessitate the development of advanced methods capable of fully utilizing both spectral and spatial information in HSI data \cite{10559660, 10475351}.

To address these challenges, various techniques have been proposed, ranging from traditional machine learning methods to advanced deep learning approaches \cite{9645266}. These methods aim to effectively capture both spectral and spatial information in HSIs, enabling more accurate and robust classification. Traditional methods like Support Vector Machines (SVMs) \cite{CHOWDHURY2024100800} and Random Forests \cite{wu2021convolutional} rely heavily on spectral information, treating each pixel independently \cite{KUMAR2024100658}. While these methods have achieved some success, they often fail to capture spatial context, resulting in noisy classifications \cite{10538297}. Moreover, the high dimensionality of HSI data exacerbates the "curse of dimensionality" \cite{6221994, RAM2024109037}, and these models require handcrafted features that may not fully capture the spectral-spatial relationships inherent in HSI data \cite{DANIEL2024100704, 10623211}.

In contrast, convolutional neural networks (CNNs) have made substantial progress in HSI classification by learning hierarchical features directly from raw data \cite{8488544, 9903062, deng2024rustqnet,8602463,li2024learning}. CNNs are particularly effective at incorporating both spectral and spatial information \cite{9307220, 8472143}. However, CNNs have limitations in capturing spectral and spatial correlations simultaneously \cite{9542960}. While 2D CNNs prioritize spatial features and neglect spectral details, 3D CNNs handle both dimensions but at the cost of increased computational complexity and reduced interpretability \cite{ahmad2024compr, WANG202336}. Additionally, CNNs' local receptive fields limit their ability to model global relationships, leading to suboptimal classification results, particularly in complex land cover scenarios \cite{MARTINS202056,wu2022uiu}.

More recently, attention mechanisms and transformer-based models have shown promising results \cite{yu2024hypersinet, 10399798, li2024s2mae,li2024casformer}. These models excel at capturing long-range dependencies and global contextual information, making them well-suited for HSI classification \cite{10604879}. However, transformers are computationally intensive due to their quadratic complexity, which can demand significant resources when working with large datasets, particularly in HSI applications where labeled samples are scarce \cite{10491347}. These models also require large amounts of labeled data to achieve optimal performance, which remains a challenge in HSI classification \cite{10440631, 10681622}. To address these limitations, graph-based transformer models have been introduced. However, these models face computational and memory challenges due to the quadratic complexity of self-attention combined with graph-based operations \cite{10638815, 10574288}. Additionally, graph-based networks can suffer from over-smoothing, where repeated information passing leads to a loss of feature distinctiveness \cite{10485641}. The reliance on graph structures also makes these models sensitive to noise and graph construction errors, requiring careful tuning \cite{10504544}.

The Vision and Spatial-Spectral Mamba models aim to reduce the computational complexity of transformers while optimizing spectral-spatial tokenization and state-space modeling for HSI classification \cite{10604894, ahmad2024multihead, WANG2024104092, SHI2024109669, yao2024spectralmamba}. The WaveMamba model incorporates wavelet transformations into the Spatial-Spectral Mamba architecture, enhancing HSI classification by capturing both fine-grained local textures and global contextual patterns \cite{10767233}. Similarly, the Spatial-Spectral Morphological Mamba model \cite{ahmad2024morphologicalmamba} integrates morphological operations with a state-space model (SSM) framework to address the computational challenges of transformers.

The MambaHSI model \cite{10604894} advances Mamba's efficient long-range modeling capabilities by incorporating spatial and spectral Mamba blocks, which adaptively model spatial-spectral interactions in hyperspectral data. The Mamba-in-Mamba (MiM) framework \cite{ZHOU2025128751} innovates HSI classification by leveraging Mamba-Cross-Scan (MCS) transformations and tokenized feature learning. Its integration of multi-scale loss and enhanced semantic tokenization achieves robust performance, especially in remote sensing applications with limited training data. HyperMamba \cite{10720896} introduces a spectral-spatial adaptive architecture through Spatial Neighborhood Adaptive Scanning (SNAS) and Spectral Adaptive Enhancement Scanning (SAES), achieving state-of-the-art performance on multiple HSI datasets. Similarly, the Local Enhanced Mamba (LE-Mamba) network \cite{WANG2024104092} employs a Local Enhanced Spatial SSM (LES-S6) and Central Region Spectral SSM (CRS-S6) for precise spatial-spectral integration, coupled with a Multi-Scale Convolutional Gated Unit (MSCGU) for robust feature aggregation and improved classification.

The HLMamba model \cite{10679212} enhances classification performance by integrating hyperspectral and LiDAR data through a Gradient Joint Algorithm (GJA) for edge contour extraction and a Multimodal Mamba Fusion Module (MMFM) to capture complex interrelationships and long-distance dependencies in multimodal datasets. The Spatial-Spectral Interaction Super-Resolution CNN-Mamba Fusion Network \cite{10695805} addresses low-resolution HSIs and high-resolution multispectral images by leveraging mutual guidance for computational efficiency. This model extracts local and global features to enhance image fusion and improve classification accuracy, solidifying Mamba's position as a versatile and powerful architecture for HSI analysis.

While Mamba models have shown promising results, they still face challenges in effectively balancing spectral and spatial information, often leading to underutilization of spectral features or overemphasis on spatial details \cite{sheng2024dualmamba}. Scalability across datasets with varying spectral band numbers is also an issue, potentially limiting their generalization capabilities. The GraphMamba model proposes to address these limitations by employing advanced spectral-spatial feature extraction and integration techniques.

\begin{figure*}[!hbt]
    \centering
    \includegraphics[width=0.99\linewidth]{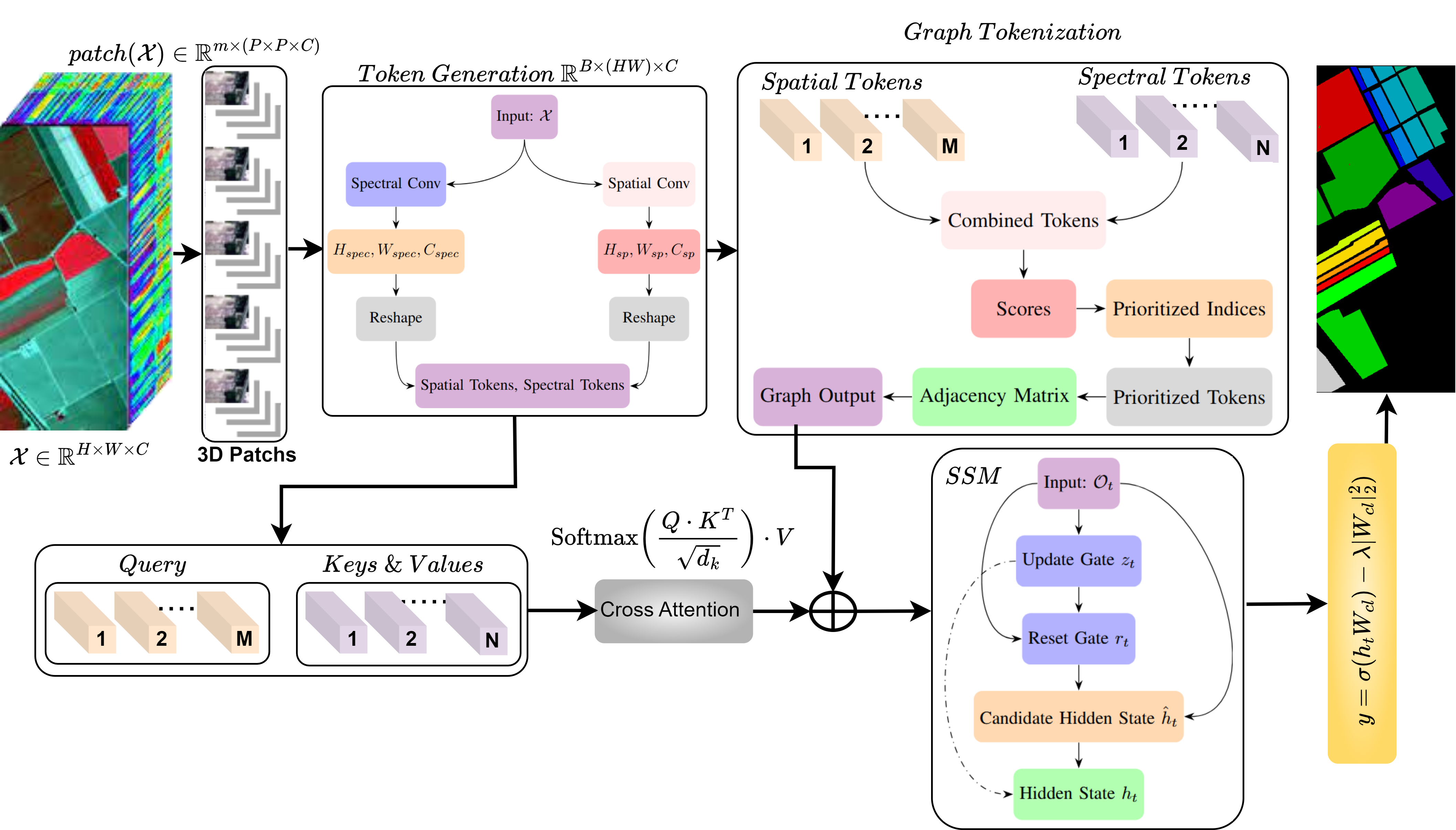}
    \caption{The proposed model processes the input HSI by slicing it into spatial and spectral patches for feature extraction. Spatial tokens, capturing spatial patterns, are generated via spatial convolutions, while spectral tokens, emphasizing spectral features, are obtained through spectral convolutions. A graph token prioritization module computes importance scores for these tokens, enabling the construction of a graph adjacency matrix that effectively captures spatial-spectral relationships. A cross-attention mechanism further refines the tokens by facilitating interactions between prioritized tokens, which are then fused with the graph output. This fused information is passed through a hybrid SSM layer, which seamlessly integrates spatial and spectral dependencies for robust classification. The arrows indicate data flow between modules, with bold arrows representing the primary flow and dashed arrows illustrating state transitions. This architecture ensures efficient feature extraction and enhanced classification performance.}
    \label{Model}
\end{figure*}

\begin{enumerate} 
    \item A dual convolutional framework is proposed to efficiently extract and tokenize both spectral and spatial features. Unlike traditional approaches, spectral tokenization utilizes 1$\times$1 convolutions to precisely isolate spectral variations, while spatial tokenization employs 3$\times$3 convolutions to effectively capture local spatial contexts. A dense projection layer follows, ensuring the generation of compact yet highly expressive token representations, optimizing both feature richness and computational efficiency.

    \item This work presents a novel graph tokenization module that dynamically prioritizes tokens through learnable scoring mechanisms. The prioritized tokens are modeled as graph nodes, with edges defined by an adjacency matrix computed from their similarities. In contrast to existing approaches that treat tokens independently, this design explicitly encodes inter-token relationships, facilitating efficient information propagation and capturing the structural dependencies intrinsic to hyperspectral data.

    \item A cross-attention layer is introduced to dynamically facilitate interactions between spectral and spatial tokens. By leveraging learnable attention weights, the module highlights critical spectral-spatial correlations, effectively enhancing discriminative feature representation while reducing redundancy.

    \item A fusion layer is designed to integrate the outputs of the graph-based modeling and attention mechanisms through a learnable projection scheme. This approach ensures the seamless combination of global structural patterns from the graph and local contextual interactions from attention, producing robust and highly discriminative feature representations for classification tasks.

    \item A hybrid SSM is proposed, incorporating a GRU to effectively capture spectral-spatial dependencies in HSI data. Unlike traditional SSMs that depend solely on linear state transitions, the GRU introduces nonlinear transitions, enabling the modeling of complex spatial-spectral dynamics. This hybrid design enhances the model's capacity to encode long-range dependencies and intricate spatial-spectral patterns while preserving the core principles of state-space modeling.
\end{enumerate}
In summary, the integration of graph-based token prioritization, cross-attention, and state-space mechanisms presents a novel solution to address the limitations of existing SSMs in HSI classification.

\section{Methodology}
\label{Meth}

An HSI cube $\mathcal{X} = \{x_i, y_i\} \in \mathbb{R}^{H \times W \times B}$ consists of spectral vectors $x_i = {x_{i,1}, x_{i,2}, x_{i,3}, \dots, x_{i,B}}$, where each $x_i$ represents the spectral information of pixel $i$, and $y_i$ denotes the corresponding class label of $x_i$. Here, $H$ and $W$ represent the spatial dimensions (height and width) of the HSI, and $B$ represents the number of spectral bands. Each pixel $i$ is associated with a spectral vector $x_i \in \mathbb{R}^{B}$. The notation $\{{x_{i,1}, x_{i,2}, \dots, x_{i,B}}\}$ represents the values of the spectral bands for pixel $i$, and $y_i$ is an integer indicating the class to which the pixel belongs.

The HSI cube is initially divided into overlapping 3D patches. Each patch is centered at the spatial coordinates $(\alpha, \beta)$ and covers $S \times S$ pixels across the spectral bands $B$. A patch is a 3D subregion of the HSI cube, encompassing both spatial and spectral information. The patch is centered at the spatial coordinates $(\alpha, \beta)$, with a spatial size of $S \times S$ and a spectral size of $B$. The total number of extracted patches, denoted by $m$, is given by $m = (H - S + 1) \times (W - S + 1)$, assuming patches are extracted with a stride of 1. If the stride $s$ is less than the patch size $S$, the patches will overlap. The overlap ratio $r$ is defined as $r = 1 - \frac{s}{S}$, indicating the proportion of overlap between adjacent patches. If $s = S$, there is no overlap ($r = 0$). If $s < S$, the patches overlap, and the overlap ratio increases as the stride decreases. The complete structure of the proposed GraphMamba model is presented in Figure \ref{Model}.

\subsection{Spatial and Spectral Token Generation}

This work proposes a dual convolutional framework for the tokenization of spectral and spatial features. This approach utilizes two types of convolutions, each optimized for a specific task. Spectral variations are captured using 1$\times$1 convolutions (shown in Equation \ref{eq1}), which focus on processing the spectral dimension of the hyperspectral data. By applying these convolutions across channels, spectral tokenization isolates important spectral features and variations, ensuring that the spectral characteristics of each pixel are preserved. This operation allows the model to focus on inter-channel dependencies and spectral correlations without mixing spatial information.

In contrast, spatial tokenization employs 3$\times$3 convolutions, which capture local spatial relationships between neighboring pixels (shown in Equation \ref{eq2}). These convolutions enable the model to identify spatial patterns and context by considering the surrounding pixels within a local receptive field. Spatial tokenization ensures that the spatial context around each pixel is accurately represented, which is crucial for understanding spatial layouts and structures in the HSI. After tokenizing the spectral and spatial features, the resulting tokens are passed through a dense projection layer, which maps them into a compact, expressive representation. This layer ensures that both spectral and spatial tokens can be effectively integrated for downstream tasks. The spatial features are denoted by $\mathcal{X}^{\text{Sp}}$, and the spectral features are denoted by $\mathcal{X}^{\text{Spc}}$, calculated as follows:

\begin{equation}
    \mathcal{X}^{\text{Sp}} = \sigma \big(\mathcal{X}_i * W_i + b_i\big) \in \mathbb{R}^{B \times N_{\text{Sp}} \times F}
    \label{eq1}
\end{equation}
\begin{equation}
    \mathcal{X}^{\text{Spc}} = \sigma \big(\mathcal{X}_i * W_i + b_i\big) \in \mathbb{R}^{B \times N_{\text{Spc}} \times F}
    \label{eq2}
\end{equation}
where $\mathcal{X}i$ represents the input patch, $W_i$ and $b_i$ are the weights and biases, respectively, $N_{\text{Sp}}$ and $N_{\text{Spc}}$ denote the number of spatial and spectral tokens, respectively, and $F$ represents the feature dimension, indicating the depth of the feature maps. The output is activated using the ReLU function $\sigma(.)$ to introduce non-linearity. The spatial and spectral tokens are then concatenated along the feature dimension to form a combined token matrix:
\begin{equation}
    \mathcal{T} = \begin{bmatrix}
    \mathcal{X}^{\text{Sp}}\\
    \mathcal{X}^{\text{Spc}}
\end{bmatrix} 
\in  \mathbb{R}^{B \times (N_{\text{Sp}} + N_{\text{Spc}}) \times F} 
\end{equation}
The concatenation operation stacks the spatial and spectral tokens along the feature dimension, forming a combined token matrix $\mathcal{T}$. This matrix contains information from both spatial and spectral features, which can be further processed in subsequent layers.

\subsection{Token Prioritization and Graph Construction}

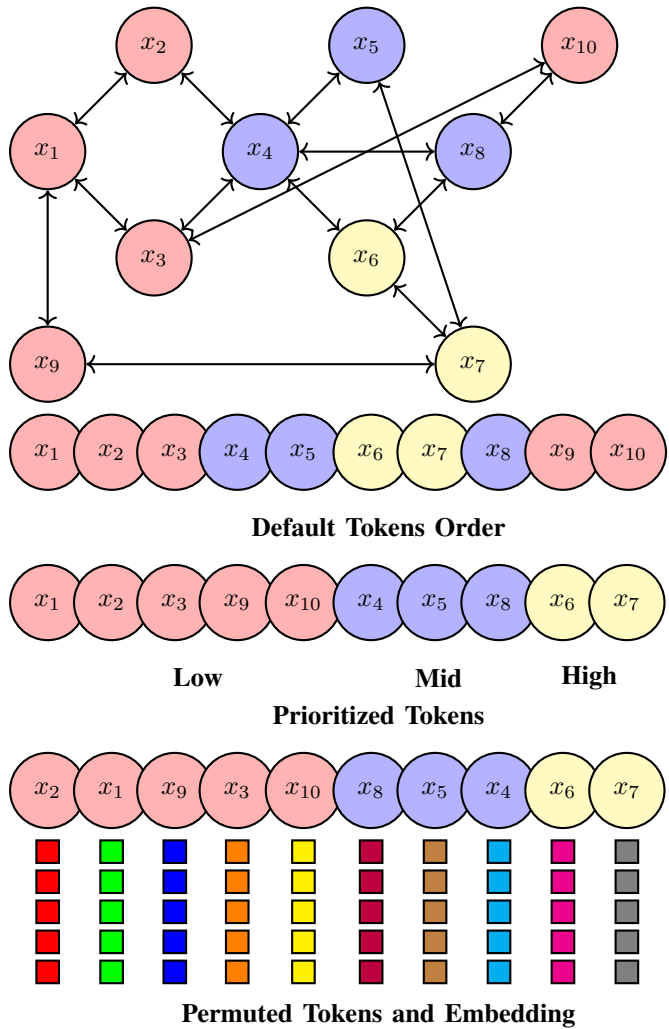
\begin{figure}[!hbt]
\centering
\begin{tikzpicture}[node distance={20mm}, thick, main/.style = {draw, circle, minimum size=10mm, inner sep=1.0pt}] 

\node[main, fill=red!30] (1) {$x_1$}; 
\node[main, fill=red!30] (2) [above right of=1] {$x_2$}; 
\node[main, fill=red!30] (3) [below right of=1] {$x_3$}; 
\node[main, fill=blue!30] (4) [above right of=3] {$x_4$}; 
\node[main, fill=blue!30] (5) [above right of=4] {$x_5$}; 
\node[main, fill=yellow!30] (6) [below right of=4] {$x_6$}; 
\node[main, fill=yellow!30] (7) [below right of=6] {$x_7$}; 
\node[main, fill=blue!30] (8) [above right of=6] {$x_8$}; 
\node[main, fill=red!30] (9) [below left of=3] {$x_9$}; 
\node[main, fill=red!30] (10) [above right of=8] {$x_{10}$};

\draw[<->, black] (1) -- (2); 
\draw[<->, black] (1) -- (3); 
\draw[<->, black] (2) -- (4); 
\draw[<->, black] (3) -- (4); 
\draw[<->, black] (5) -- (4); 
\draw[<->, black] (6) -- (4); 
\draw[<->, black] (7) -- (5); 
\draw[<->, black] (8) -- (6); 
\draw[<->, black] (9) -- (7); 
\draw[<->, black] (10) -- (8); 
\draw[<->, black] (1) -- (9); 
\draw[<->, black] (4) -- (8); 
\draw[<->, black] (6) -- (7); 
\draw[<->, black] (3) -- (10);

\begin{scope}[yshift=-4.0cm]
    \node[main, fill=red!30] (f1) at (0,0) {$x_1$};
    \node[main, fill=red!30] (f2) at (0.85cm,0) {$x_2$};
    \node[main, fill=red!30] (f3) at (1.69cm,0) {$x_3$};
    \node[main, fill=blue!30] (f4) at (2.52cm,0) {$x_4$};
    \node[main, fill=blue!30] (f5) at (3.40cm,0) {$x_5$};
    \node[main, fill=yellow!30] (f6) at (4.30cm,0) {$x_6$};
    \node[main, fill=yellow!30] (f7) at (5.15cm,0) {$x_7$};
    \node[main, fill=blue!30] (f8) at (6.00cm,0) {$x_8$};
    \node[main, fill=red!30] (f9) at (6.85cm,0) {$x_9$};
    \node[main, fill=red!30] (f10) at (7.72cm,0) {$x_{10}$};
    \node at (4.4cm, -1cm) {\textbf{Default Tokens Order}};
\end{scope}

\begin{scope}[yshift=-6.0cm]
    \node[main, fill=red!30] (d1) at (0,0) {$x_1$}; 
    \node[main, fill=red!30] (d2) at (0.85cm,0) {$x_2$}; 
    \node[main, fill=red!30] (d3) at (1.69cm,0) {$x_3$}; 
    \node[main, fill=red!30] (d9) at (2.52cm,0) {$x_9$}; 
    \node[main, fill=red!30] (d10) at (3.40cm,0) {$x_{10}$}; 
    \node[main, fill=blue!30] (d4) at (4.30cm,0) {$x_4$}; 
    \node[main, fill=blue!30] (d5) at (5.15cm,0) {$x_5$}; 
    \node[main, fill=blue!30] (d8) at (6.00cm,0) {$x_8$}; 
    \node[main, fill=yellow!30] (d6) at (6.85cm,0) {$x_6$}; 
    \node[main, fill=yellow!30] (d7) at (7.70cm,0) {$x_7$}; 
    \node at (2.0cm, -1cm) {\textbf{Low}};
    \node at (5.2cm, -1cm) {\textbf{Mid}};
    \node at (7.2cm, -1cm) {\textbf{High}};
    \node at (4.4cm, -1.5cm) {\textbf{Prioritized Tokens}};
\end{scope}

\begin{scope}[yshift=-8.5cm]
    \node[main, fill=red!30] (p2) at (0,0) {$x_2$}; 
    \node[main, fill=red!30] (p1) at (0.85cm,0) {$x_1$}; 
    \node[main, fill=red!30] (p9) at (1.69cm,0) {$x_9$}; 
    \node[main, fill=red!30] (p3) at (2.52cm,0) {$x_3$}; 
    \node[main, fill=red!30] (p10) at (3.40cm,0) {$x_{10}$}; 
    \node[main, fill=blue!30] (p8) at (4.30cm,0) {$x_8$}; 
    \node[main, fill=blue!30] (p5) at (5.15cm,0) {$x_5$}; 
    \node[main, fill=blue!30] (p4) at (6.00cm,0) {$x_4$}; 
    \node[main, fill=yellow!30] (p6) at (6.85cm,0) {$x_6$}; 
    \node[main, fill=yellow!30] (p7) at (7.70cm,0) {$x_7$}; 
\end{scope}

\begin{scope}[yshift=-10cm]
    \foreach \x/\color in {p2/red, p1/green, p9/blue, p3/orange, p10/yellow, p8/purple, p5/brown, p4/cyan, p6/magenta, p7/gray} {
        \foreach \i in {0, 0.4, 0.8, 1.2, 1.6} {
            \node[draw, fill=\color, minimum size=3mm, inner sep=0pt] at (\x.south) [shift={(0,-\i-0.3)}] {};
        }
    }
    
    \node at (4.4cm, -1.5cm) {\textbf{Permuted Tokens and Embedding}};
\end{scope}

\end{tikzpicture}
\caption{Visualization of permuted tokens with embeddings. Each node represents a token, categorized by degree order: Low (red), Mid (blue), and High (yellow). Below each node, small colored squares illustrate different embeddings, with distinct colors used to emphasize varying embedding representations. The arrangement highlights the distribution and variety of embeddings for different tokens in a visual format.}
\label{Graph}
\end{figure}

A dense layer is applied to compute the prioritization scores for each token:
\begin{equation}
    \mathcal{S} = \big(\mathcal{T} * W_{\mathcal{T}} + b_{\mathcal{T}}\big) \in  \mathbb{R}^{B \times (N_{\text{Sp}} + N_{\text{Spc}}) \times 1} 
\end{equation}
where $W_{\mathcal{T}}$ is a learned weight matrix and $b_{\mathcal{T}}$ is a bias term. This operation projects the features into a new space where the output values, $\mathcal{S}$, can take any real values. To ensure non-negative and interpretable prioritization scores, we apply a ReLU activation function:
\begin{equation}
     \mathcal{S} = max(0,\mathcal{S})
\end{equation}
The ReLU activation function clips any negative values to zero, resulting in a non-negative score. These scores are then used to indicate the relative importance of different elements, with higher values signifying greater priority in the subsequent process. This mechanism ensures that the computed prioritization scores are valid and easy to interpret. The output, $\mathcal{S}$, contains the prioritization scores for each token, reflecting their importance. The indices of the top $N$ prioritized tokens are selected based on these scores.
\begin{equation}
    \mathcal{I} = \text{TopIndices}\big(\mathcal{S},N\big) \in \mathbb{R}^{B\times N}
\end{equation}
Using the prioritized indices $\mathcal{I}$, the prioritized tokens are gathered:
\begin{equation}
    \mathcal{X}_{\text{Pro}} = \text{Gather}\big(\mathcal{T}, \mathcal{I}\big) \in \mathbb{R}^{B \times N \times F}
\end{equation}
To model the relationships between the prioritized tokens, an adjacency matrix $\mathcal{A}$ is calculated as the dot product of $\mathcal{X}_{\text{Pro}}$ with itself:
\begin{equation}
    \mathcal{A} = \mathcal{X}_{\text{Pro}} \cdot \mathcal{X}_{\text{Pro}}^T \in \mathbb{R}^{B \times N \times N}
\end{equation}
The adjacency matrix $\mathcal{A}$ represents the relationships between the prioritized tokens. The dot product operation computes the similarity between tokens, forming the edges of the graph. This matrix is then used to aggregate information from neighboring tokens by multiplying it with the prioritized tokens, which helps capture contextual information across the graph. Finally, the graph output is computed by multiplying the adjacency matrix with the prioritized tokens:
\begin{equation}
    \mathcal{Y} = \mathcal{A} \cdot \mathcal{X}_{\text{Pro}} \in \mathbb{R}^{B \times N \times F}
\end{equation}
Finally, this output is passed through another dense layer to obtain the final result:
\begin{equation}
    \mathcal{\hat{Y}} =  \sigma \big(\mathcal{Y} * W_{\mathcal{Y}} + b_{\mathcal{Y}}\big) \in  \mathbb{R}^{B \times N \times F} 
\end{equation}
where $F$ is the number of output channels specified for the layer. The details about graph token prioritization, permutations, and graph embeddings are presented in Figure \ref{Graph}.

\subsection{Cross-Attention Mechanism}
\begin{figure}[!hbt]
    \centering
    \begin{tikzpicture}[
        node distance=1.2cm,
        every node/.style={rectangle, rounded corners, text centered, minimum height=1cm},
        arrow/.style={-Stealth},
        scale=0.85, transform shape]

        \node (spatial_tokens) [fill=violet!30, minimum width=1.5cm] {Spatial Tokens};
        \node (spectral_tokens) [fill=blue!30, right=3cm of spatial_tokens, minimum width=1.5cm] {Spectral Tokens};

        \node (query) [fill=pink!30, below=0.5cm of spatial_tokens, minimum width=1.8cm] {Query $Q$};
        \node (key) [fill=red!30, below=0.5cm of spectral_tokens, minimum width=1.8cm] {Key $K$};
        \node (value) [fill=orange!30, below=2cm of key, minimum width=1.8cm] {Value $V$};

        \node (attention_weights) [fill=gray!30, below=1.0cm of {$(query)!0.5!(key)$}, minimum width=1.8cm] {$A = \text{softmax}(\frac{QK^T}{\sqrt{d_k}})$};
        \node (attention_output) [fill=violet!30, below=0.5cm of attention_weights] {Output ($A \cdot V$)};

        \draw[arrow] (spatial_tokens) -- (query);
        \draw[arrow] (spectral_tokens) -- (key);
        \draw[arrow] (spectral_tokens) to[out=360,in=0] (value);
        \draw[arrow] (query) to [out=270,in=180] (attention_weights);
        \draw[arrow] (key) to[out=270,in=0] (attention_weights);
        \draw[arrow] (attention_weights) -- (attention_output);
        \draw[arrow] (value) -- (attention_output);

    \end{tikzpicture}
    \caption{Cross Attention Process over spatial and spectral tokens.}
    \label{FigCrossAttention}
\end{figure}
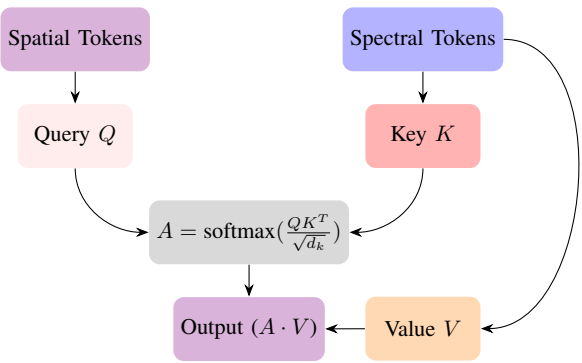

The spatial tokens $\mathcal{X}^{\text{Sp}}$ and spectral tokens $\mathcal{X}^{\text{Spc}}$ are transformed into query, key, and value representations for the cross-attention mechanism, as shown in Figure \ref{FigCrossAttention}:
\begin{equation}
    Q \in \mathbb{R}^{B \times N_{\text{Sp}} \times F} ; K \in \mathbb{R}^{B \times N_{\text{Spc}} \times F} ; V \in \mathbb{R}^{B \times N_{\text{Sp}} \times F}
\end{equation}
The attention weights are computed using the dot product of the query and key matrices, followed by a Softmax operation to normalize the scores:
\begin{equation}
    \mathcal{A}_{\text{Weights}} = \text{Softmax} \bigg(\frac{Q \cdot K^T}{\sqrt{d_k}}\bigg) \in \mathbb{R}^{B \times N_{\text{Sp}} \times N_{\text{Spc}}}
\end{equation}
where $d_k$ is the dimensionality of the key representations. The final output of the Cross-Attention layer is obtained by applying the attention weights to the value representations:
\begin{equation}
    \mathcal{A}_{\text{output}} = (A_{\text{Weights}} \cdot V) \in \mathbb{R}^{B \times N_{\text{Sp}} \times F}
\end{equation}

\subsection{GRU-based State Space Model}
The projected graph output ($\mathcal{\hat{Y}}$) and the attention output ($\mathcal{A}_{\text{output}}$) are concatenated along the second dimension after applying a dense layer to both the graph and attention outputs:
\begin{equation}
    \mathcal{O} = \begin{bmatrix}
        \mathcal{\hat{Y}} \\ 
        \mathcal{A}_{\text{output}}
\end{bmatrix} 
\in  \mathbb{R}^{B \times (N_{\text{Nodes}} + N_{\text{Sp}}) \times F} 
\end{equation}
Finally, the input $\mathcal{O}$ is processed by SSM sequences:
\begin{equation}
    z_t = \sigma(W_z \cdot x_t + U_z \cdot h_{t-1} + b_z)
\end{equation}
\begin{equation}
    r_t = \sigma(W_r \cdot x_t + U_r \cdot h_{t-1} + b_r)
\end{equation}
\begin{equation}
    \hat{h}_t  = \text{tanh} (W_h \cdot x_t + U_h \cdot (r_t \odot h_{t-1}) + b_h)
\end{equation}
\begin{equation}
    h_t = (1 - z_t) \odot h_{t-1} + z_t \odot \hat{h}_t
\end{equation}
The update gate ($z_t$) controls how much of the previous state to retain, while the reset gate ($r_t$) determines how much of the previous state to discard. The candidate hidden state ($\hat{h}t$) serves as a potential update for the hidden state, modulated by the reset gate. The final hidden state $h_t$ is a weighted combination of the previous state $h{t-1}$ and the candidate hidden state $\hat{h}_t$. The SSM processes the sequence of tokens and generates a final hidden state, representing the information learned from the sequence. Finally, a classifier is applied to the final hidden state as follows:
\begin{equation}
    y = \sigma(h \cdot W_{\text{classifier}}) - \lambda |W_{\text{classifier}}|_2^2)
\end{equation}
The final hidden state is passed through the classifier to generate the output predictions $y$. $l_2$ regularization is applied to prevent overfitting by penalizing large weights in the classifier, with the regularization strength controlled by $\lambda = 0.01$.

\section{Experimental Datasets}
\label{Data}

To evaluate the performance of GraphMamba, it was tested on several publicly available HSI datasets. The datasets used in the experiments include Pavia University (PU), Pavia Center (PC), Salinas (SA), University of Houston (UH), and WHU-Hi-HanChuan (HC). Table \ref{Tab.1} summarizes the key characteristics of each dataset used.
\begin{table}[!hbt]
    \centering
    \caption{Summary of the HSI datasets used for experimental evaluation.}
    \resizebox{\columnwidth}{!}{\begin{tabular}{cccccc} \hline 
        --- & \textbf{SA} & \textbf{UH} & \textbf{PU} & \textbf{PC} & \textbf{HC} \\  \hline 
        \textbf{Sensor} & AVIRIS & CASI & ROSIS-03 & ROSIS-03 & Headwall Nano \\
        \textbf{Wavelength} & $350-1050$ & $350-1050$ & $430-860$ & $430-860$ & $400-1000$ \\
        \textbf{Resolution} & $3.7m$ & $2.5m$ & $1.3m$ & $1.3m$ & $0.109m$ \\ 
        \textbf{Spatial} & $512 \times 217$ & $340\times 1905$ & $610 \times 610$ & $1096 \times 1096$ & $1217 \times 303$ \\
        \textbf{Spectral} & 224 & 144 & 103 & 102 & 274 \\
        \textbf{Classes} & 16 & 15 & 9 & 9 & 16 \\
        \textbf{Source} & Aerial & Aerial & Aerial & Aerial & Aerial \\ \hline 
    \end{tabular}}
    \label{Tab.1}
\end{table}

The \textbf{WHU-Hi-HanChuan (HC)} dataset was collected on June 17, 2016, from 17:57 to 18:46 in Hanchuan, Hubei Province, China, using a 17-mm focal length Headwall Nano-Hyperspec imaging sensor mounted on a Leica Aibot X6 UAV V1. The weather conditions were clear, with a temperature of approximately 30$^\circ\mathrm{C}$ and a relative humidity of about 70\%. The study area is a rural-urban fringe featuring buildings, water bodies, and cultivated land, including seven crop species: strawberry, cowpea, soybean, sorghum, water spinach, watermelon, and greens. The UAV flew at an altitude of 250m, capturing imagery at a resolution of 1217 $\times$ 303 pixels, with 274 spectral bands ranging from 400 to 1000 nm and a spatial resolution of approximately 0.109m. The images contain numerous shadowed areas because the dataset was acquired in the afternoon when the solar elevation angle was low.

The \textbf{Salinas (SA)} dataset was collected by the 224-band Airborne Visible/Infrared Imaging Spectrometer (AVIRIS) sensor over Salinas Valley, California. It features high spatial resolution with 3.7-meter pixels and covers an area of 512 lines by 217 samples. The dataset, available only as at-sensor radiance data, includes vegetation, bare soils, and vineyard fields. The ground truth for SA comprises 16 classes.

The \textbf{University of Houston (UH)} dataset, published by the IEEE Geoscience and Remote Sensing Society in 2013 as part of its Data Fusion Contest, was collected by the Compact Airborne Spectrographic Imager (CASI). This dataset has a spatial resolution of 2.5 meters per pixel (MPP), with dimensions of 340 $\times$ 1905 pixels and 144 spectral bands spanning wavelengths from 0.38 to 1.05 $\mu$m. The ground truth includes 15 different land-cover classes.

\textbf{Pavia University (PU) and Pavia Center (PC)} are two scenes acquired by the Reflective Optics System Imaging Spectrometer (ROSIS) sensor during a flight campaign over Pavia, northern Italy. The number of spectral bands is 102 for PC and 103 for PU. PC is a 1096 $\times$ 1096 pixels image, while PU is 610 × 610 pixels. The geometric resolution is 1.3 meters. Both image ground truths differentiate nine classes each. 

\section{Ablation Study}
\label{Training}
The weights of the GraphMamba model were randomly initialized and optimized over 50 epochs using the Adam optimizer with a learning rate of 0.001 and cross-entropy loss. The training was performed in mini-batches of 56 samples per epoch. The embedding dimensions of the Mamba block and the graph output nodes were set to 64, while the dimensions for cross-attention and state space were configured to 128. This configuration allowed the model to learn effectively by minimizing the loss function. The optimal settings identified from these experiments were subsequently used to evaluate the GraphMamba model in comparison to other methods. All experiments were conducted on a system with an Intel i9 processor, an RTX 4060 GPU, and 64GB of RAM, using Jupyter Notebook in a Linux environment.

\subsection{Impact of Training Samples}

To identify the optimal performance, experiments were conducted with various numbers of training samples, specifically 0.5\%, 1\%, 2\%, 5\%, 10\%, 15\%, 20\%, and 25\%. Testing these different splits is crucial to assess how the model performs with varying amounts of training data, which directly impacts its accuracy and generalization capabilities. Additionally, determining the most effective training ratio ensures that the model remains robust even when data is limited.

\begin{figure}[!hbt]
    \centering
    \begin{subfigure}{0.49\textwidth}
        \begin{tikzpicture}
            \begin{axis}[
                width=\textwidth, 
                height=0.70\textwidth, 
                xlabel={Percentage of Training Samples},
                ylabel={OA},
                grid=both,
                grid style={solid, gray!50},
                legend style={
                    at={(0.75,0.60)}, 
                    anchor=north west, 
                    font=\small,
                /tikz/every even column/.append style={column sep=0.1cm} 
            },
                ymax=100,
                xtick={1,2,3,4,5,6,7,8},
                xticklabels={$0.5\%$, $1\%$,$2\%$,$5\%$,$10\%$,$15\%$,$20\%$,$25\%$},
                mark options={solid},
                cycle list name=color list
            ]
            \addplot[color=blue, thick, mark=square*] coordinates {(1, 56.7265) (2, 72.6946) (3, 82.84) (4, 91.4703) (5, 95.5954) (6, 97.6580) (7, 98.3499) (8, 98.0705)};
            \addplot[color=red, thick, mark=triangle*] coordinates {(1, 73.1812) (2, 92.1496) (3, 92.9620) (4, 97.8445) (5, 98.9620) (6, 98.6693) (7, 99.5417) (8, 97.7276)};
            \addplot[color=green, thick, mark=*] coordinates {(1, 87.2344) (2, 89.4439) (3, 94.5944) (4, 98.5627) (5, 99.4199) (6, 99.4494) (7, 99.7118) (8, 99.6489)};
            \addplot[color=purple, thick, mark=o] coordinates {(1, 82.1768) (2, 87.3793) (3, 92.8963) (4, 94.8495) (5, 97.6903) (6, 97.2313) (7, 96.4408) (8, 96.8267)};
            \addplot[color=black, thick, mark=o] coordinates {(1, 96.8046) (2, 98.1032) (3, 98.9362) (4, 96.3861) (5, 99.4168) (6, 99.6179) (7, 99.1913) (8, 99.6058)};
            \legend{UH, PU, SA, HC, PC}
            \end{axis}
        \end{tikzpicture}
    \end{subfigure}
\caption{Overall accuracy (OA) across different datasets as a function of the percentage of training samples, along with the execution time for each run.}
\label{Trin}
\end{figure}
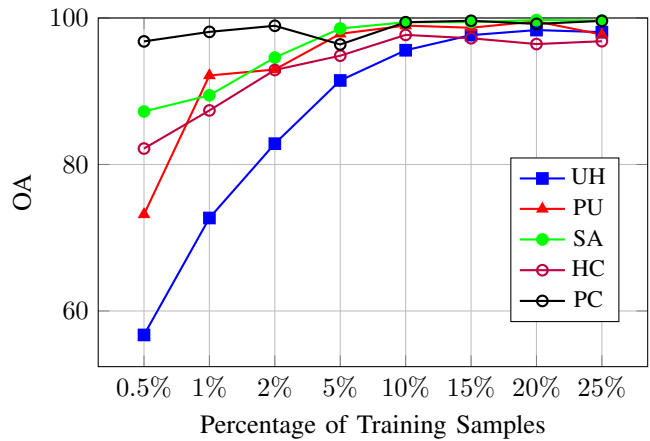

The results in Figure \ref{Trin} demonstrate the overall accuracy (OA) of the GraphMamba model as a function of the training data ratio. Accuracy consistently improves as the percentage of training data increases. Notably, the HC dataset achieved the highest accuracy of 97.69\% with 10\% of the data, highlighting the model's robustness with larger training sets. The PU, PC, and SA datasets also performed well, with peak accuracies of 99.54\%, 99.61\%, and 99.71\%, respectively. The UH dataset showed significant gains, starting at 82.17\% accuracy with 1\% of the data and reaching 97.69\% at 10\%, illustrating the model's ability to enhance performance with more training data. These findings emphasize GraphMamba's effectiveness in handling diverse datasets, demonstrating that the model's performance scales positively with increased data.

\subsection{Impact of Attention and Graph Tokenization}

This section presents the results of an ablation study designed to evaluate the individual contributions of the graph tokenization and attention mechanisms to the overall performance of GraphMamba. Table \ref{Tab2A} summarizes the model's performance under various configurations, specifically comparing scenarios where only one mechanism is used versus when both are integrated.

\textbf{Graph-only Configuration:} When the graph tokenization is used without the attention mechanism, the model performs well across all datasets, with notable improvements on datasets that require capturing spatial dependencies, such as PU and PC. For instance, the OA on the PC dataset reaches 95.04\%. However, the model's performance is still lower compared to when both mechanisms are integrated.

\textbf{Attention-only Configuration:} When only the attention mechanism is employed, the model demonstrates competitive performance, particularly on the SA and UH datasets. However, the HC dataset shows a decrease in performance, indicating that the attention mechanism alone may not fully capture the spatial correlations necessary for more complex tasks.

\begin{table*}[!hbt]
    \centering
    \caption{Performance comparison of GraphMamba with different graph and attention configurations. The results highlight the impact of integrating both graph tokenization and attention mechanisms on model performance.}
    \resizebox{\textwidth}{!}{\begin{tabular}{cc||ccc|ccc|ccc|ccc|ccc} \hline 
        \multirow{2}{*}{Graph} & \multirow{2}{*}{Attention} & \multicolumn{3}{c|}{HC} & \multicolumn{3}{c|}{UH} & \multicolumn{3}{c|}{SA} & \multicolumn{3}{c|}{PU} & \multicolumn{3}{c}{PC} \\ \cline{3-17}
        & & OA & AA & $\kappa$ & OA & AA & $\kappa$ & OA & AA & $\kappa$ & OA & AA & $\kappa$ & OA & AA & $\kappa$ \\ \hline 
        \cmark & \xmark & 87.3350 & 72.4539 & 85.1157 & 84.8436 & 84.4133 & 83.6200 & 94.1954 & 94.2040 & 93.5273 & 88.3345 & 80.3201 & 84.4336 & 95.0415 & 85.5914 & 92.9624 \\ \hline
        \xmark & \cmark & 94.6825 & 89.8931 & 93.7803 & 94.2834 & 93.3364 & 94.1441 & 99.5566 & 99.5117 & 99.5063 & 99.5511 & 99.1814 & 99.4051 & 98.9065 & 97.4988 & 98.4550 \\ \hline
        \cmark & \cmark & 97.3651 & 94.2446 & 96.9133 & 97.4979 & 97.1501 & 97.2951 & 99.8891 & 99.8803 & 99.8765 & 98.7422 & 97.5725 & 98.3333 & 99.4924 & 98.5007 & 99.2813 \\ \hline
    \end{tabular}}
    \label{Tab2A}
\end{table*}
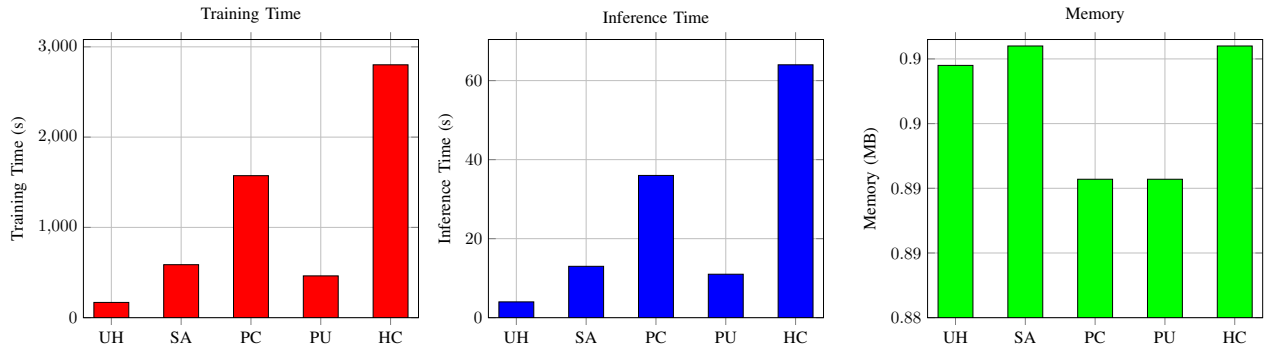
\begin{figure*}[!hbt]
    \centering
    \begin{subfigure}{0.3\textwidth} 
        \centering
        \begin{tikzpicture}[scale=0.65] 
            \begin{axis}[
                ylabel={Training Time (s)},
                ybar,
                bar width=0.5, 
                xtick={1, 2, 3, 4, 5},
                xticklabels={UH, SA, PC, PU, HC},
                ymin=0,
                grid=major,
                title={Training Time},
            ]
            \addplot[fill=red] coordinates {(1, 169) (2, 587) (3, 1573) (4, 463) (5, 2801)};
            \end{axis}
        \end{tikzpicture}
    \end{subfigure}
    \begin{subfigure}{0.3\textwidth}
        \centering
        \begin{tikzpicture}[scale=0.65] 
            \begin{axis}[
                ylabel={Inference Time (s)},
                ybar,
                bar width=0.5, 
                xtick={1, 2, 3, 4, 5},
                xticklabels={UH, SA, PC, PU, HC},
                ymin=0,
                grid=major,
                title={Inference Time},
            ]
            \addplot[fill=blue] coordinates {(1, 4) (2, 13) (3, 36) (4, 11) (5, 64)};
            \end{axis}
        \end{tikzpicture}
    \end{subfigure}
    \begin{subfigure}{0.32\textwidth} 
        \centering
        \begin{tikzpicture}[scale=0.65] 
            \begin{axis}[
                ylabel={Memory (MB)},
                ybar,
                bar width=0.5, 
                xtick={1, 2, 3, 4, 5},
                xticklabels={UH, SA, PC, PU, HC},
                ymin=0.8800,
                ymax=0.9015,
                grid=major,
                title={Memory},
            ]
            \addplot[fill=green] coordinates {(1, 0.8995) (2, 0.9010) (3, 0.8907) (4, 0.8907) (5, 0.9010)};
            \end{axis}
        \end{tikzpicture}
    \end{subfigure}
    \caption{Comparison of training time, inference time, and memory usage across different datasets. The dataset sizes are as follows: $512 \times 217 \times 224$ for SA, $340 \times 1905 \times 144$ for UH, $610 \times 610 \times 103$ for PU, $1096 \times 1096 \times 102$ for PC, and $1217 \times 303 \times 274$ for HC. The HC dataset exhibits significantly higher training and inference times, indicating that its larger size requires more time for both processes.}
    \label{TrainInf}
\end{figure*}

\textbf{Combined Graph and Attention Configuration:} The best performance is achieved when both the graph tokenization and attention mechanisms are integrated. This configuration consistently outperforms the individual components across all metrics. For instance, the PC dataset reaches an OA of 98.50\%, while PU achieves an OA of 99.89\%. This highlights that the synergy between both mechanisms enhances performance by effectively capturing both global and local feature relationships.

The results in Table \ref{Tab2A} clearly show that while each mechanism contributes to the model's performance, their combination delivers superior results. The attention mechanism improves the model’s focus on relevant features, while the graph tokenization captures essential spatial dependencies for HSI classification. Together, these mechanisms significantly boost the model’s ability to generalize across diverse datasets, positioning GraphMamba as a robust and effective method for HSI classification.

\subsection{Training Time, Inference Time, and Memory Requirements}

In HSI classification tasks, training and inference time are key metrics for assessing the model's efficiency. Training time refers to the duration required for the model to learn from the training data, while inference time indicates how quickly the trained model can generate predictions on new data. This section presents both training and inference times to offer insights into the computational demands of each dataset. These metrics are particularly important in ablation studies, where the objective is to evaluate the impact of different model components. By comparing these times across datasets of varying sizes, we gain a deeper understanding of how model complexity, dataset size, and computational resources interact.

The results shown in Figure \ref{TrainInf} illustrate the relative training and inference times for each dataset. For example, the SA dataset (512 $\times$ 217 $\times$ 224) exhibits a moderate training time of 587 seconds and an inference time of 13 seconds, indicating that the model requires a moderate amount of time to learn from this dataset. In contrast, the HC dataset (1217 $\times$ 303 $\times$ 274) shows significantly higher training and inference times, at 2801 seconds and 64 seconds, respectively. This suggests that the larger size of the HC dataset demands more time for both training and inference. The PU dataset (610 $\times$ 610 $\times$ 103) demonstrates relatively lower training and inference times, implying that smaller datasets may lead to faster computations, but possibly at the cost of lower model performance due to reduced data complexity. On the other hand, the PC dataset (1096 $\times$ 1096 $\times$ 102) presents an inverse trend, with a higher training time of 1573 seconds but a relatively moderate inference time of 36 seconds. In summary, the training and inference times not only reflect the dataset sizes but also provide valuable insights into the model's scalability and efficiency.

\begin{figure*}[!hbt]
    \centering
	\begin{subfigure}{0.19\textwidth}
		\includegraphics[width=0.99\textwidth]{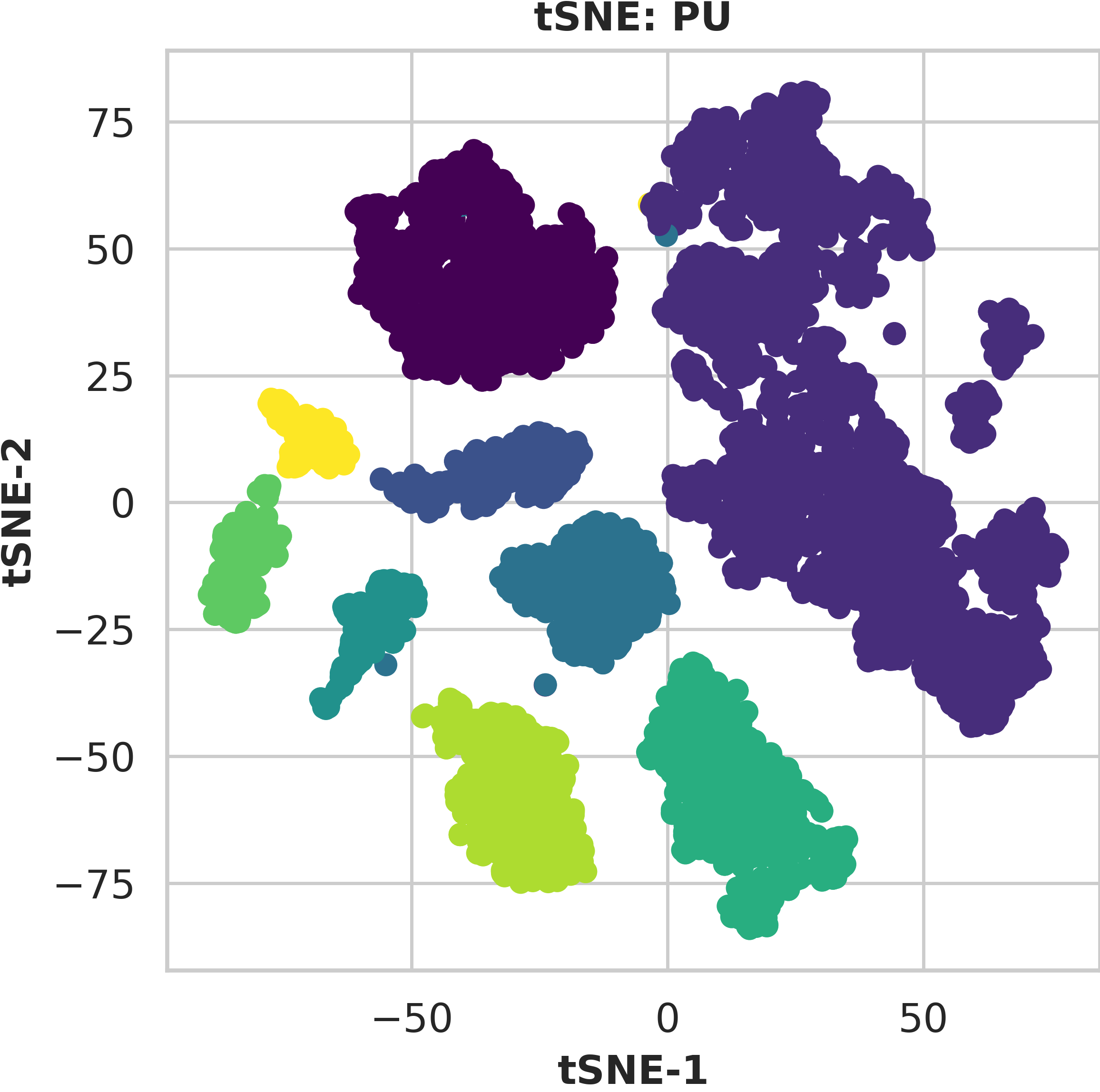}
		\caption{PU} 
		\label{Fig2A}
	\end{subfigure}
    \begin{subfigure}{0.19\textwidth}
		\includegraphics[width=0.99\textwidth]{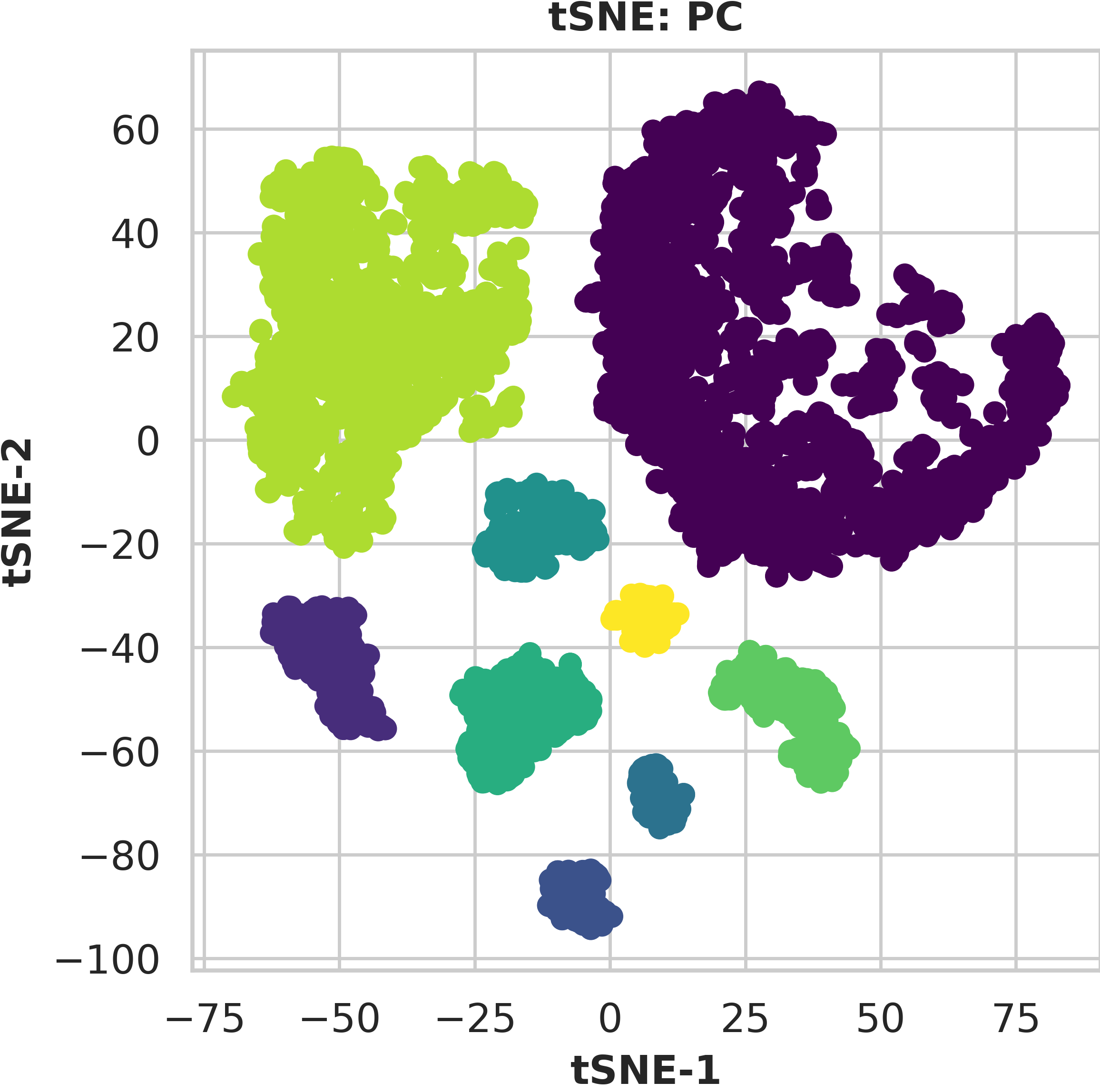}
		\caption{PC}
		\label{Fig2B}
	\end{subfigure}
	\begin{subfigure}{0.19\textwidth}
		\includegraphics[width=0.99\textwidth]{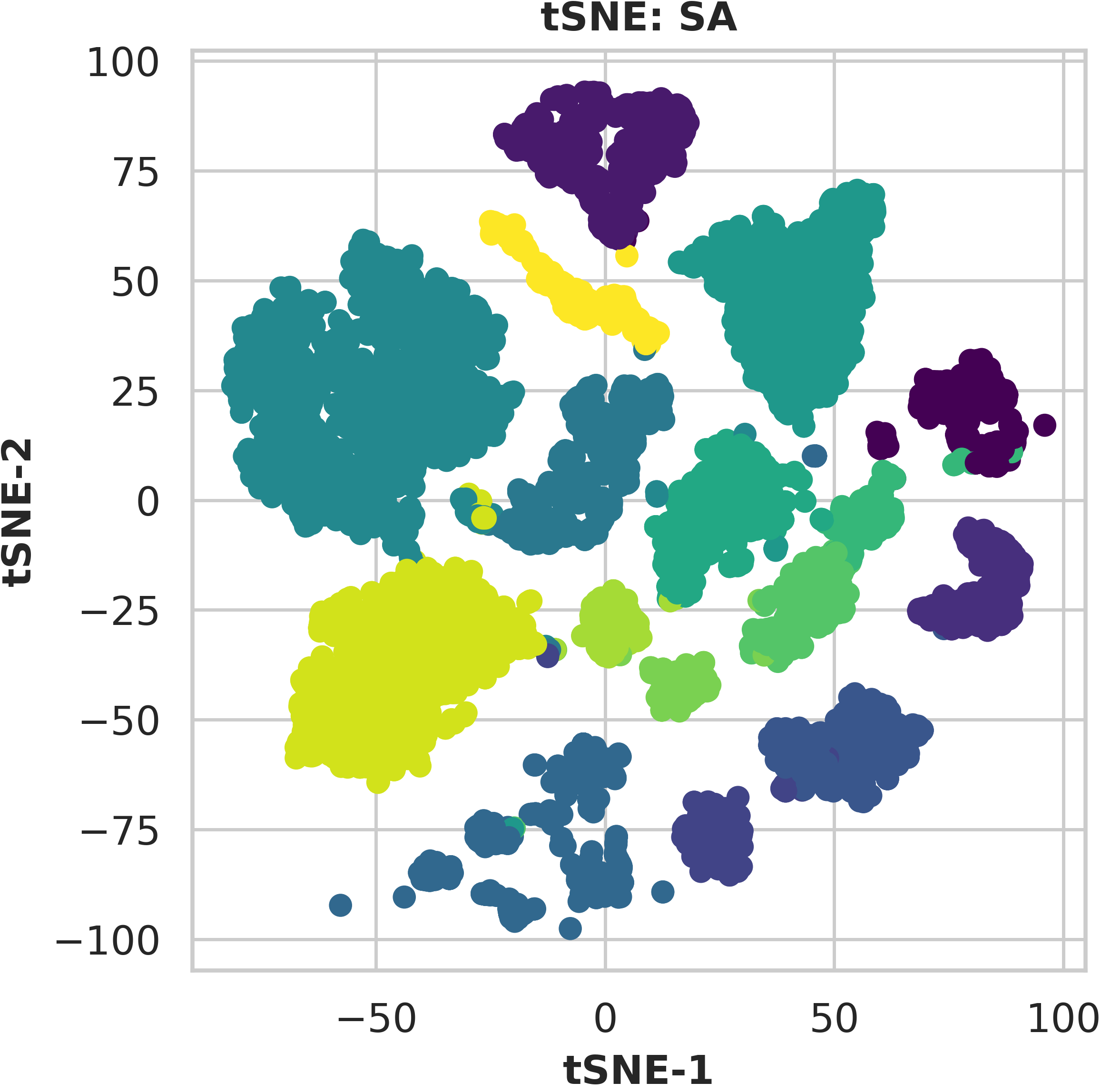}
		\caption{SA}
		\label{Fig2C}
	\end{subfigure} 
	\begin{subfigure}{0.19\textwidth}
		\includegraphics[width=0.99\textwidth]{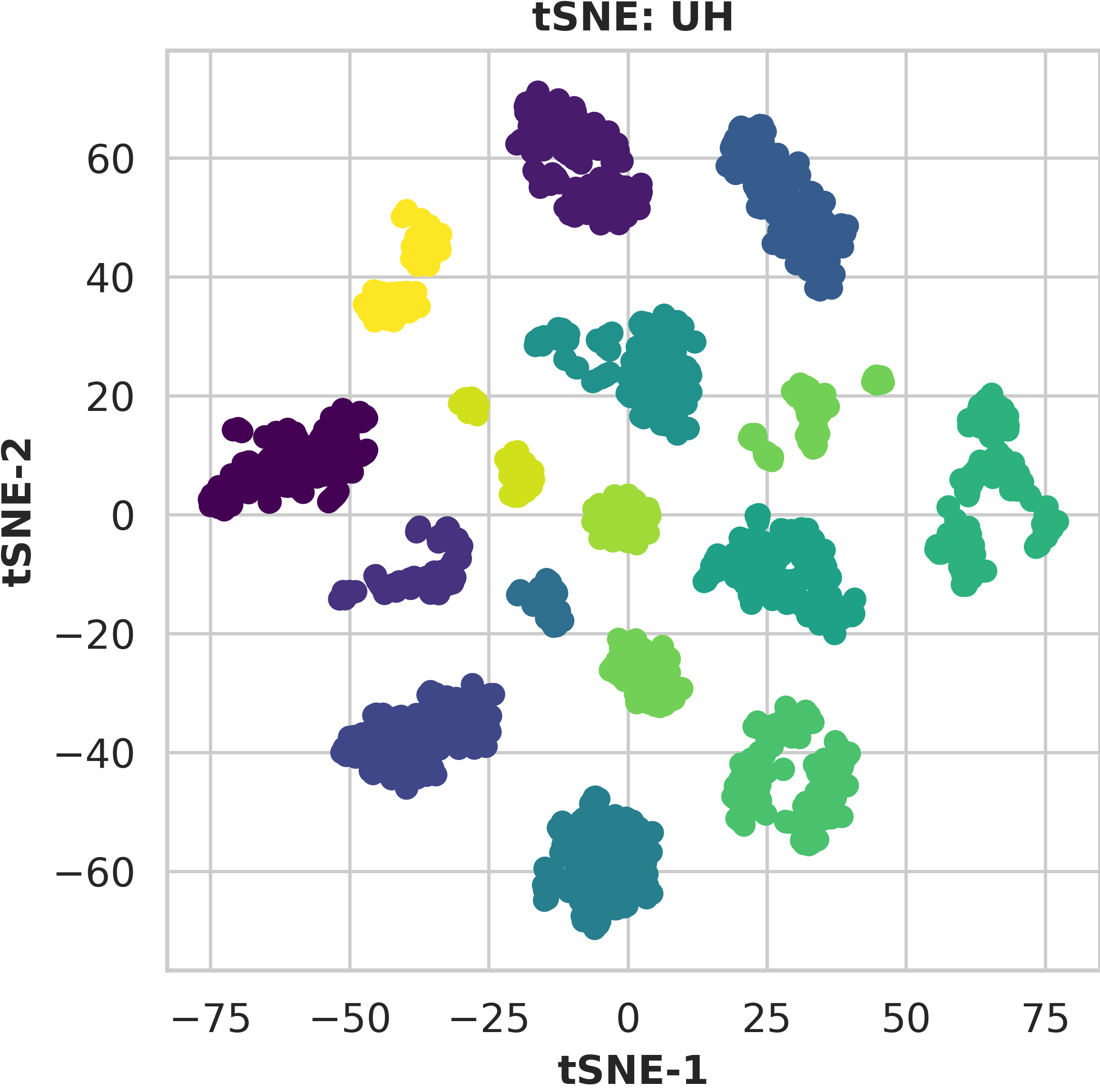}
		\caption{UH}
		\label{Fig2D}
	\end{subfigure}
    \begin{subfigure}{0.19\textwidth}
		\includegraphics[width=0.99\textwidth]{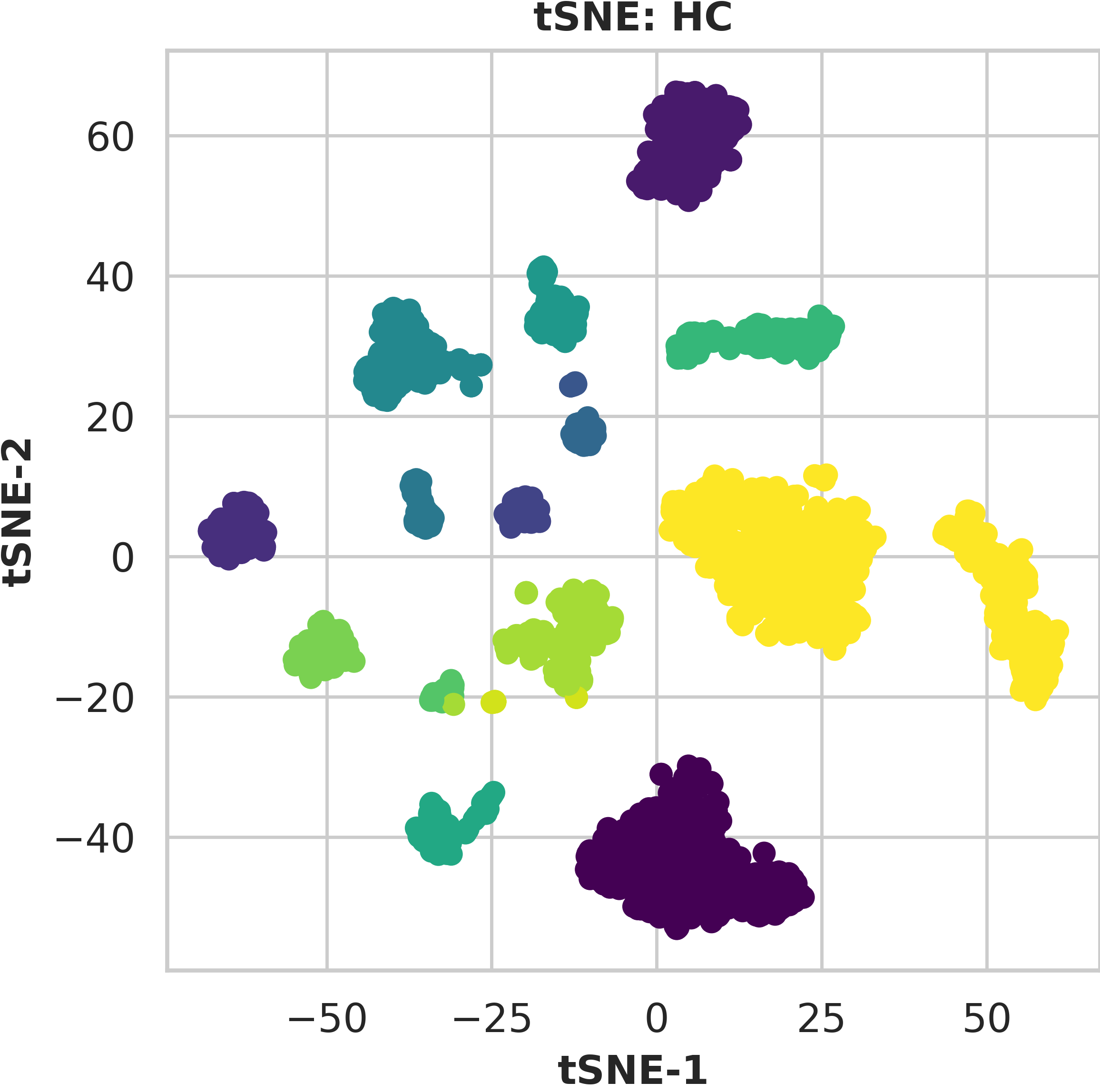}
		\caption{HC (0.5\% samples)}
		\label{Fig2E}
	\end{subfigure}
\caption{t-SNE visualization of the learned feature representations for the PU, PC, SA, UH, and HC datasets, showcasing the GraphMamba model's ability to effectively distinguish and cluster features for classification in a lower-dimensional space.}
\label{Fig2}
\end{figure*}

\begin{table*}[!hbt]
    \centering
    \caption{The table presents OA, AA, $\kappa$ coefficient, and the number of training parameters for various CNN, Transformer models, and GraphMamba.}
    \resizebox{\textwidth}{!}{\begin{tabular}{c|ccccccc|ccc|c} \hline 
       \multirow{2}{*}{\textbf{Measures}} & \textbf{2DCNN} & \textbf{3DCNN} & \textbf{HybCNN} & \textbf{2DIN} & \textbf{3DIN} & \textbf{HyIN} & \textbf{GraphCNN} & \textbf{Hybrid-ViT} & \textbf{Hir-ViT} & \textbf{SSViT} & \multirow{2}{*}{\textbf{GraphMamba}} \\ \cline{2-11}

        \multicolumn{11}{c}{\textbf{UH dataset}} \\ \hline 
        \textbf{Parameters} & 1109055 & 15839231 & 1380351 & 10658493 & 189006375 & 4495511 & 510626 & 836559 & 16680533 & 836559 & \textbf{144272} \\ \hline
        \textbf{Kappa} & 94.86 & 96.81 & 96.98 & 96.49 & 94.84 & 95.65 & 87.85 & 96.23 & 91.24 & 96.13 & 97.29 \\ \hline
        \textbf{OA} & 95.25 & 97.05 & 97.21 & 96.75 & 95.23 & 95.98 & 88.77 & 96.51 & 91.89 & 96.42 & 97.49 \\ \hline
        \textbf{AA} & 95.24 & 97.10 & 97.41 & 96.82 & 95.35 & 95.87 & 87.20 & 95.93 & 92.17 & 95.27 & 97.15 \\ \hline 
        
        \multicolumn{11}{c}{\textbf{PU dataset}} \\ \hline 
        \textbf{Parameters}  & 1108281 & 16679759 & 1379577 & 10657587 & 189005985 & 4495121 & 510500 & 835017 & 16679759 & 835017 & \textbf{143498} \\ \hline 
        \textbf{Kappa}  & 98.06 & 98.06 & 98.91 & 98.03 & 96.50 & 98.74 & 97.55 & 96.05 & 97.33 & 96.07 & 98.33 \\ \hline 
        \textbf{OA}  & 98.54 & 98.54 & 99.18 & 98.51 & 97.36 & 99.05 & 98.15 & 97.02 & 97.99 & 97.03 & 98.74 \\ \hline 
        \textbf{AA}  & 98.10 & 97.73 & 98.72 & 97.18 & 96.06 & 98.43 & 96.89 & 95.48 & 96.92 & 95.18 & 97.57 \\ \hline \hline

        \multicolumn{11}{c}{\textbf{PC dataset}} \\ \hline 
        \textbf{Parameters}  & 1108281 & 15838457 & 1379577 & 10657587 & 189005985 & 4495121 & 510380 & 835017 & 16679759 & 835017 & \textbf{143498} \\ \hline 
        \textbf{Kappa}  & 99.23 & 99.83 & 99.58 & 99.56 & 99.24 & 99.74 & 99.59 & 99.53 & 97.50 & 99.44 & 99.28 \\ \hline 
        \textbf{OA}  & 99.46 & 99.88 & 99.70 & 99.69 & 99.47 & 99.82 & 99.71 & 99.66 & 98.23 & 99.60 & 99.49 \\ \hline 
        \textbf{AA}  & 97.70 & 99.61 & 98.96 & 98.97 & 97.86 & 99.40 & 98.86 & 99.05 & 94.42 & 98.56 & 98.50 \\ \hline \hline

        \multicolumn{11}{c}{\textbf{HC dataset}} \\ \hline 
        \textbf{Parameters}  & 1109184 & 15839360 & 1380480 & 10658644 & 189006440 & 4495576 & 510647 & 836816 & 16680662 & 836816 & \textbf{144401} \\ \hline 
        \textbf{Kappa}  & 94.14 & 96.59 & 95.78 & 96.11 & 95.06 & 96.71 & 93.77 & 93.36 & 97.03 & 93.83 & 96.91 \\ \hline 
        \textbf{OA}  & 95.00 & 97.08 & 96.39 & 96.67 & 95.78 & 97.19 & 94.68 & 94.33 & 97.46 & 94.72 & 97.36 \\ \hline 
        \textbf{AA}  & 89.48 & 93.60 & 92.96 & 93.44 & 90.98 & 94.03 & 90.17 & 88.62 & 95.38 & 89.62 & 94.24 \\ \hline \hline
        
        \multicolumn{11}{c}{\textbf{SA dataset}} \\ \hline 
        \textbf{Parameters}  & 1109184 & 15839360 & 1380480 & 10658644 & 189006440 & 4495576 & 510647 & 836816 & 16680662 & 836816 & \textbf{144401} \\ \hline
        \textbf{Kappa} & 97.97 & 98.26 & 98.87 & 97.55 & 97.36 & 97.18 & 98.50 & 97.29 & 97.48 &  97.43 & 99.87 \\ \hline 
        \textbf{OA} & 98.17 & 98.43 & 98.99 & 97.80 & 97.63 & 97.47 & 98.66 & 97.56 & 97.74 & 97.69 & 99.88 \\ \hline 
        \textbf{AA} & 98.96 & 99.00 & 99.51 & 98.85 & 98.65 & 93.28 & 99.15 & 98.81 & 98.87 & 98.87 & 99.88 \\ \hline \hline 
    \end{tabular}}
    \label{comperative}
\end{table*}

To ensure a comprehensive understanding of the memory requirements during model training, we employed an enhanced method for estimating memory consumption. This approach considers various components of memory usage, including parameters, activations, gradients, and optimizer states. The memory for parameters is calculated by multiplying the total number of parameters by 4 bytes (assuming float32 precision). Activation memory is estimated by iterating through each layer of the model and computing the size of the output tensors based on the layer's output shape and batch size. Gradient memory, which holds the computed gradients during backpropagation, is approximated as equivalent to activation memory. Additionally, the memory required for optimizer states, such as momentum and velocity in Adam, is estimated to be twice the memory for parameters. These components are then summed to provide the total memory required during training. While this method does not capture certain nuanced aspects of memory usage, such as layer-specific optimization or memory sharing in distributed environments, it provides a practical and reliable estimate of the memory footprint.

\subsection{Feature Representation}

t-SNE is a dimensionality reduction technique used to visualize high-dimensional data by projecting it into a lower-dimensional space. It works by preserving the pairwise distances between data points, ensuring that similar points in the high-dimensional space remain close in the reduced space. This is achieved through a cost function based on Kullback-Leibler divergence, which measures the difference between probability distributions in both spaces. The algorithm iteratively adjusts the positions of points in the lower-dimensional space to minimize this divergence, ensuring that meaningful spatial relationships are retained.

In this study, t-SNE is employed to assess the feature representations learned by the GraphMamba model. Figures \ref{Fig2A}, \ref{Fig2B}, \ref{Fig2C}, \ref{Fig2D}, and \ref{Fig2E} show the t-SNE visualizations for the PU, PC, SA, UH, and HC datasets, respectively. These visualizations demonstrate the model's ability to distinguish between different classes, validating its effectiveness in feature learning across various datasets. The clear clustering observed in these visualizations indicates that the model successfully captures the inherent structure in the data, enhancing classification accuracy.

\begin{figure*}[!htb]
    \centering
        \begin{subfigure}{0.085\textwidth}
            \includegraphics[width=0.99\textwidth]{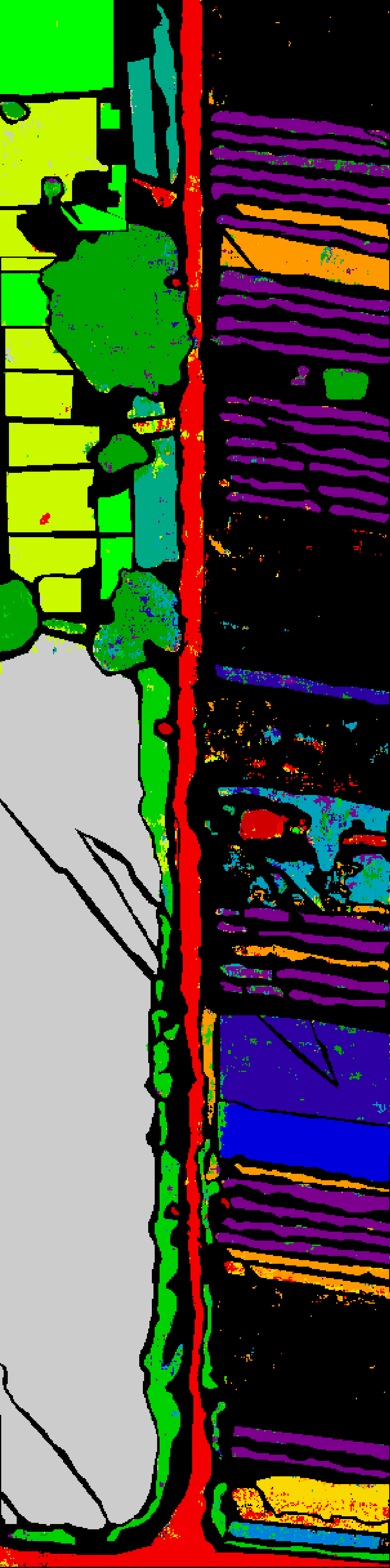}
            \caption{2DCNN}
        \end{subfigure}
        \begin{subfigure}{0.085\textwidth}
            \centering
            \includegraphics[width=0.99\textwidth]{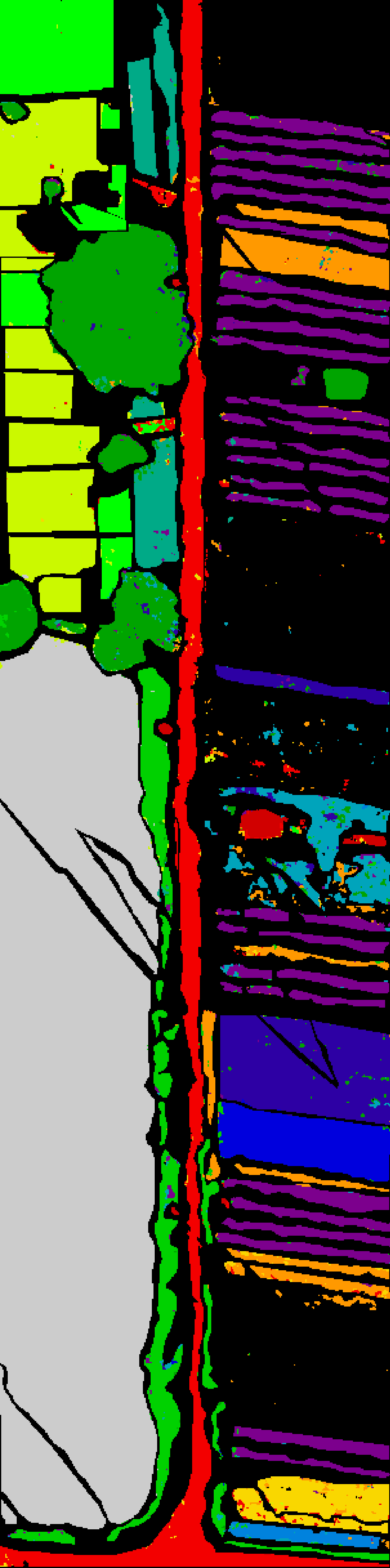}
            \caption{3DCNN}
        \end{subfigure}
        \begin{subfigure}{0.085\textwidth}
            \centering
            \includegraphics[width=0.99\textwidth]{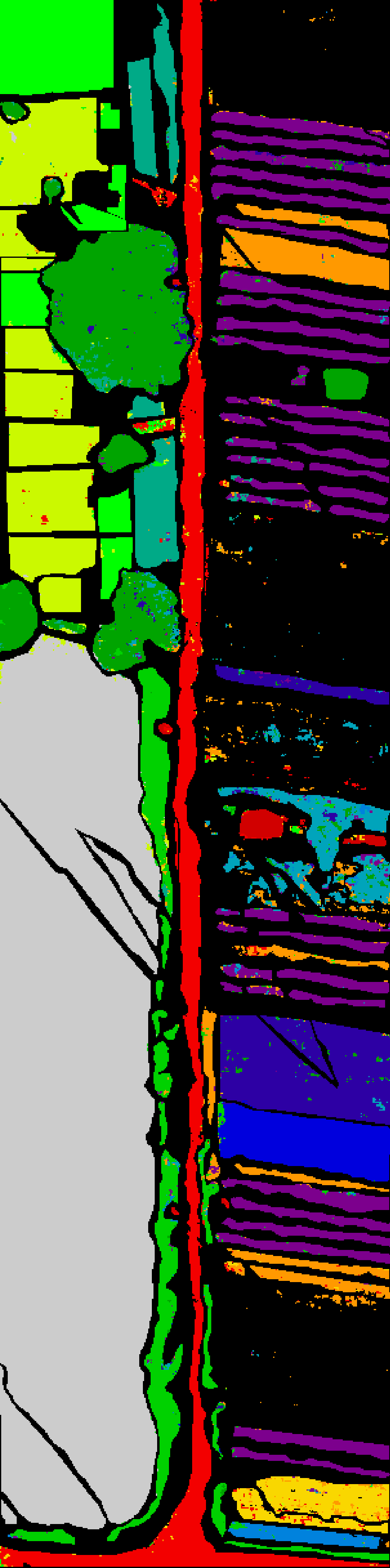}
            \caption{HCNN}
        \end{subfigure}
        \begin{subfigure}{0.085\textwidth}
            \centering
            \includegraphics[width=0.99\textwidth]{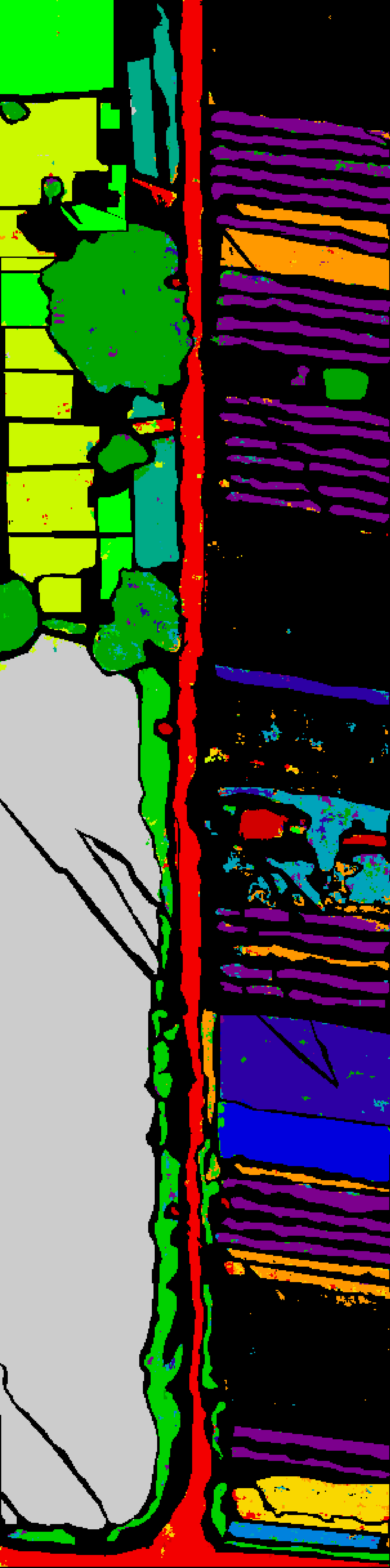}
            \caption{2DIN}
        \end{subfigure}
        \begin{subfigure}{0.085\textwidth}
            \centering
            \includegraphics[width=0.99\textwidth]{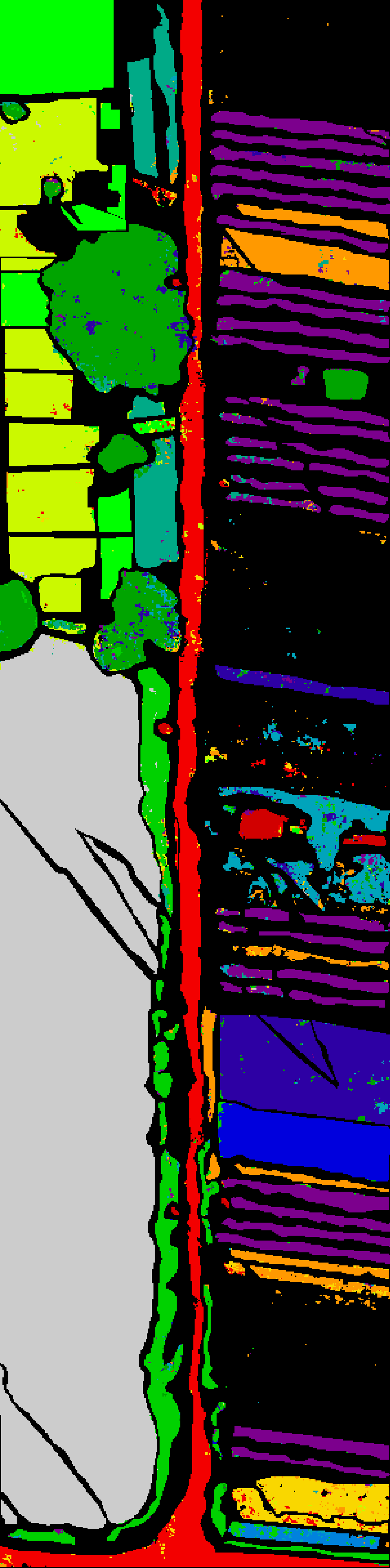}
            \caption{3DIN}
        \end{subfigure}
        \begin{subfigure}{0.085\textwidth}
            \centering
            \includegraphics[width=0.99\textwidth]{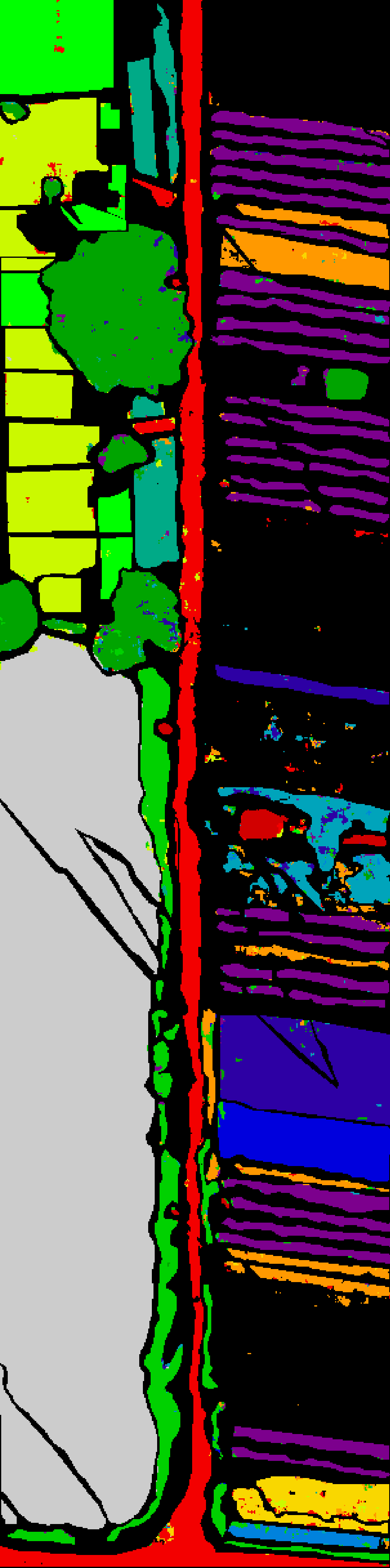}
            \caption{HIN}
        \end{subfigure}
        \begin{subfigure}{0.085\textwidth}
            \centering
            \includegraphics[width=0.99\textwidth]{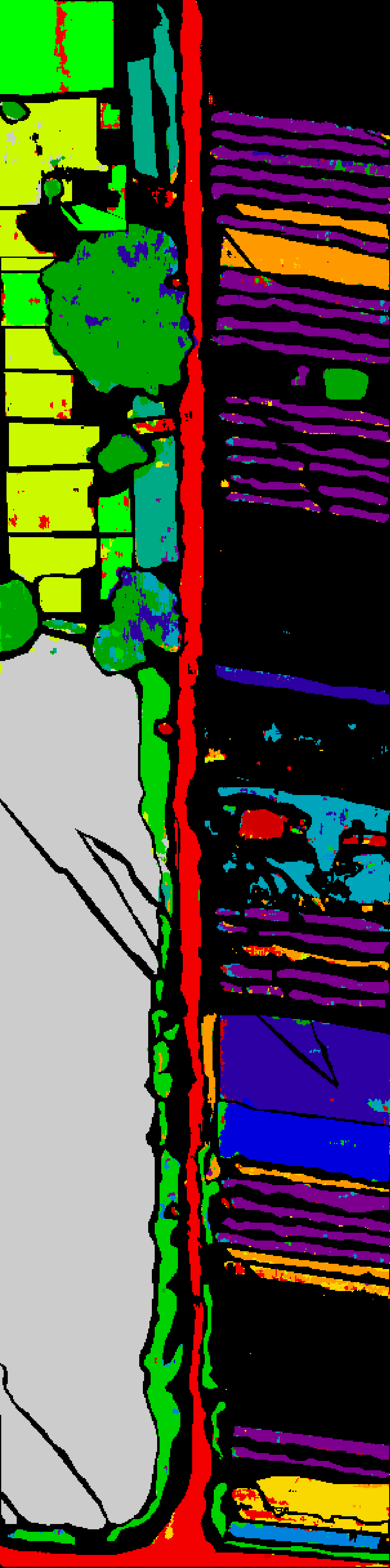}
            \caption{GCNN}
        \end{subfigure}
        \begin{subfigure}{0.085\textwidth}
            \centering
            \includegraphics[width=0.99\textwidth]{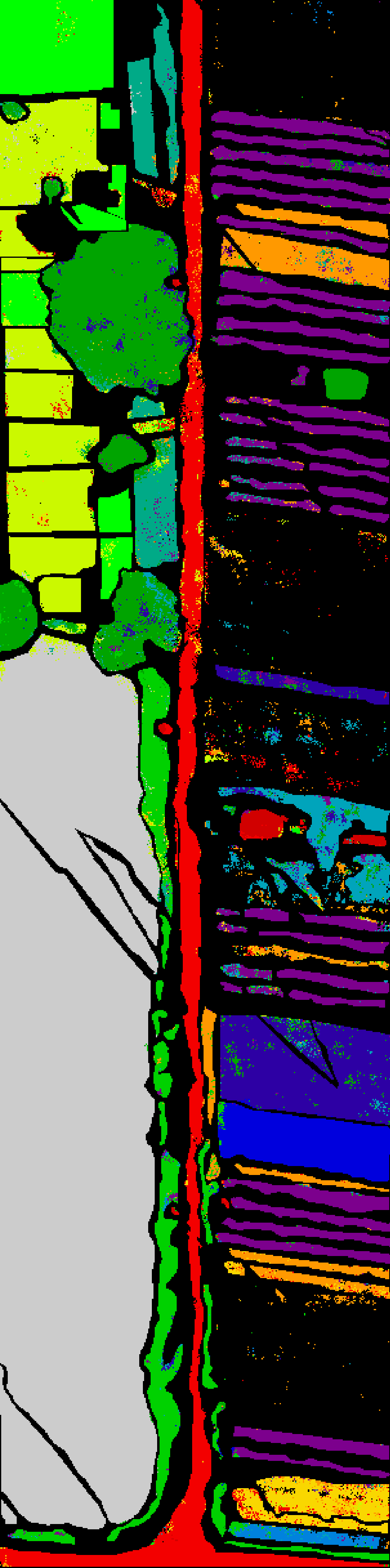}
            \caption{HViT}
        \end{subfigure}
        \begin{subfigure}{0.085\textwidth}
            \centering
            \includegraphics[width=0.99\textwidth]{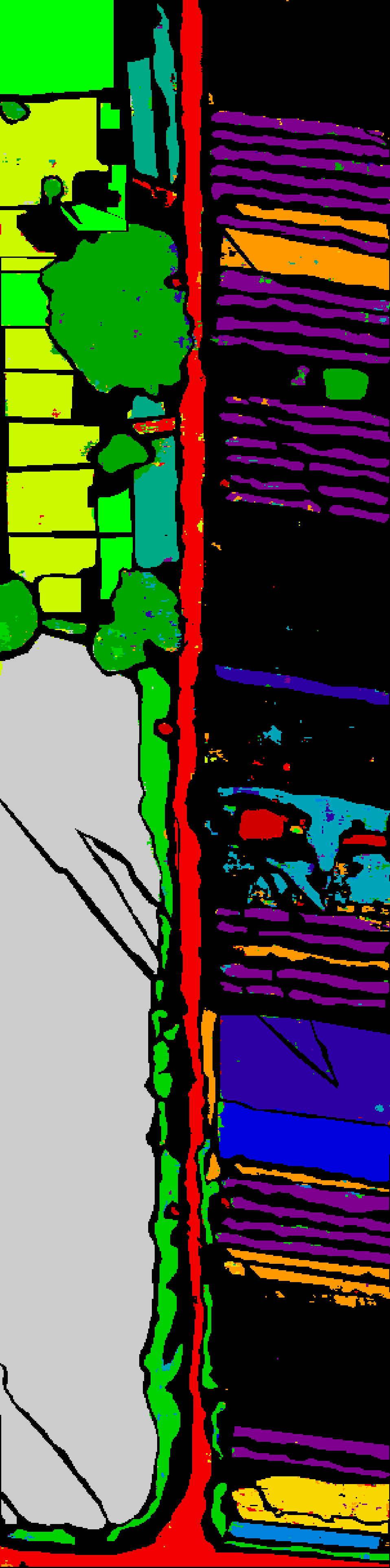}
            \caption{Hir\_ViT}
        \end{subfigure}
        \begin{subfigure}{0.085\textwidth}
            \centering
            \includegraphics[width=0.99\textwidth]{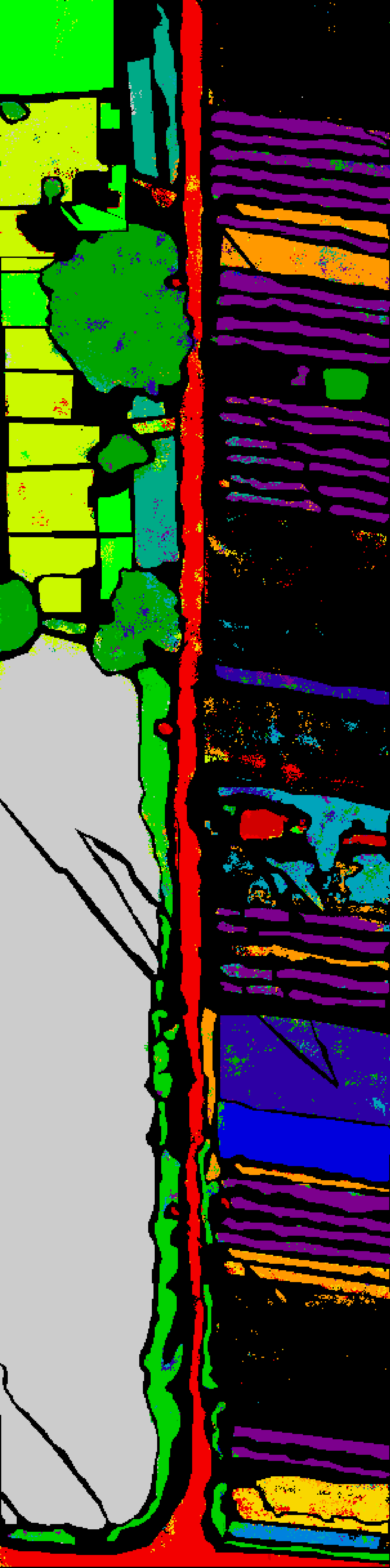}
            \caption{SST}
        \end{subfigure}
        \begin{subfigure}{0.085\textwidth}
            \centering
            \includegraphics[width=0.99\textwidth]{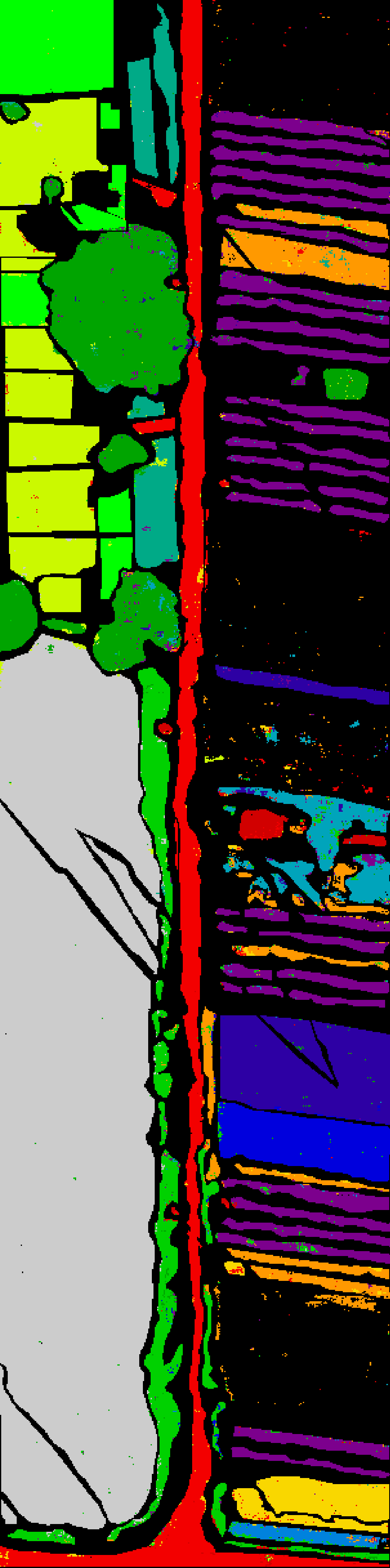}
            \caption{GMamba}
        \end{subfigure}
    \caption{The predicted ground truth maps for the HC dataset are presented for various state-of-the-art methods alongside GraphMamba. While most competing methods achieve similar accuracy, their parameter efficiency is limited and too high for deployment on resource-constrained devices, in contrast to GraphMamba.}
    \label{fig:HC_results}
\end{figure*}
\begin{figure*}[!htb]
    \centering
        \begin{subfigure}{0.085\textwidth}
            \includegraphics[width=0.99\textwidth]{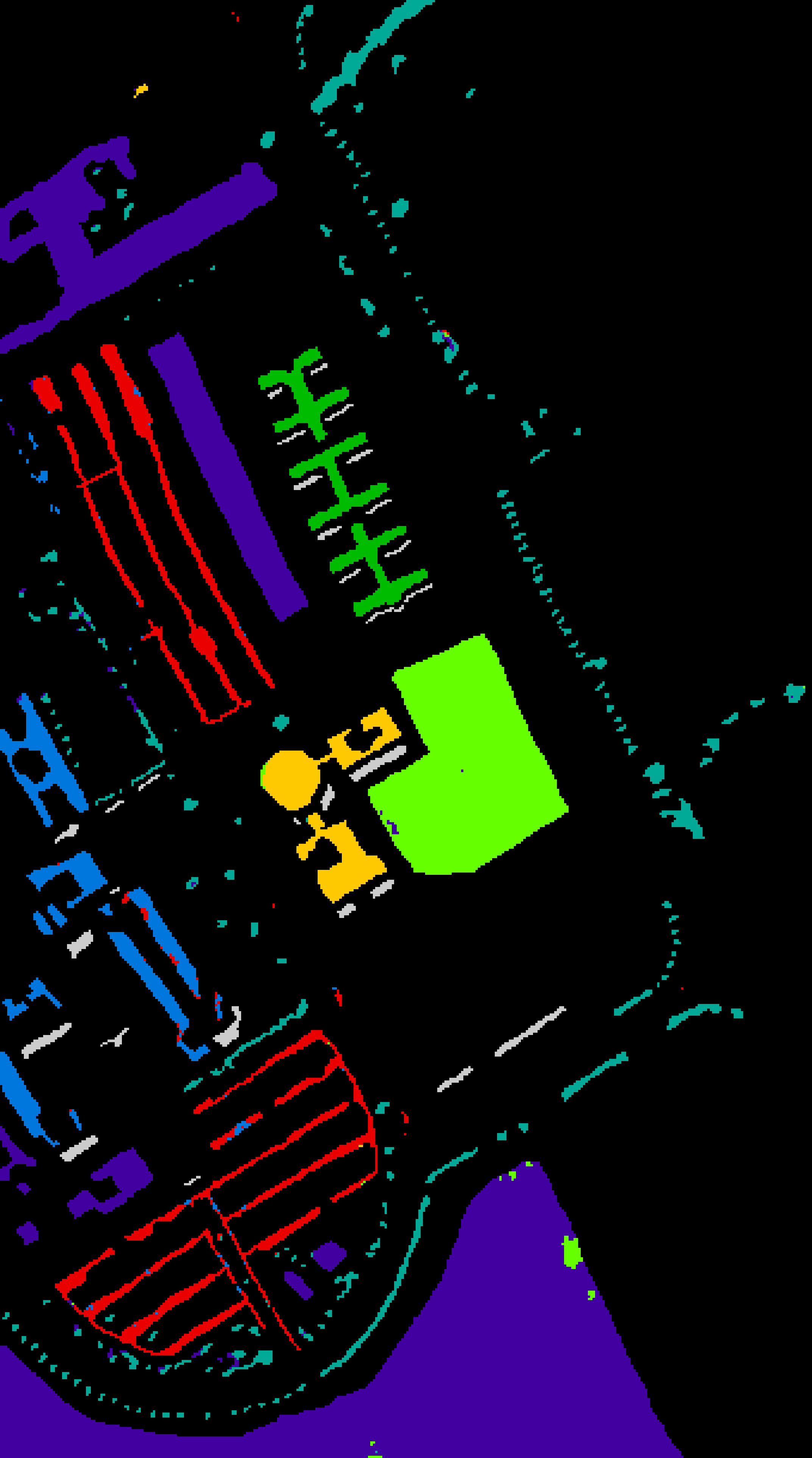}
            \caption{2DCNN}
        \end{subfigure}
        \begin{subfigure}{0.085\textwidth}
            \centering
            \includegraphics[width=0.99\textwidth]{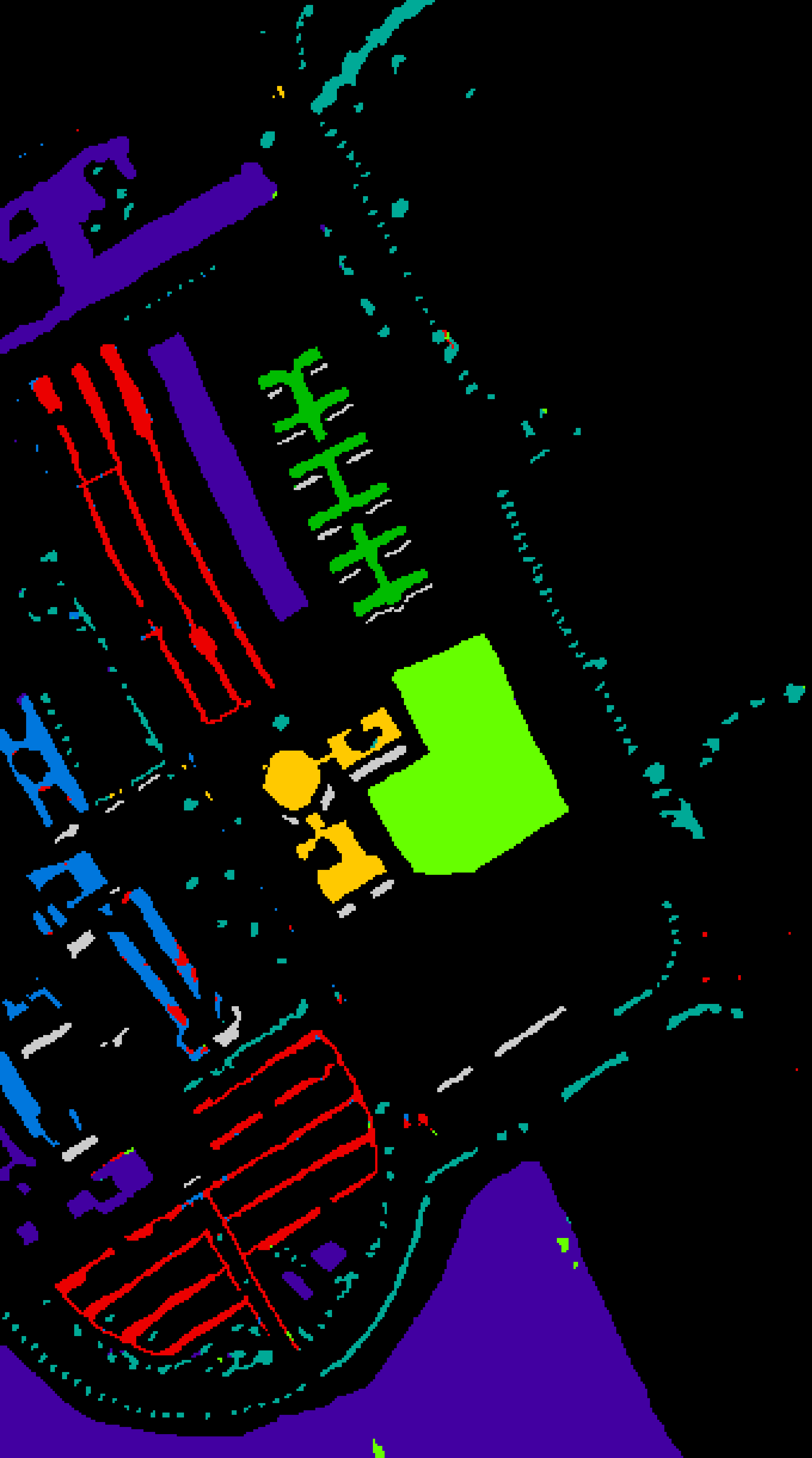}
            \caption{3DCNN}
        \end{subfigure}
        \begin{subfigure}{0.085\textwidth}
            \centering
            \includegraphics[width=0.99\textwidth]{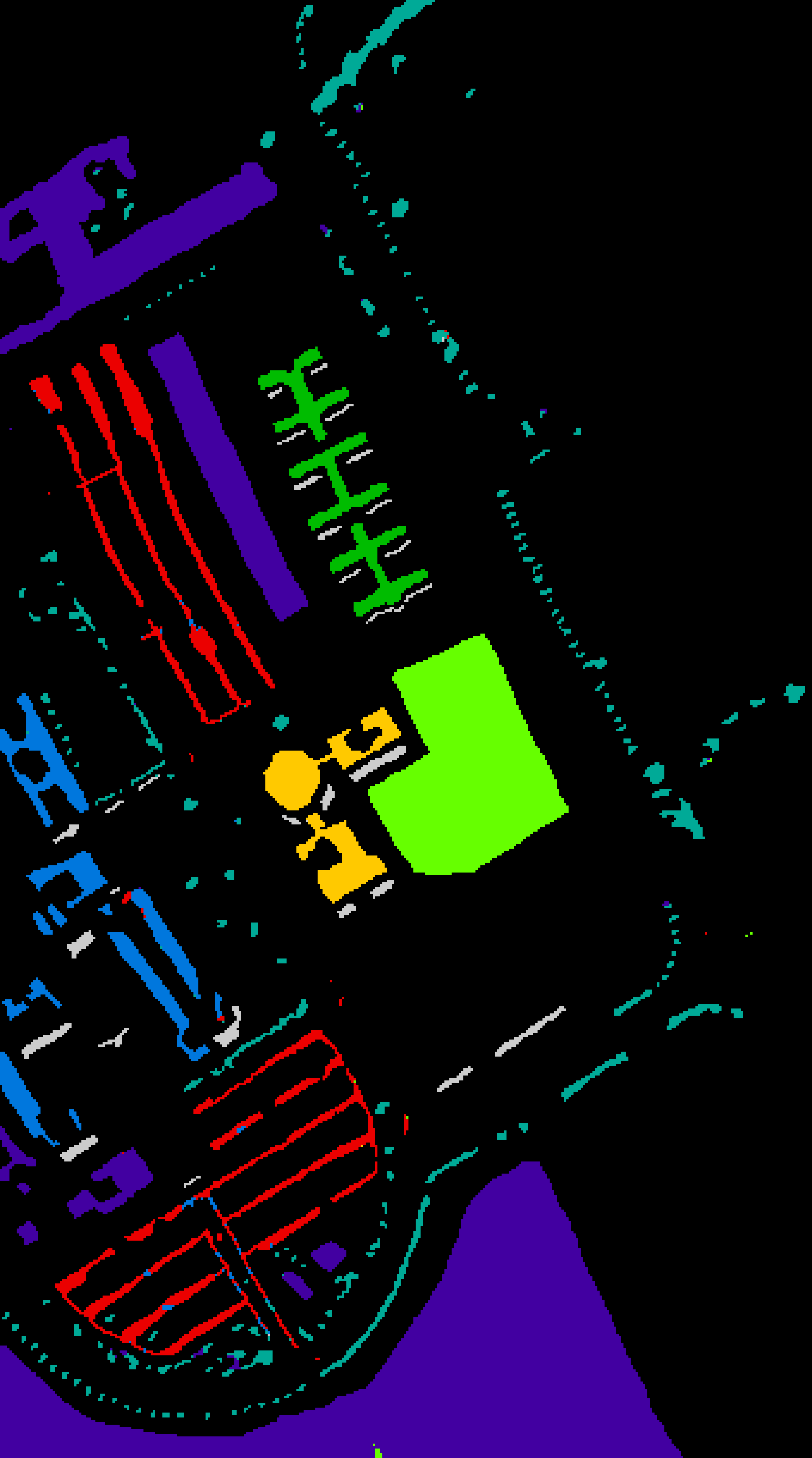}
            \caption{HCNN}
        \end{subfigure}
        \begin{subfigure}{0.085\textwidth}
            \centering
            \includegraphics[width=0.99\textwidth]{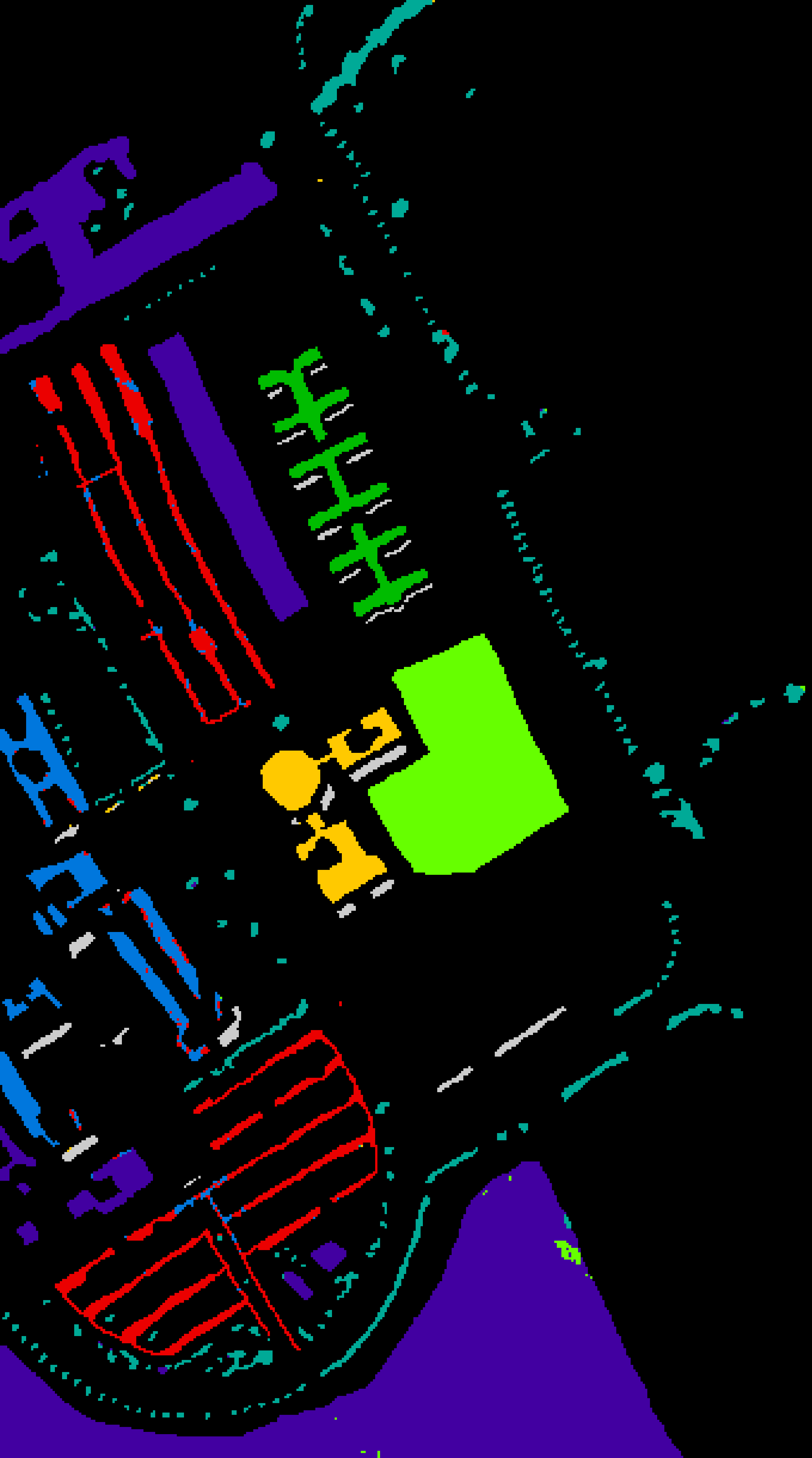}
            \caption{2DIN}
        \end{subfigure}
        \begin{subfigure}{0.085\textwidth}
            \centering
            \includegraphics[width=0.99\textwidth]{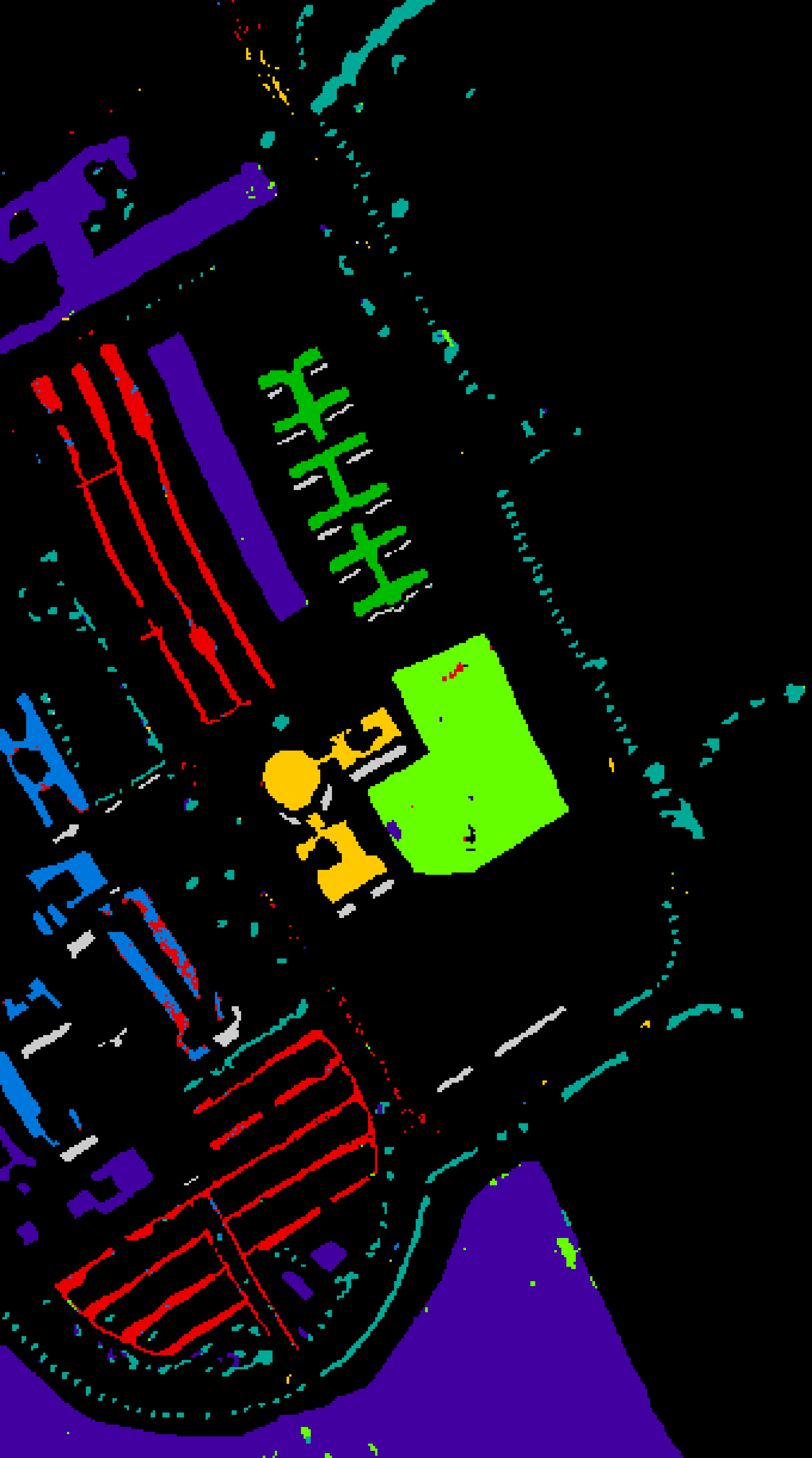}
            \caption{3DIN}
        \end{subfigure}
        \begin{subfigure}{0.085\textwidth}
            \centering
            \includegraphics[width=0.99\textwidth]{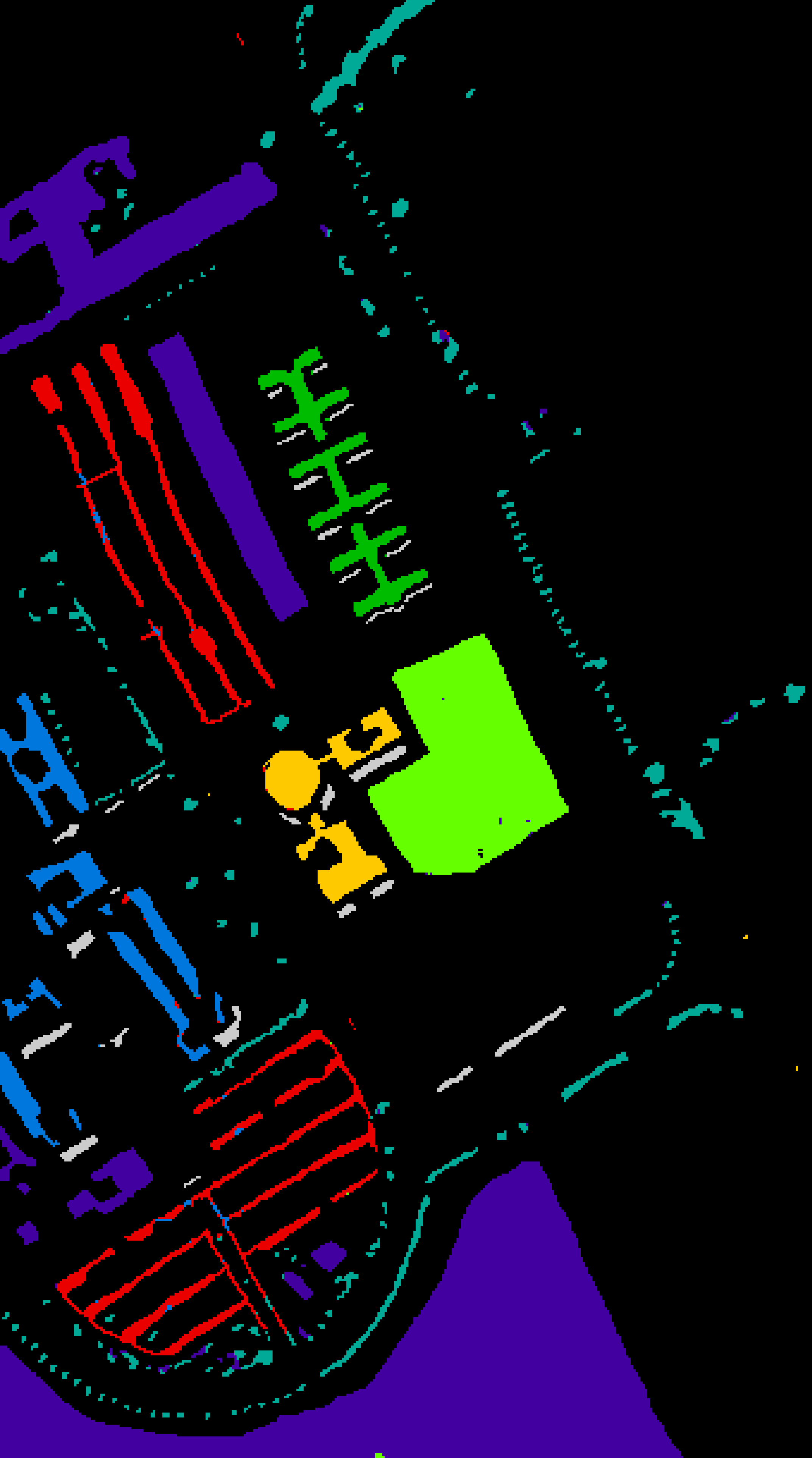}
            \caption{HyIN}
        \end{subfigure}
        \begin{subfigure}{0.085\textwidth}
            \centering
            \includegraphics[width=0.99\textwidth]{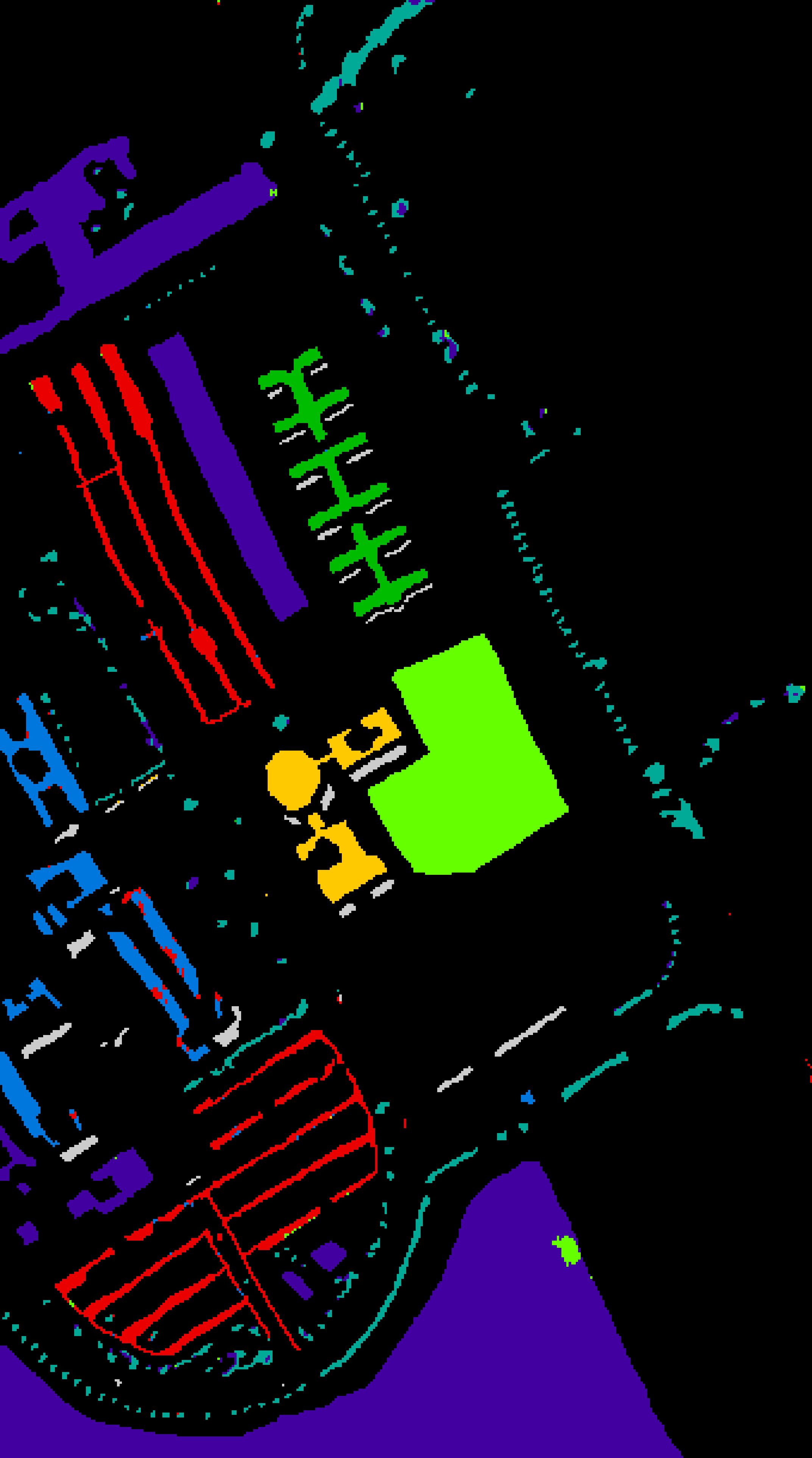}
            \caption{GCNN}
        \end{subfigure}
        \begin{subfigure}{0.085\textwidth}
            \centering
            \includegraphics[width=0.99\textwidth]{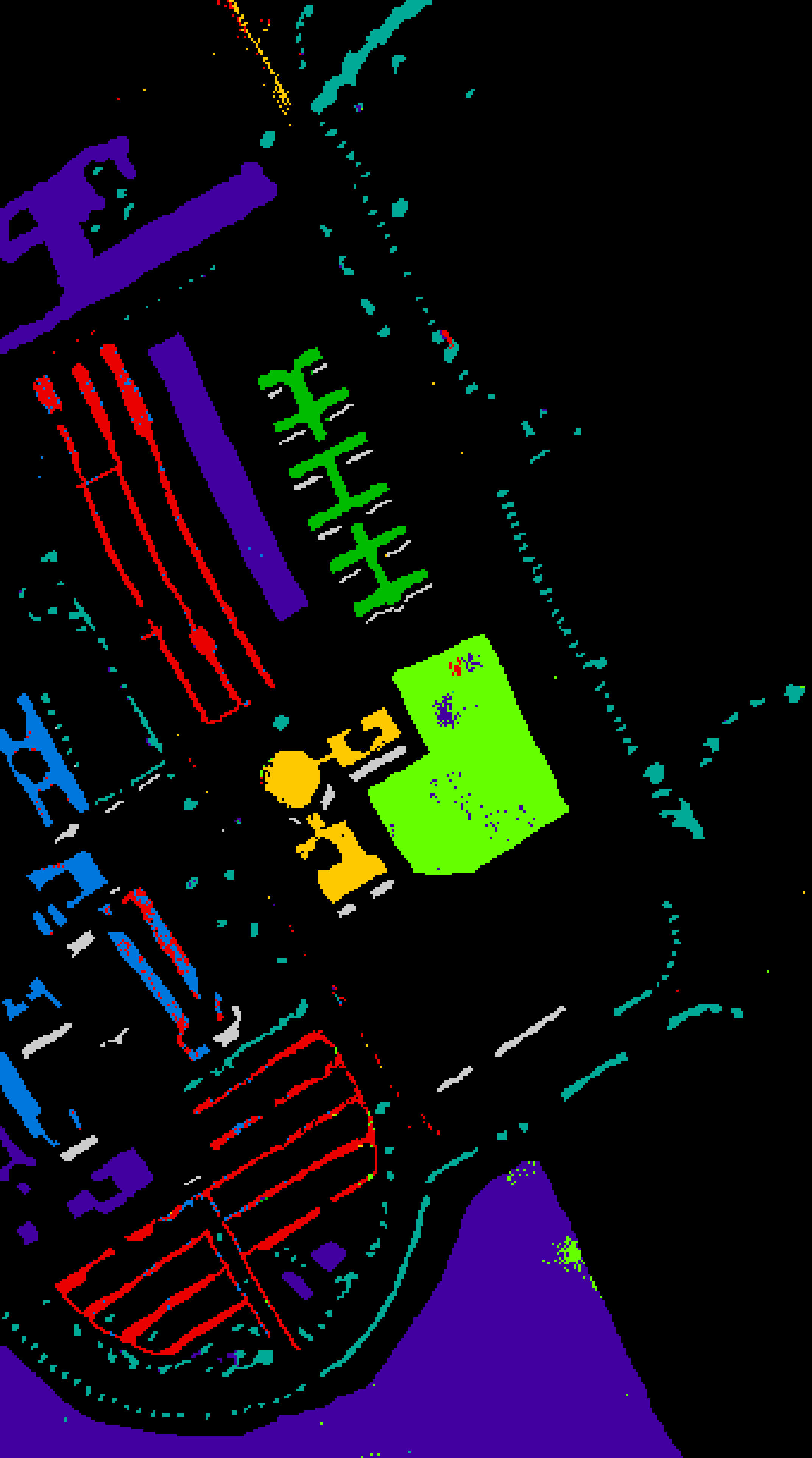}
            \caption{HViT}
        \end{subfigure}
        \begin{subfigure}{0.085\textwidth}
            \centering
            \includegraphics[width=0.99\textwidth]{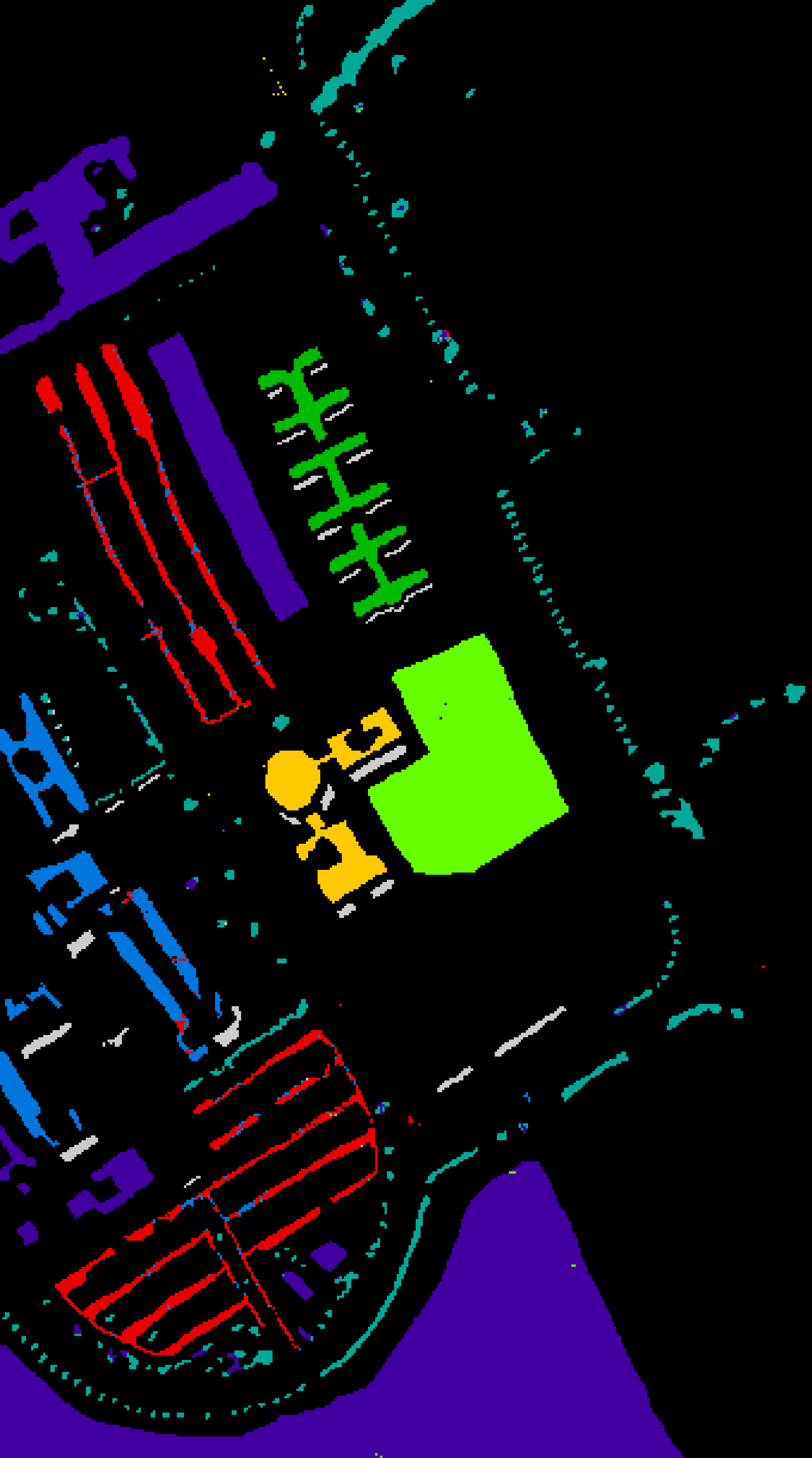}
            \caption{Hir\_ViT}
        \end{subfigure}
        \begin{subfigure}{0.085\textwidth}
            \centering
            \includegraphics[width=0.99\textwidth]{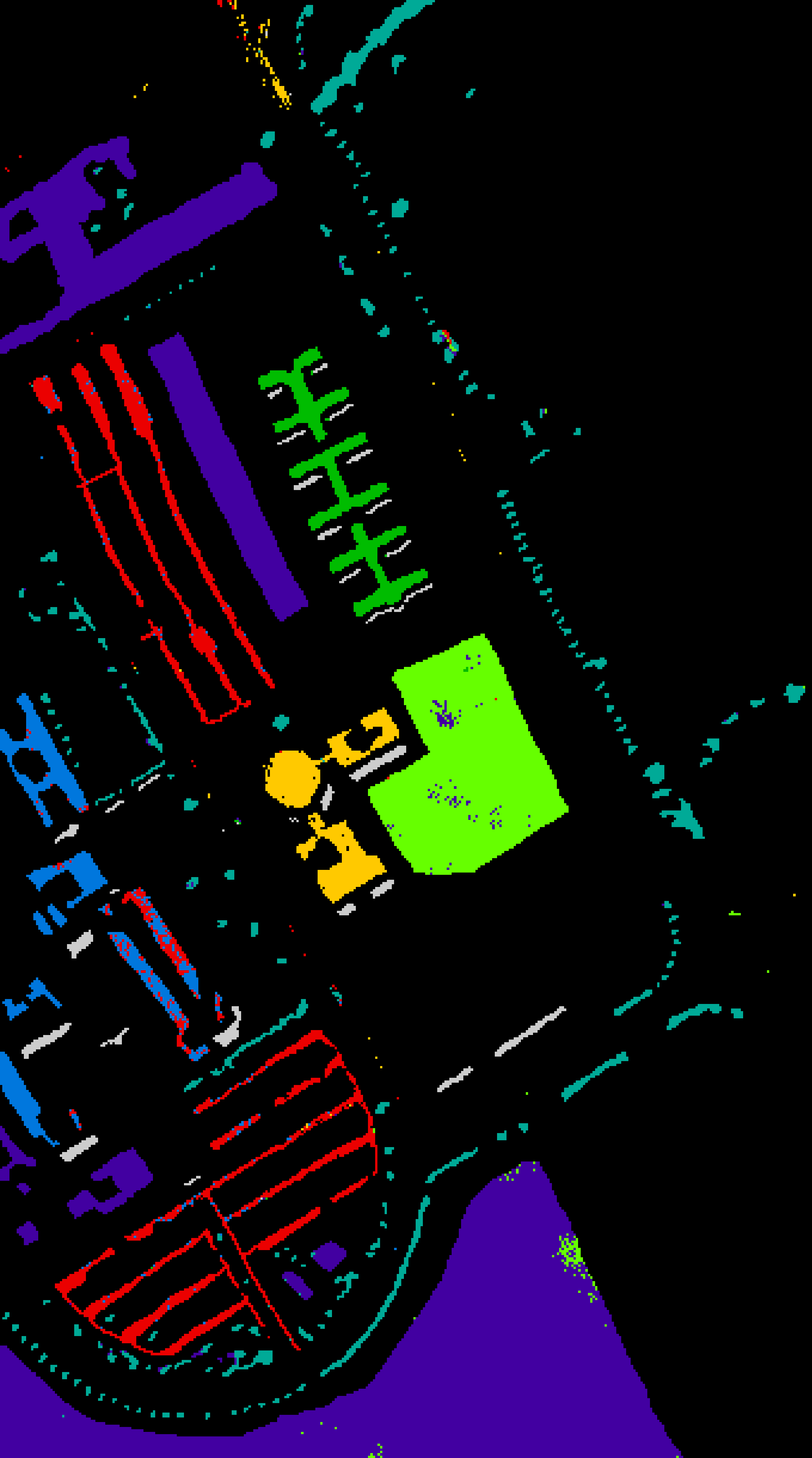}
            \caption{SST}
        \end{subfigure}
        \begin{subfigure}{0.085\textwidth}
            \centering
            \includegraphics[width=0.99\textwidth]{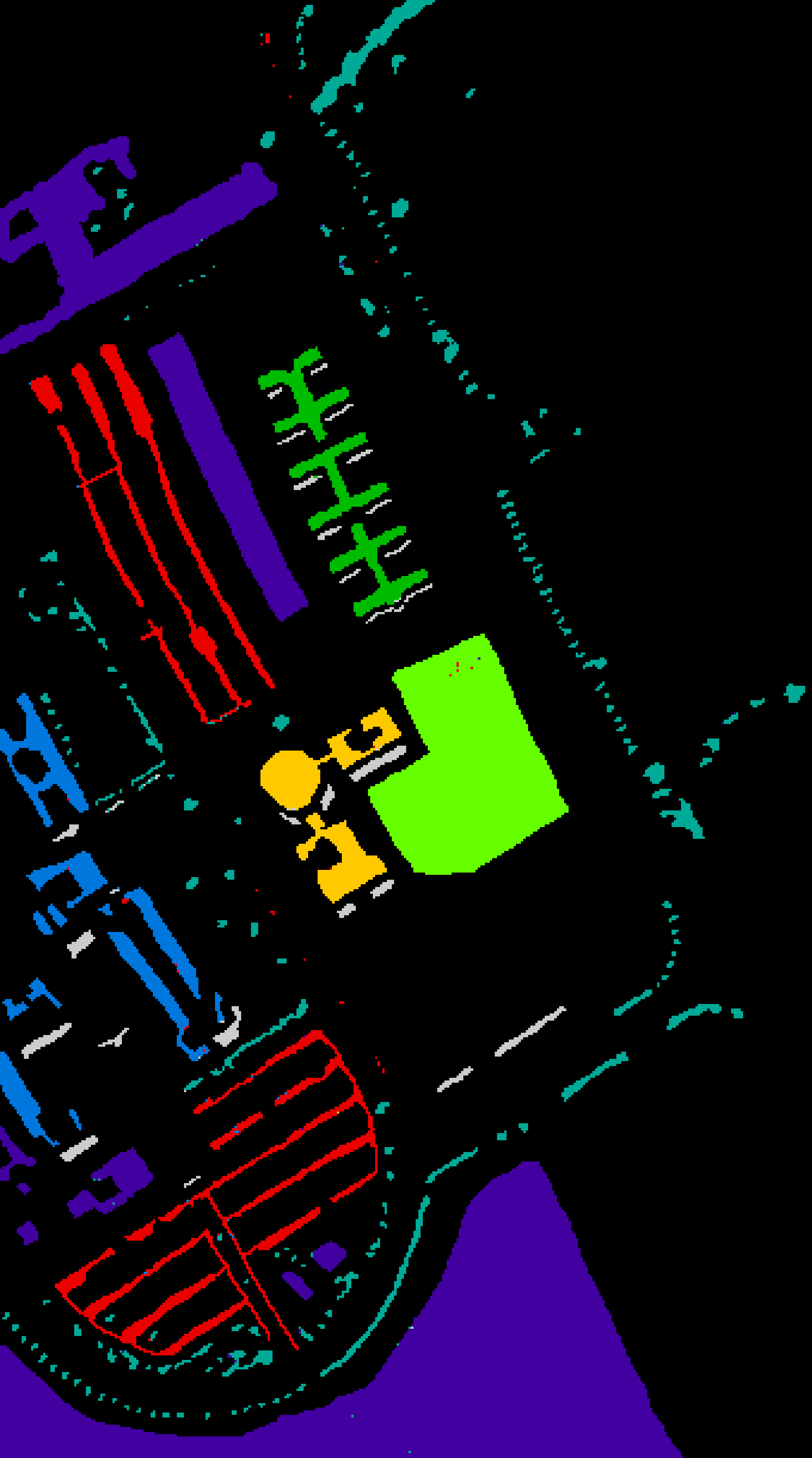}
            \caption{GMamba}
        \end{subfigure}
    \caption{The predicted ground truth maps for the PU dataset are presented for various state-of-the-art methods along with GraphMamba.}
    \label{fig:PU_results}
\end{figure*}
\begin{figure*}[!htb]
    \centering
        \begin{subfigure}{0.085\textwidth}
            \includegraphics[width=0.99\textwidth]{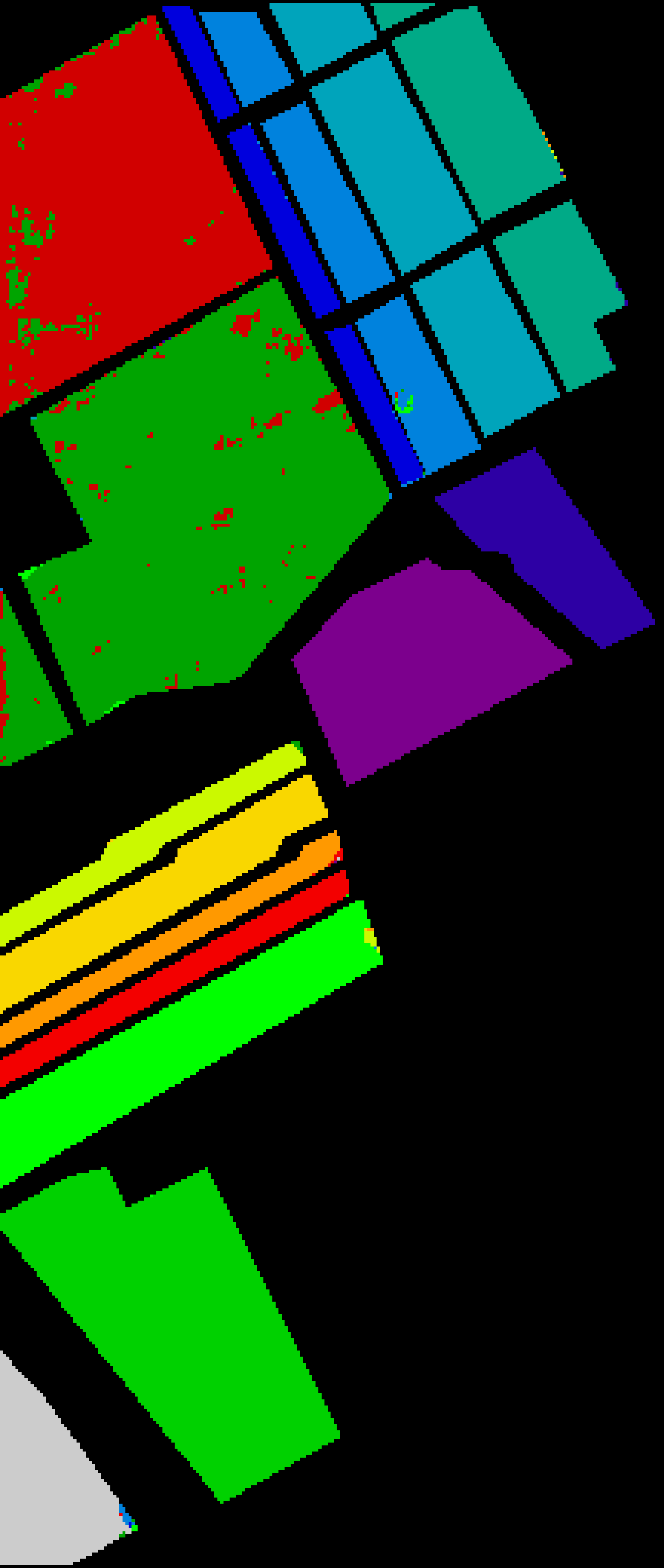}
            \caption{2DCNN}
        \end{subfigure}
        \begin{subfigure}{0.085\textwidth}
            \centering
            \includegraphics[width=0.99\textwidth]{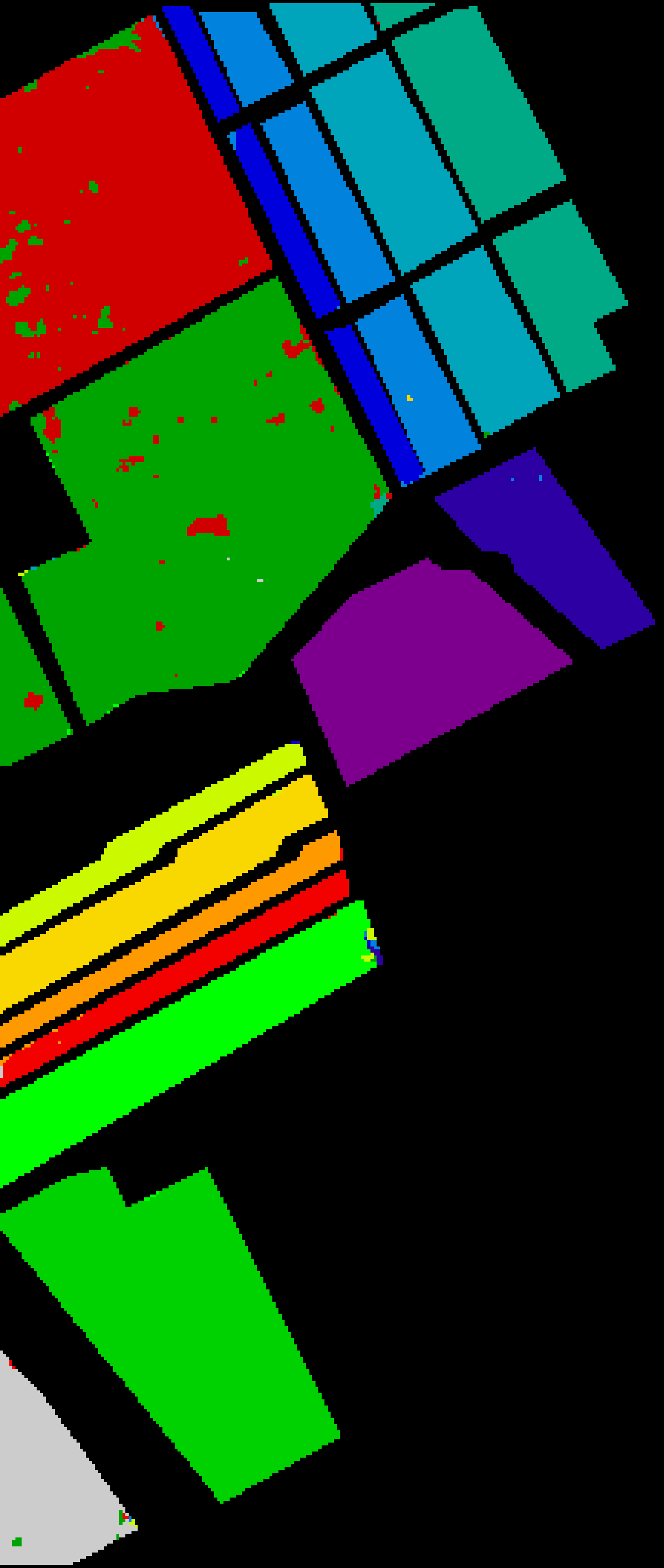}
            \caption{3DCNN}
        \end{subfigure}
        \begin{subfigure}{0.085\textwidth}
            \centering
            \includegraphics[width=0.99\textwidth]{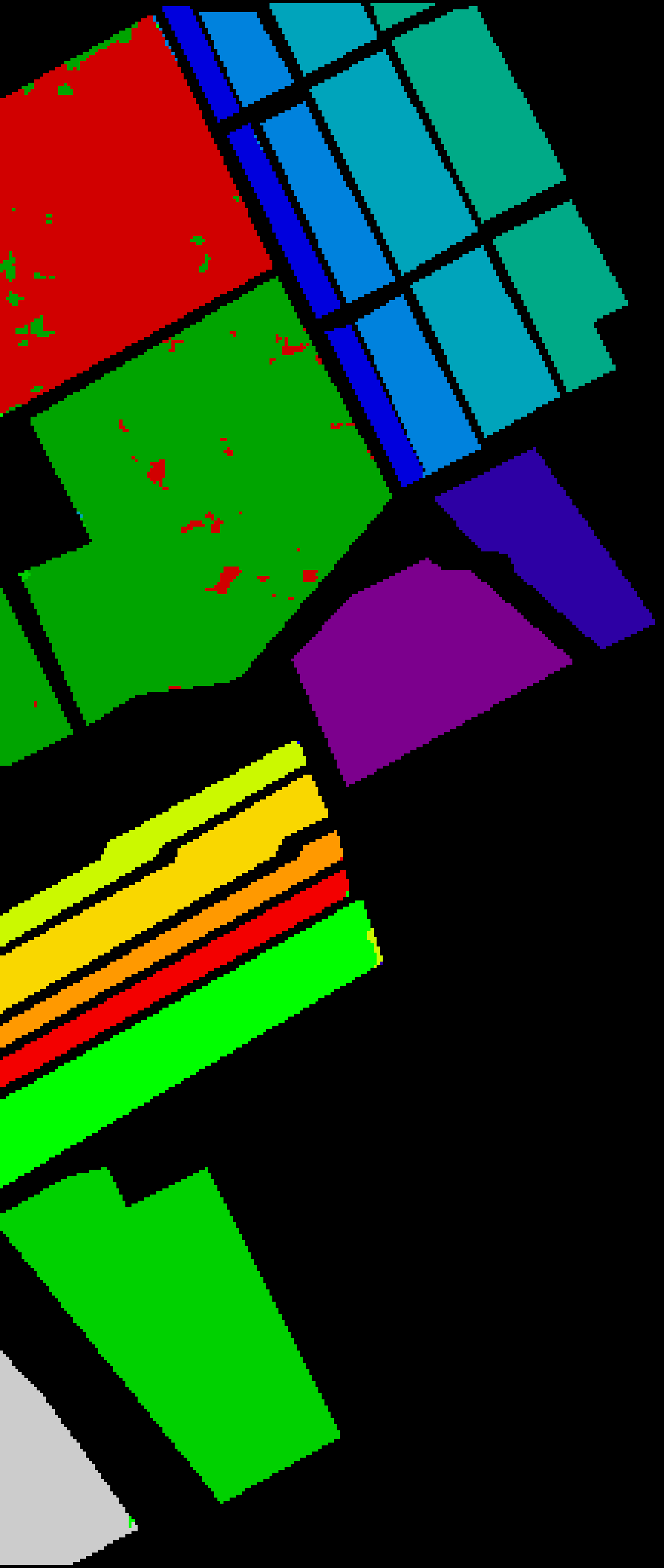}
            \caption{HCNN}
        \end{subfigure}
        \begin{subfigure}{0.085\textwidth}
            \centering
            \includegraphics[width=0.99\textwidth]{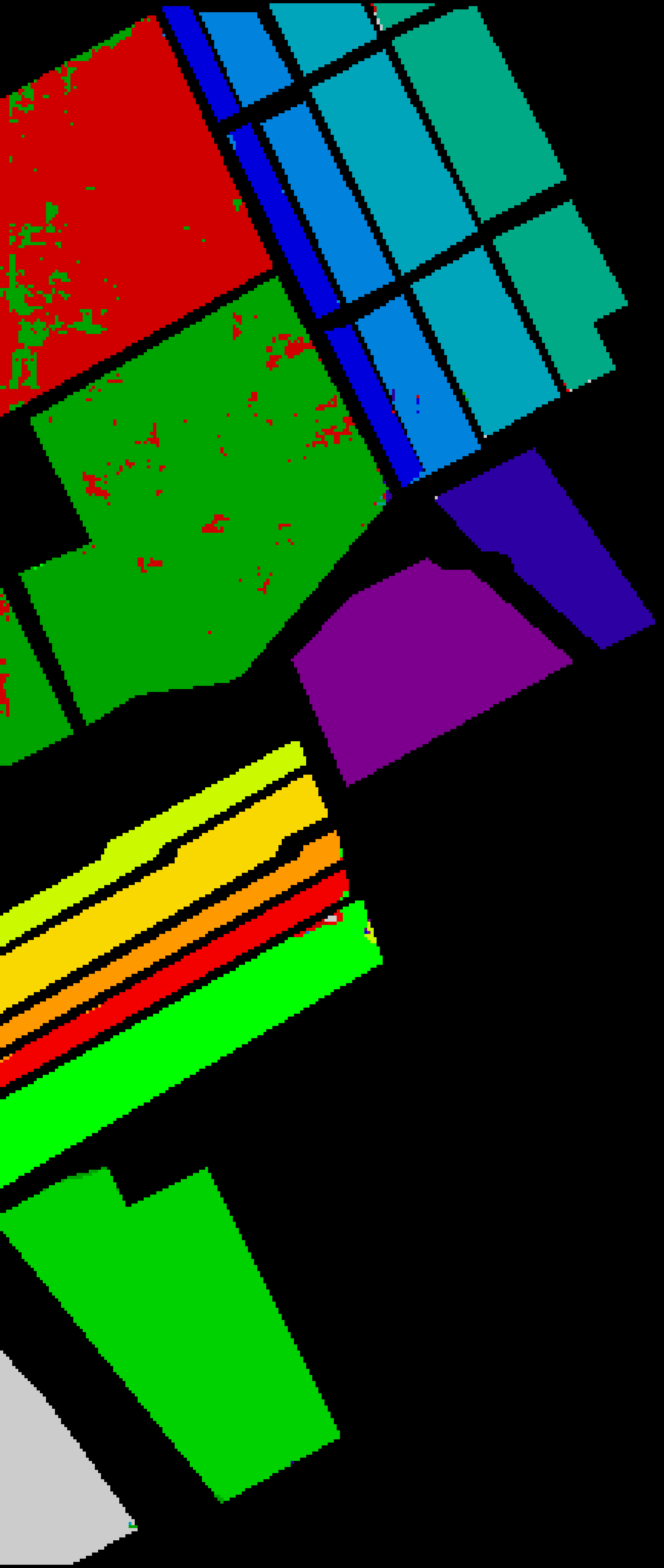}
            \caption{2DIN}
        \end{subfigure}
        \begin{subfigure}{0.085\textwidth}
            \centering
            \includegraphics[width=0.99\textwidth]{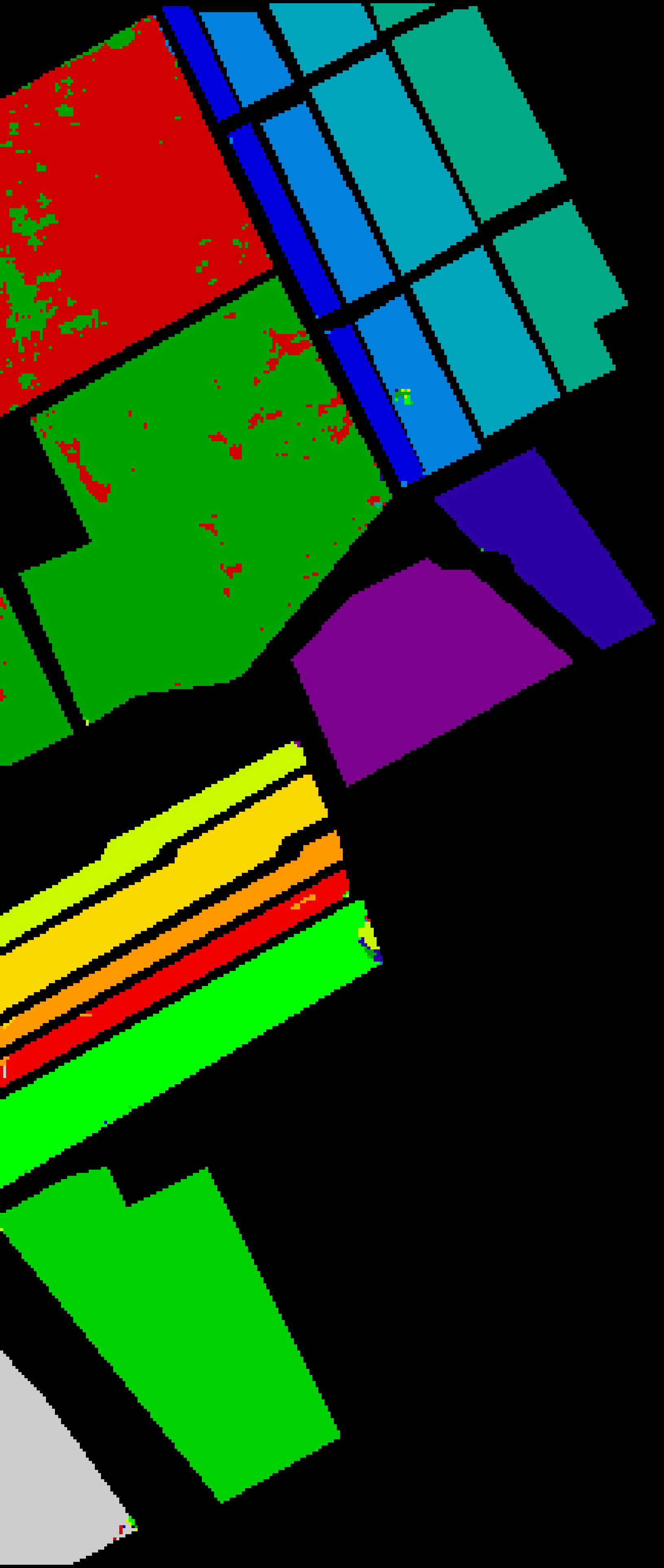}
            \caption{3DIN}
        \end{subfigure}
        \begin{subfigure}{0.085\textwidth}
            \centering
            \includegraphics[width=0.99\textwidth]{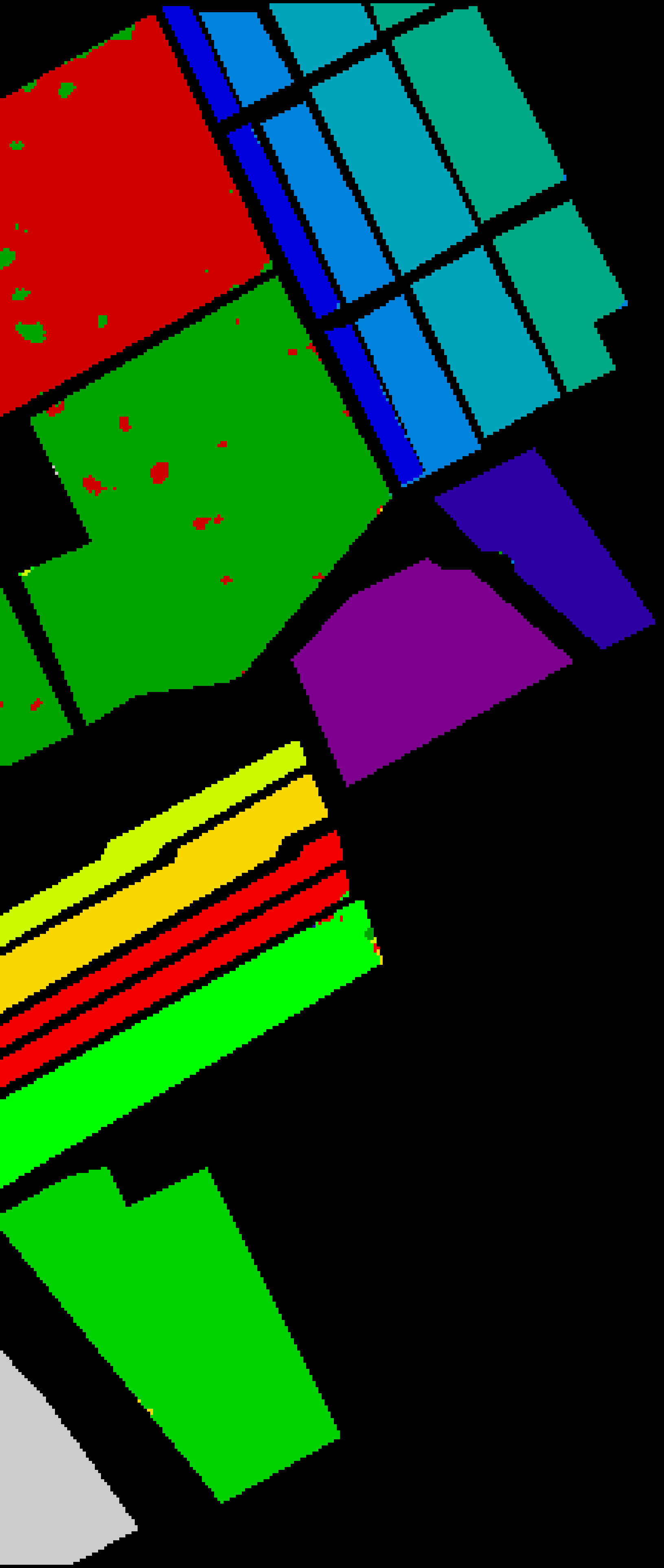}
            \caption{HybIN}
        \end{subfigure}
        \begin{subfigure}{0.085\textwidth}
            \centering
            \includegraphics[width=0.99\textwidth]{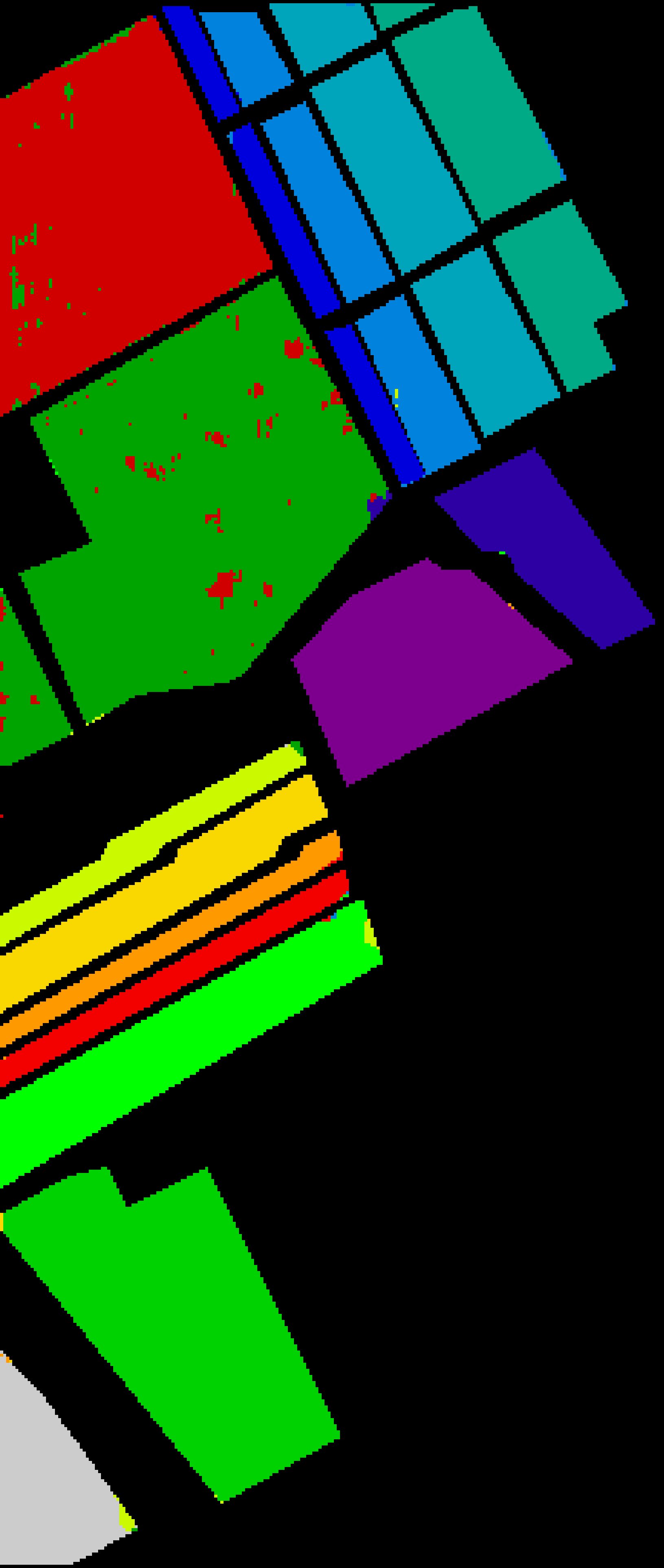}
            \caption{GCNN}
        \end{subfigure}
        \begin{subfigure}{0.085\textwidth}
            \centering
            \includegraphics[width=0.99\textwidth]{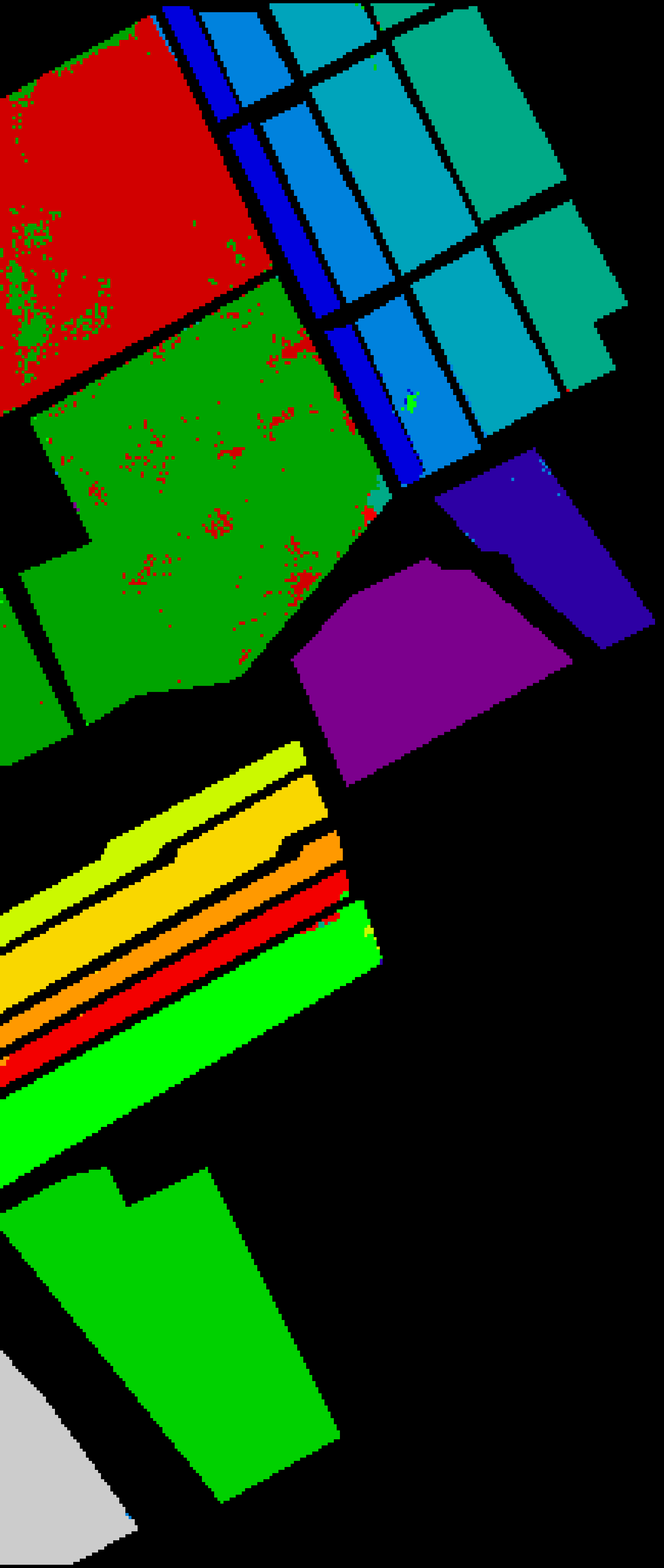}
            \caption{HViT}
        \end{subfigure}
        \begin{subfigure}{0.085\textwidth}
            \centering
            \includegraphics[width=0.99\textwidth]{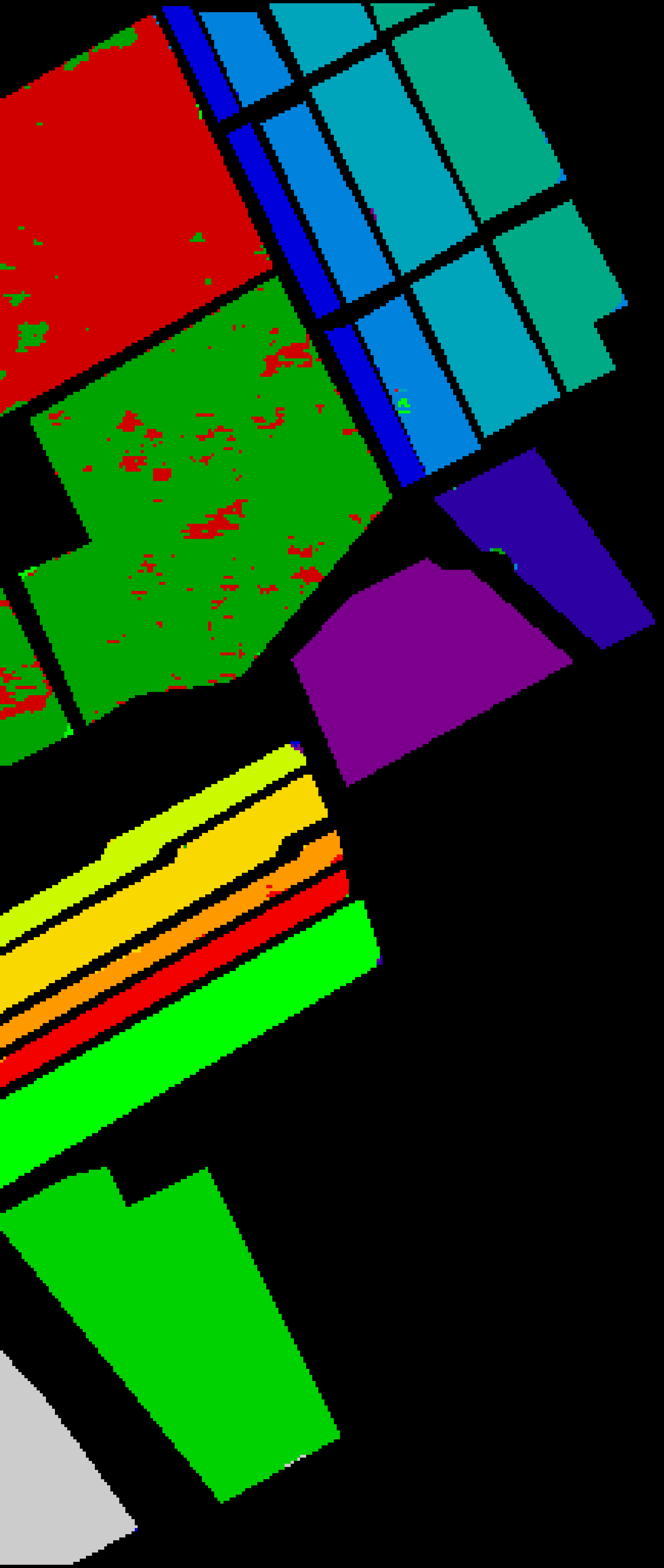}
            \caption{Hir\_ViT}
        \end{subfigure}
        \begin{subfigure}{0.085\textwidth}
            \centering
            \includegraphics[width=0.99\textwidth]{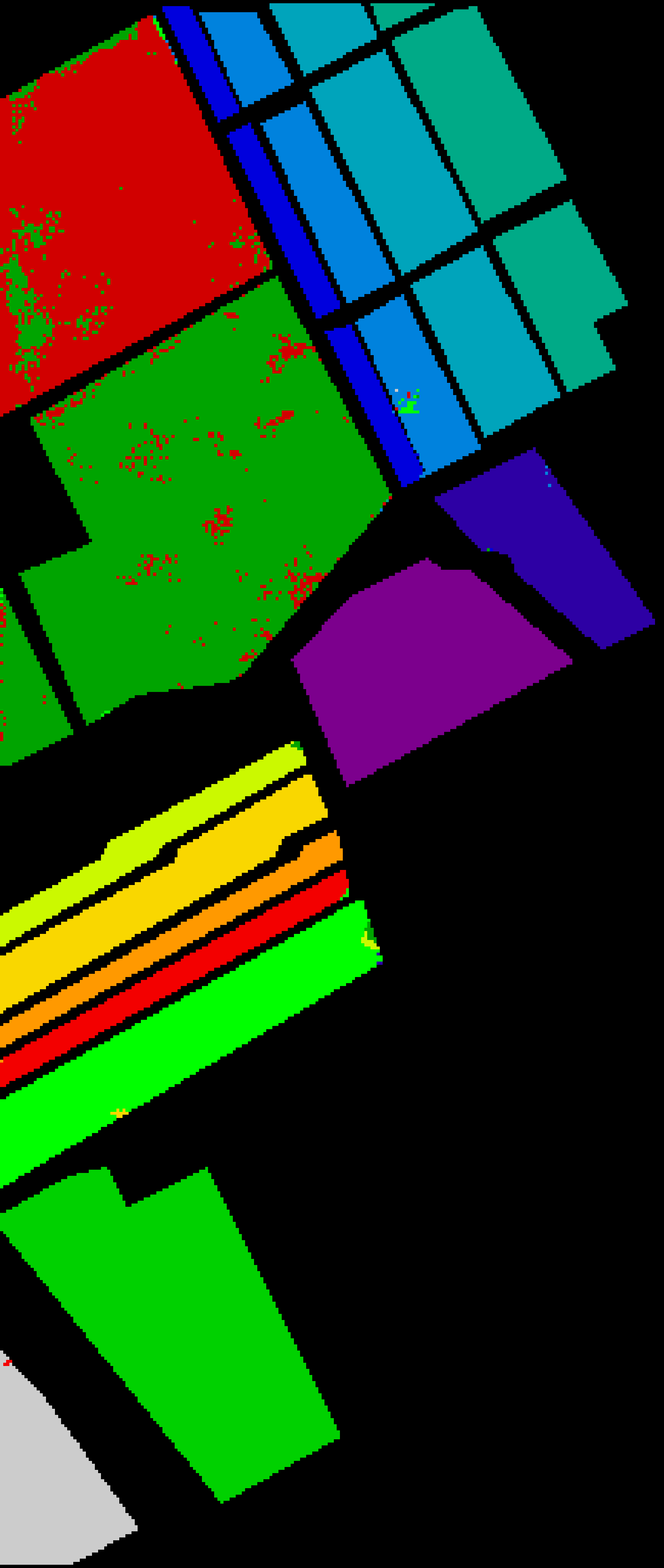}
            \caption{SST}
        \end{subfigure}
        \begin{subfigure}{0.085\textwidth}
            \centering
            \includegraphics[width=0.99\textwidth]{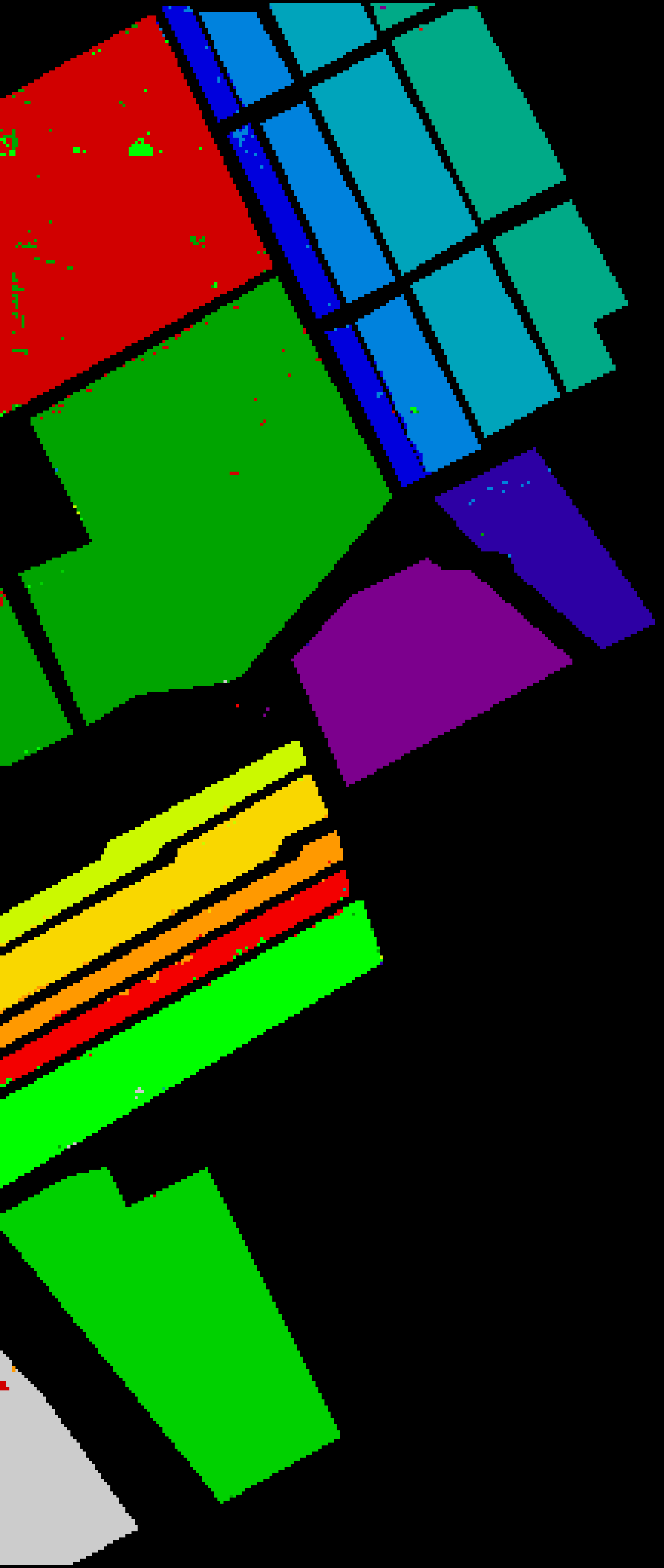}
            \caption{GMamba}
        \end{subfigure}
    \caption{The predicted ground truth maps for the SA dataset are presented for various state-of-the-art methods along with GraphMamba.}
    \label{fig:SA_results}
\end{figure*}
\begin{figure*}[!htb]
    \centering
        \begin{subfigure}{0.085\textwidth}
            \includegraphics[width=0.99\textwidth]{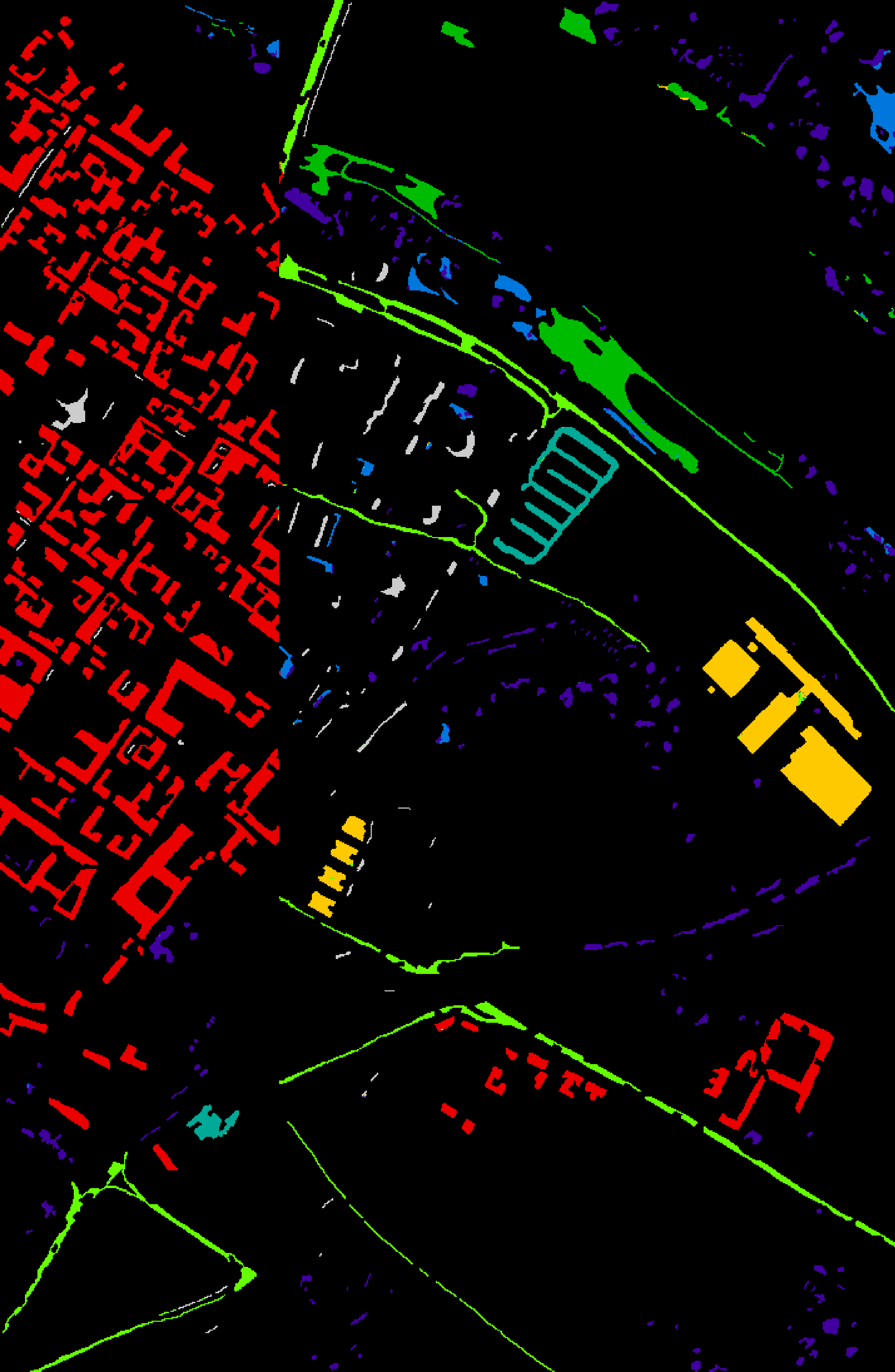}
            \caption{2DCNN}
        \end{subfigure}
        \begin{subfigure}{0.085\textwidth}
            \centering
            \includegraphics[width=0.99\textwidth]{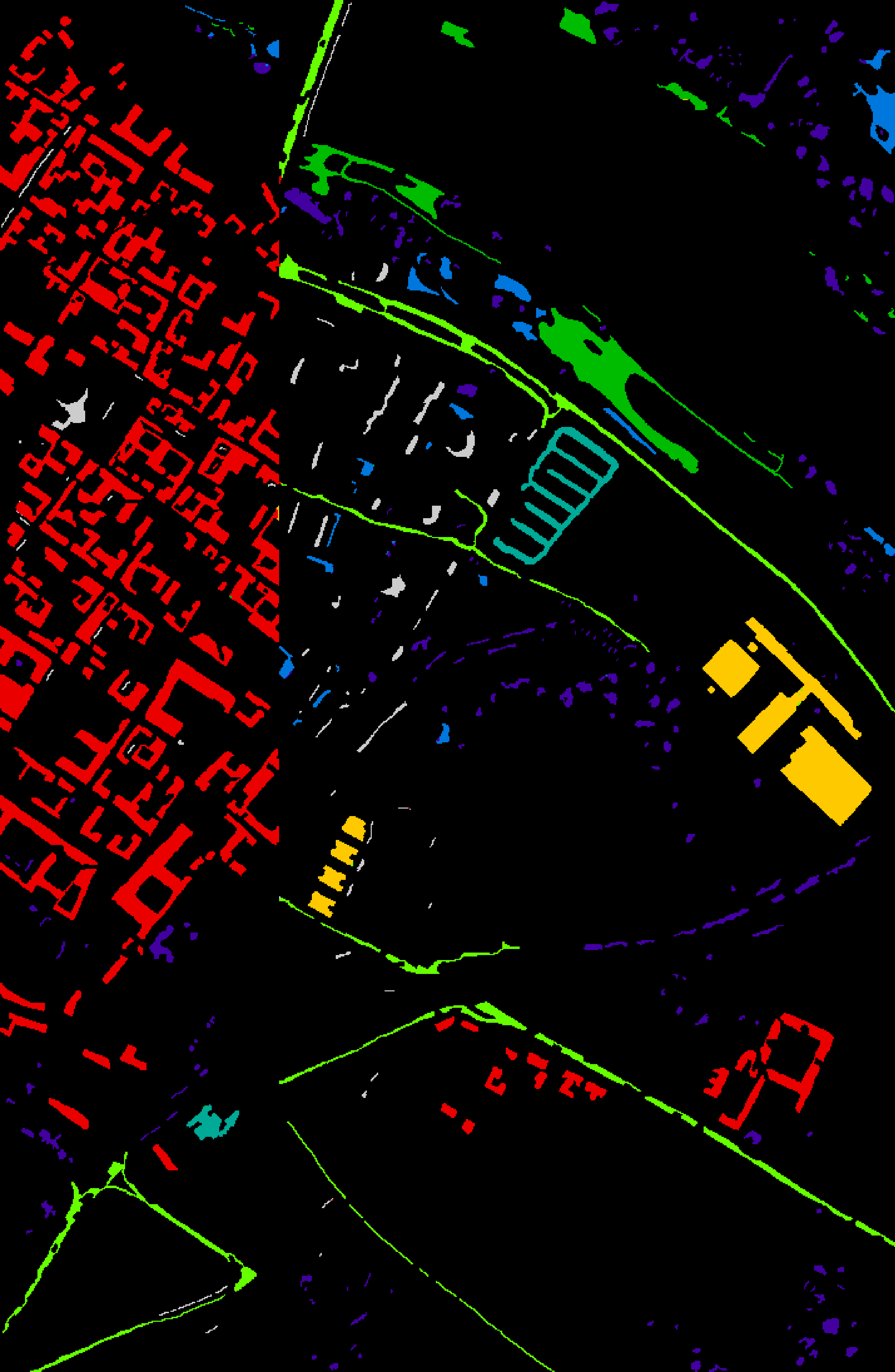}
            \caption{3DCNN}
        \end{subfigure}
        \begin{subfigure}{0.085\textwidth}
            \centering
            \includegraphics[width=0.99\textwidth]{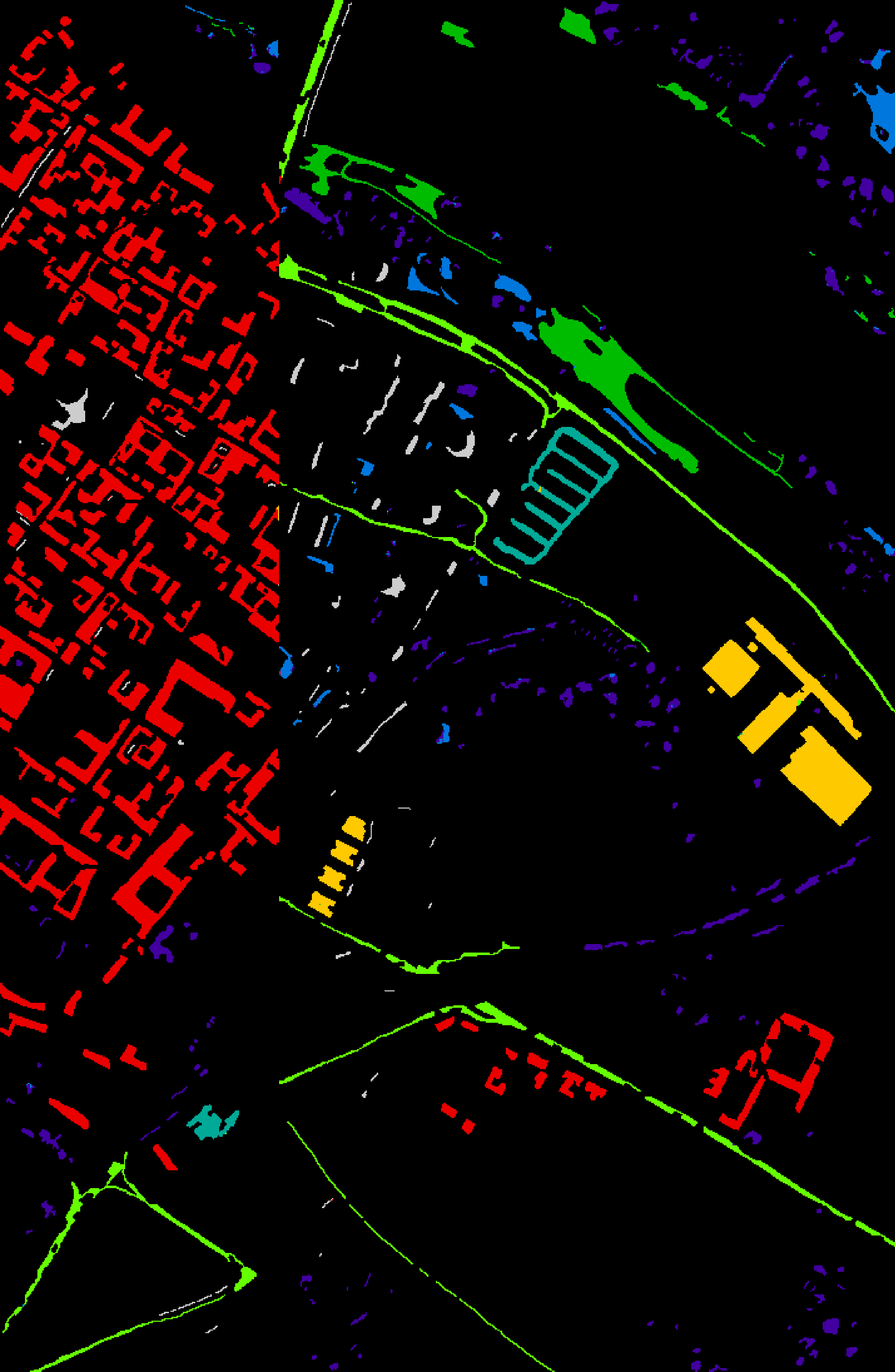}
            \caption{HCNN}
        \end{subfigure}
        \begin{subfigure}{0.085\textwidth}
            \centering
            \includegraphics[width=0.99\textwidth]{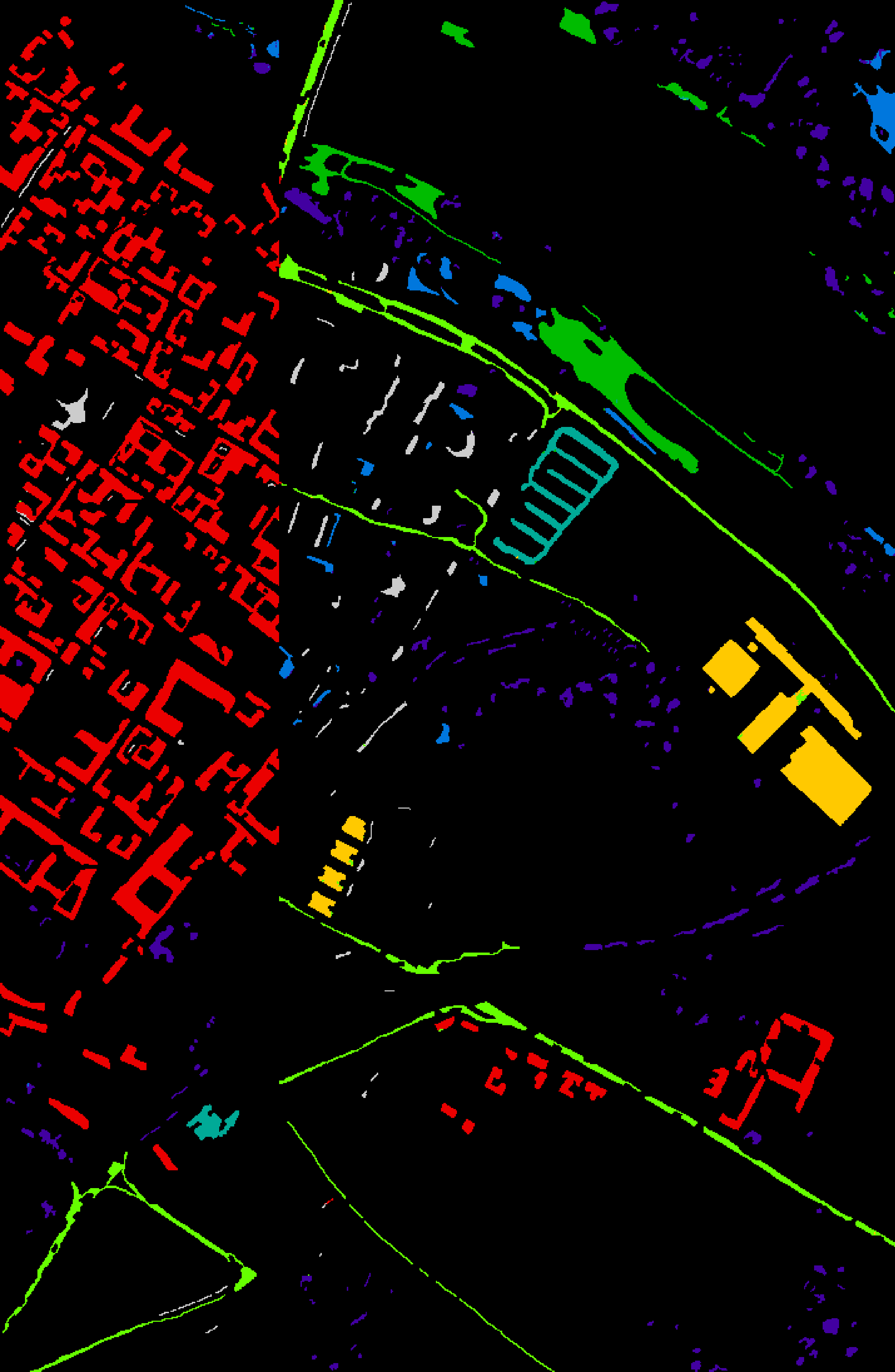}
            \caption{2DIN}
        \end{subfigure}
        \begin{subfigure}{0.085\textwidth}
            \centering
            \includegraphics[width=0.99\textwidth]{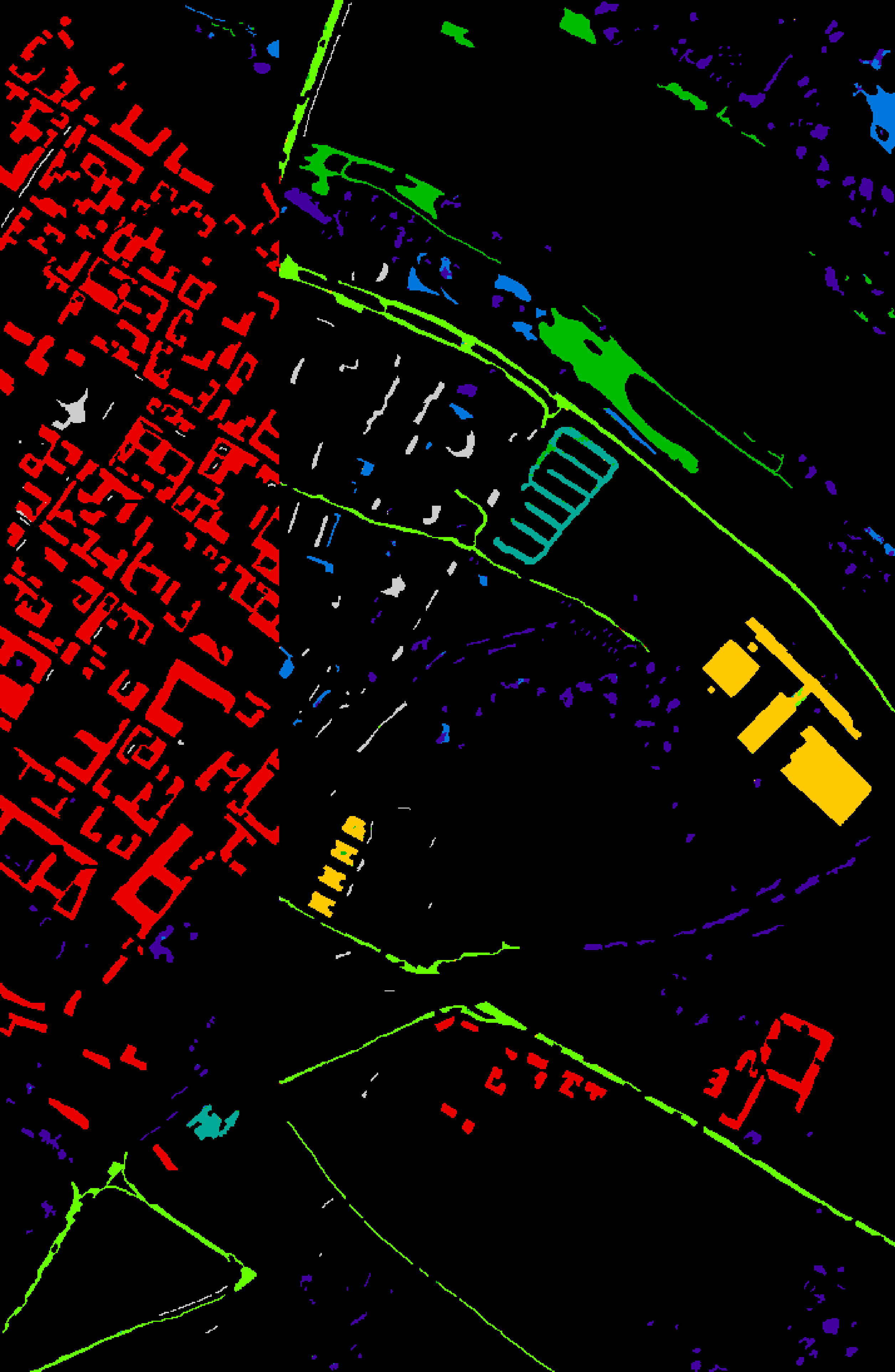}
            \caption{3DIN}
        \end{subfigure}
        \begin{subfigure}{0.085\textwidth}
            \centering
            \includegraphics[width=0.99\textwidth]{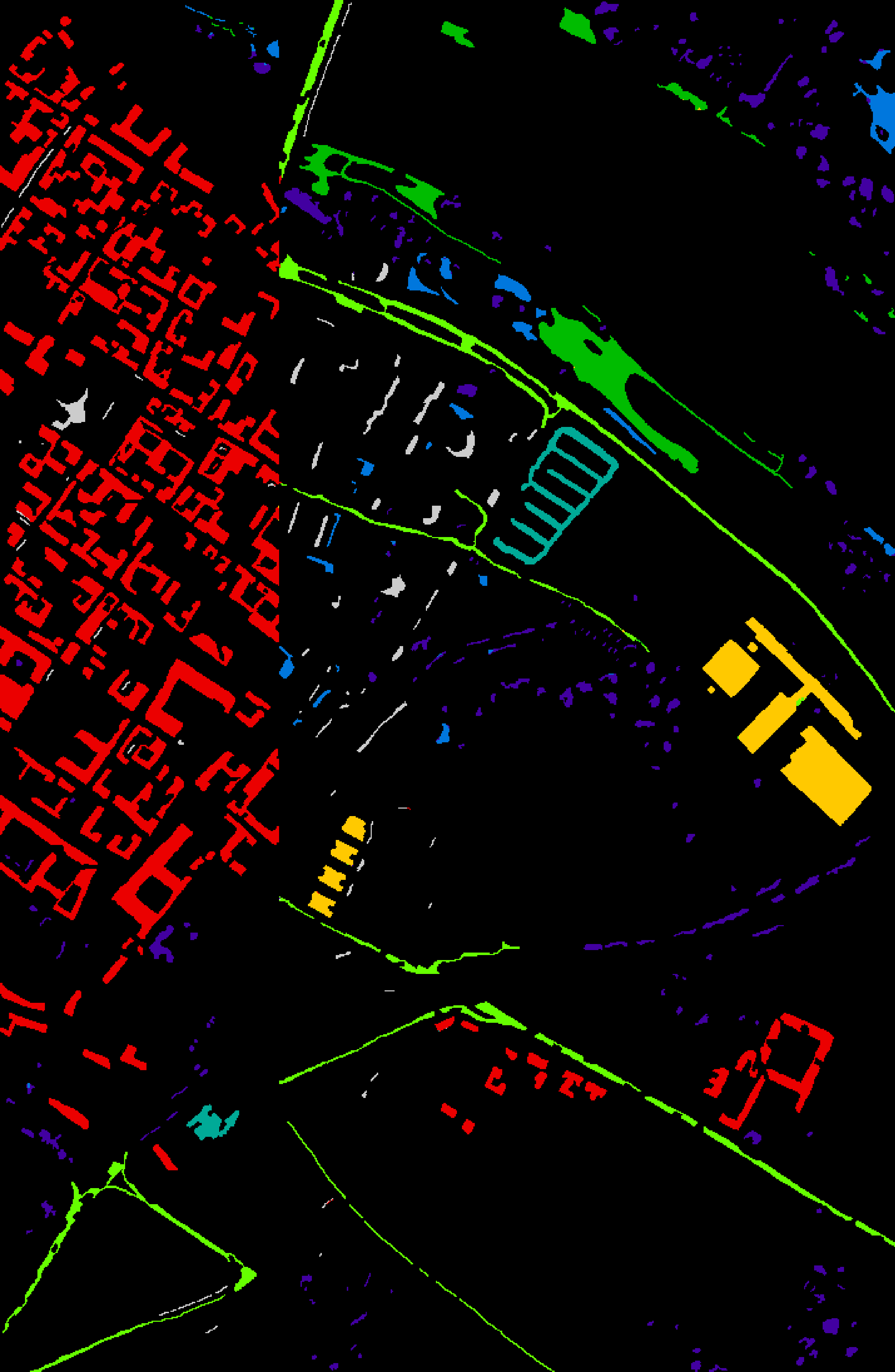}
            \caption{HIN}
        \end{subfigure}
        \begin{subfigure}{0.085\textwidth}
            \centering
            \includegraphics[width=0.99\textwidth]{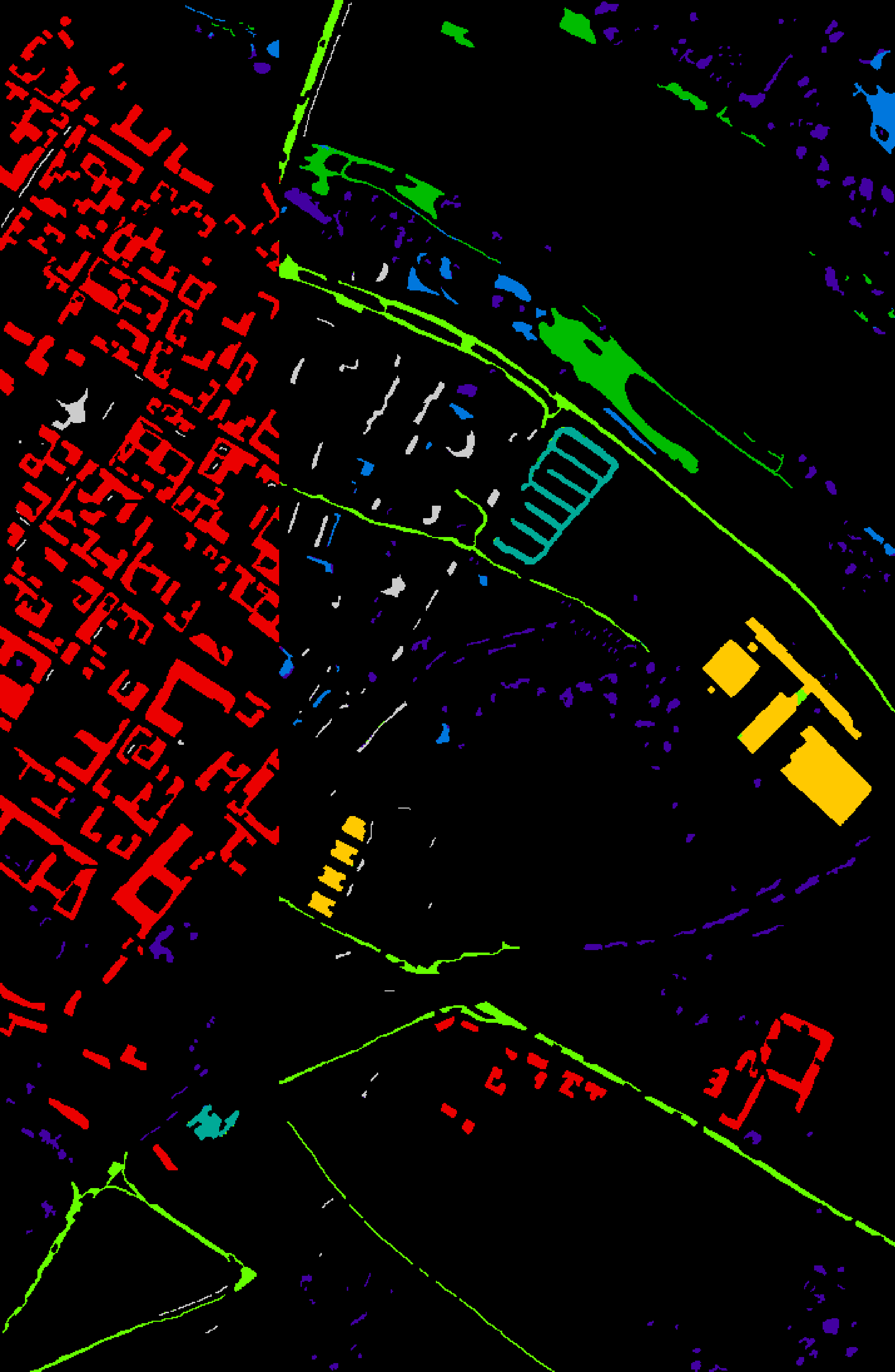}
            \caption{GCNN}
        \end{subfigure}
        \begin{subfigure}{0.085\textwidth}
            \centering
            \includegraphics[width=0.99\textwidth]{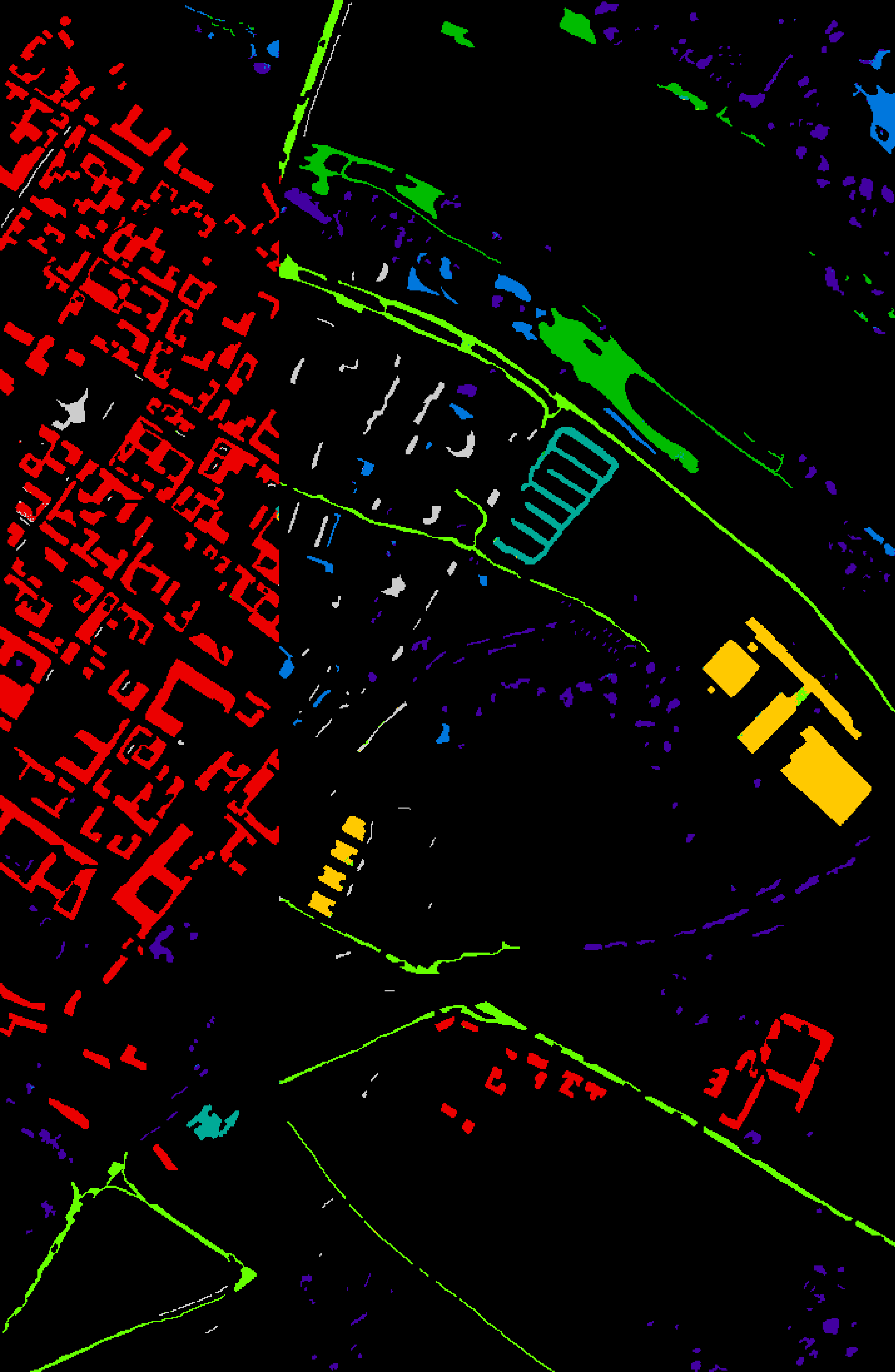}
            \caption{HViT}
        \end{subfigure}
        \begin{subfigure}{0.085\textwidth}
            \centering
            \includegraphics[width=0.99\textwidth]{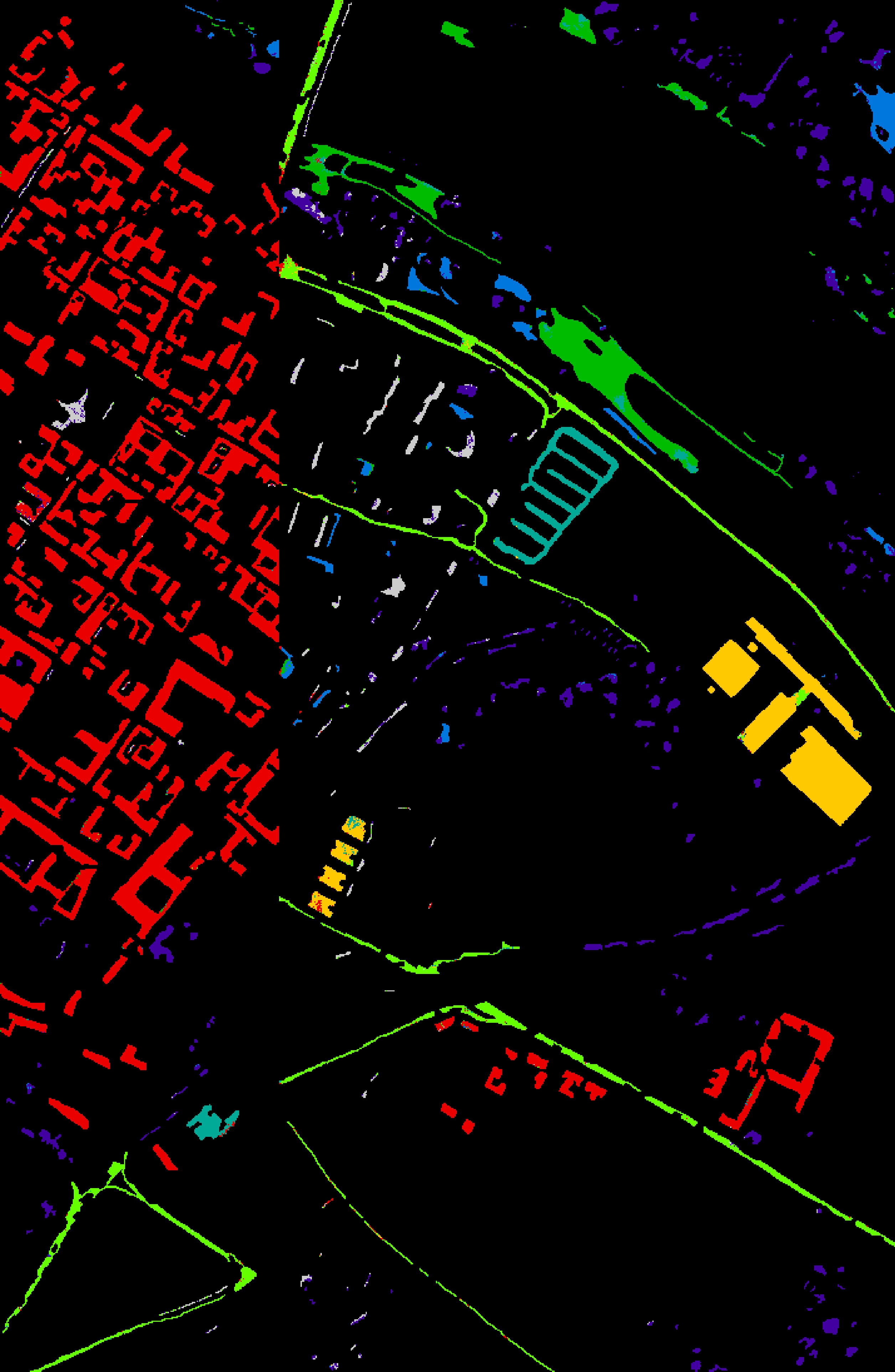}
            \caption{Hir\_ViT}
        \end{subfigure}
        \begin{subfigure}{0.085\textwidth}
            \centering
            \includegraphics[width=0.99\textwidth]{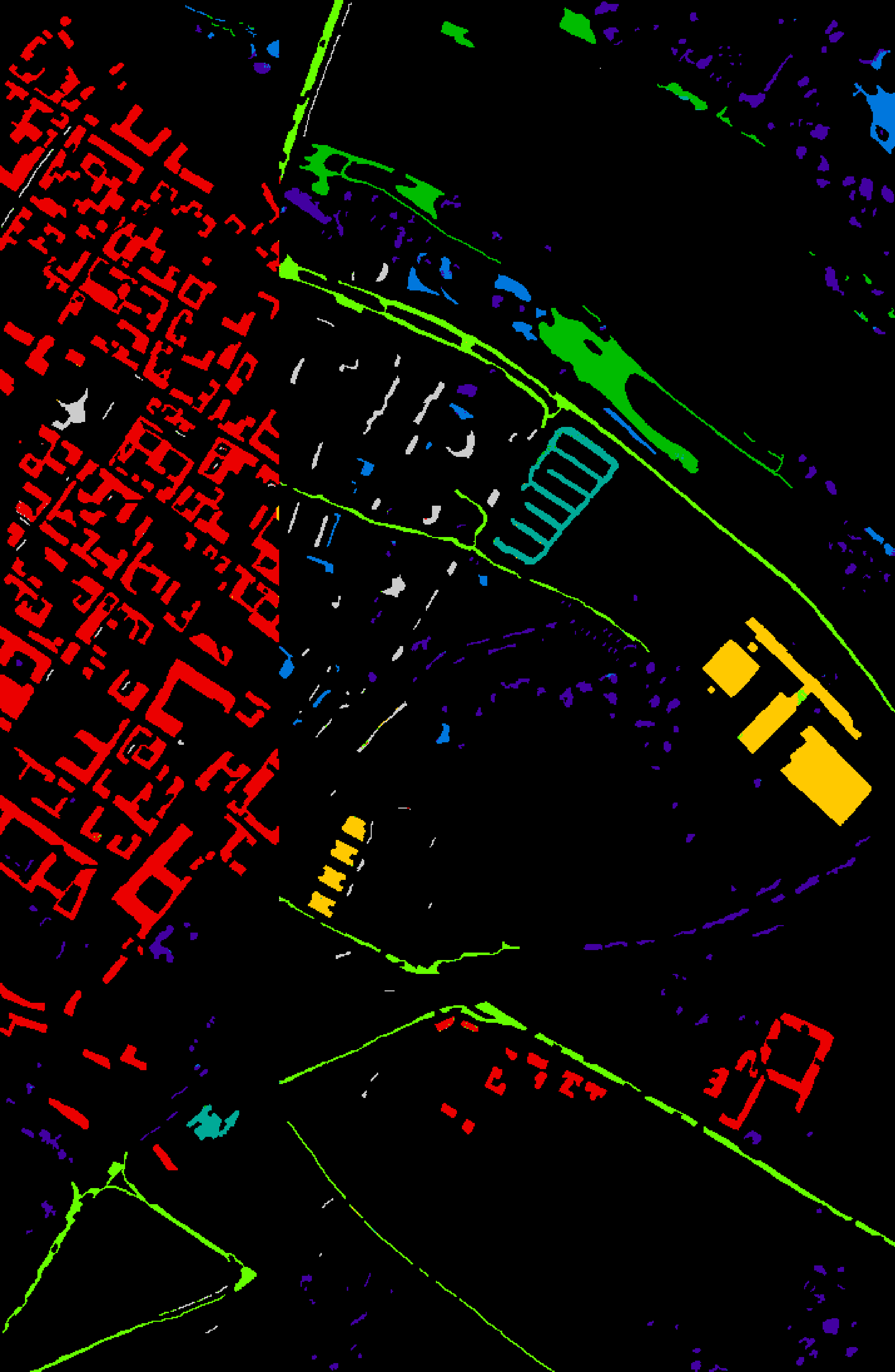}
            \caption{SST}
        \end{subfigure}
        \begin{subfigure}{0.085\textwidth}
            \centering
            \includegraphics[width=0.99\textwidth]{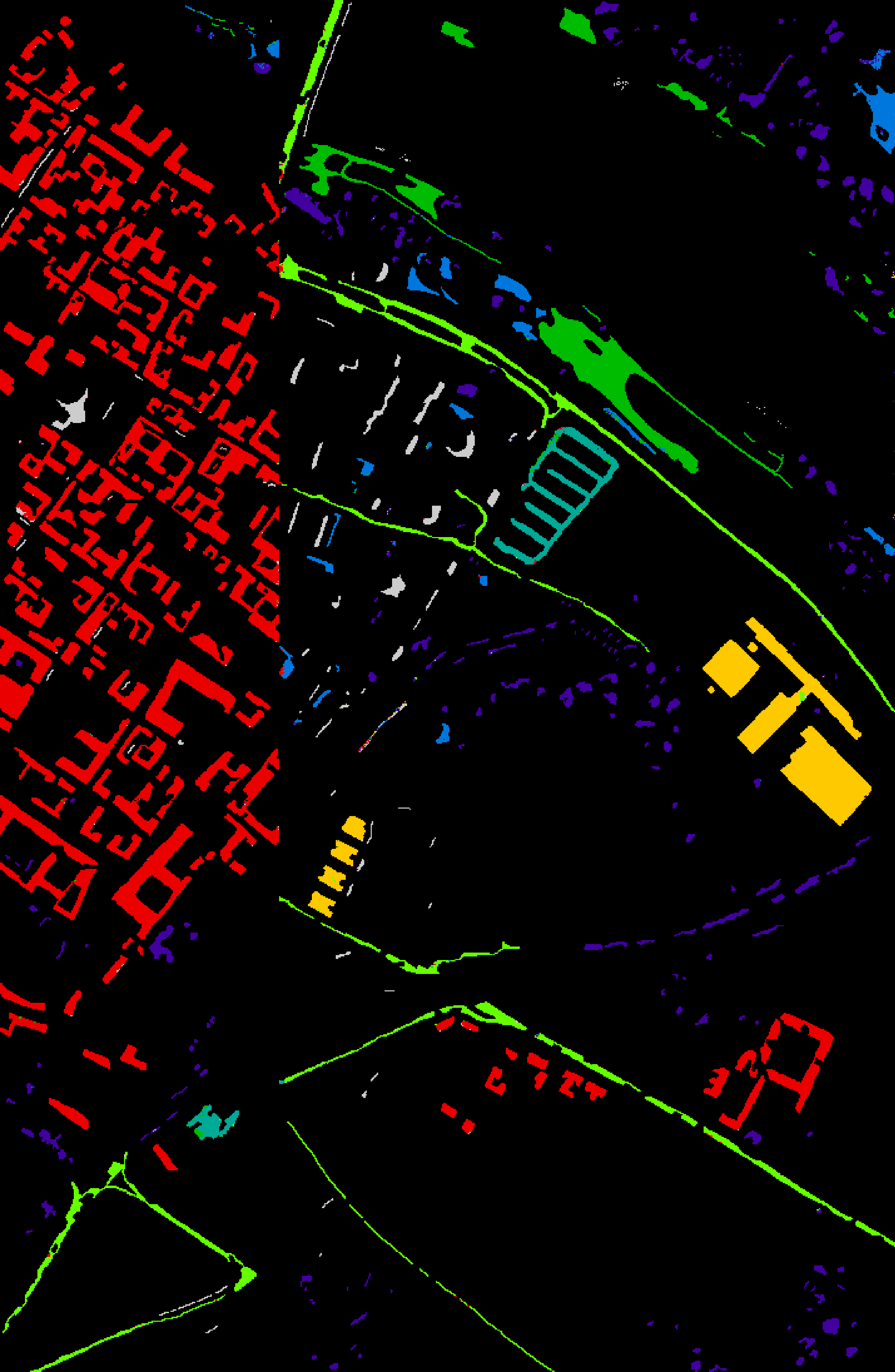}
            \caption{GMamba}
        \end{subfigure}
    \caption{The predicted ground truth maps for the PC dataset are presented for various state-of-the-art methods along with GraphMamba.}
    \label{fig:PC_results}
\end{figure*}
\begin{figure*}[!htb]
    \centering
        \begin{subfigure}{0.48\textwidth}
            \includegraphics[width=0.99\textwidth]{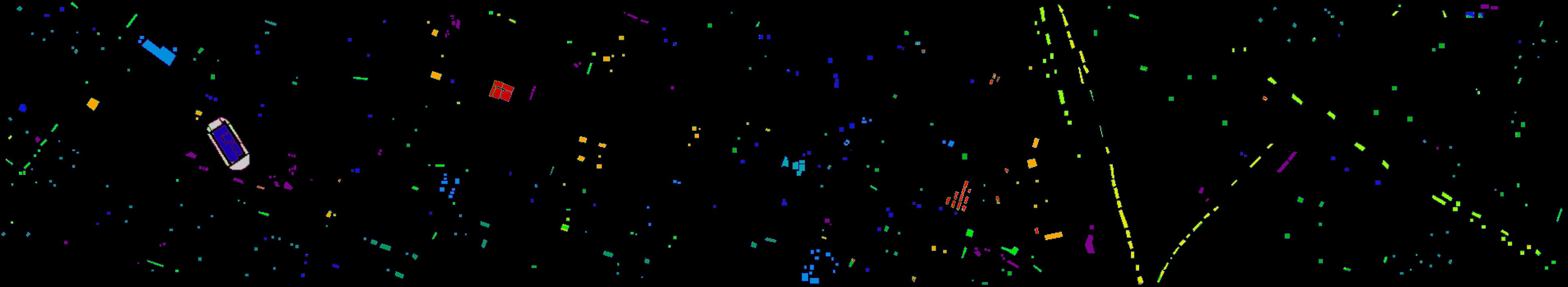}
            \caption{2DCNN}
        \end{subfigure}
        \begin{subfigure}{0.48\textwidth}
            \centering
            \includegraphics[width=0.99\textwidth]{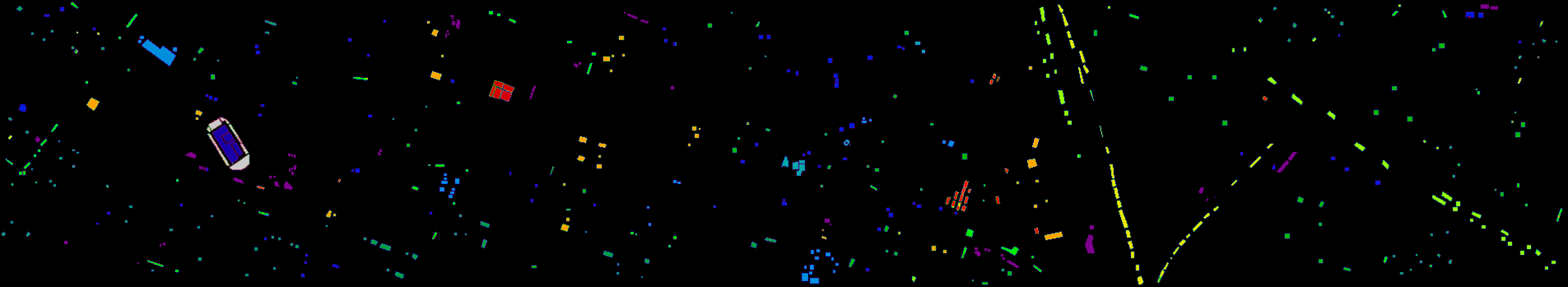}
            \caption{3DCNN}
        \end{subfigure}
        \begin{subfigure}{0.48\textwidth}
            \centering
            \includegraphics[width=0.99\textwidth]{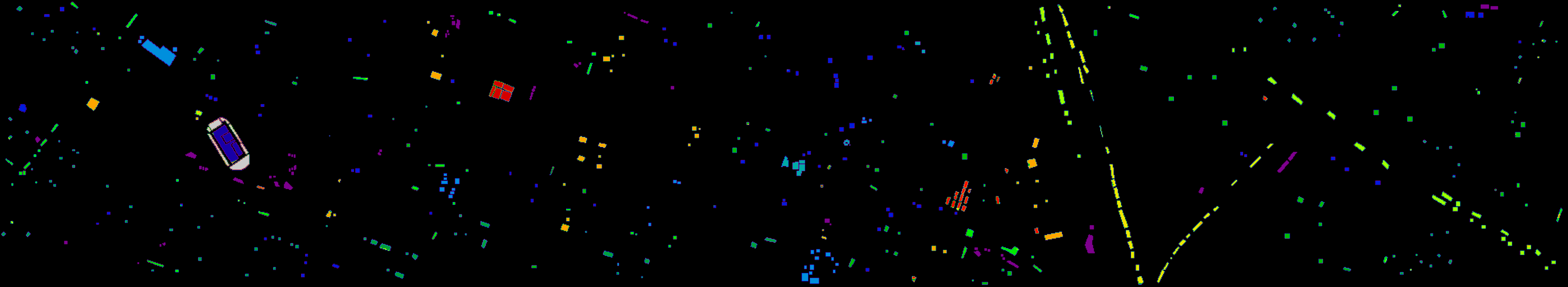}
            \caption{HybCNN}
        \end{subfigure}
        \begin{subfigure}{0.48\textwidth}
            \includegraphics[width=0.99\textwidth]{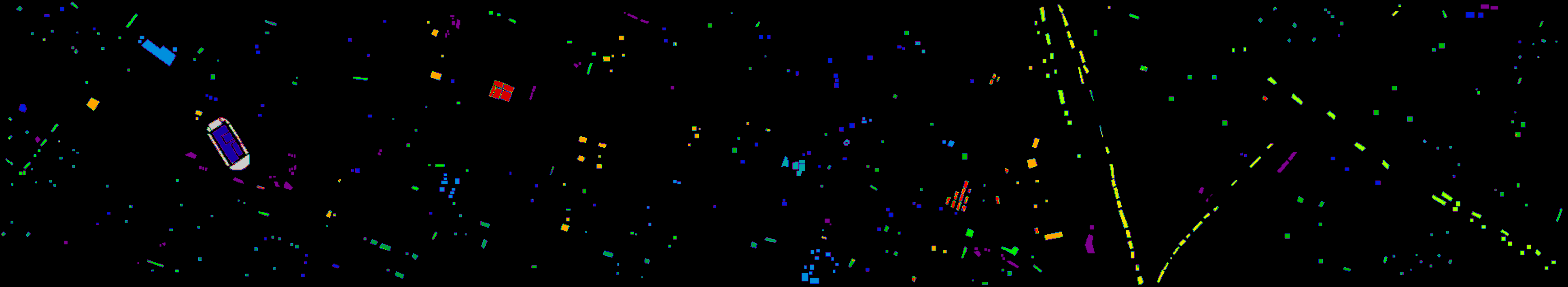}
            \caption{2DIN}
        \end{subfigure}
        \begin{subfigure}{0.48\textwidth}
            \centering
            \includegraphics[width=0.99\textwidth]{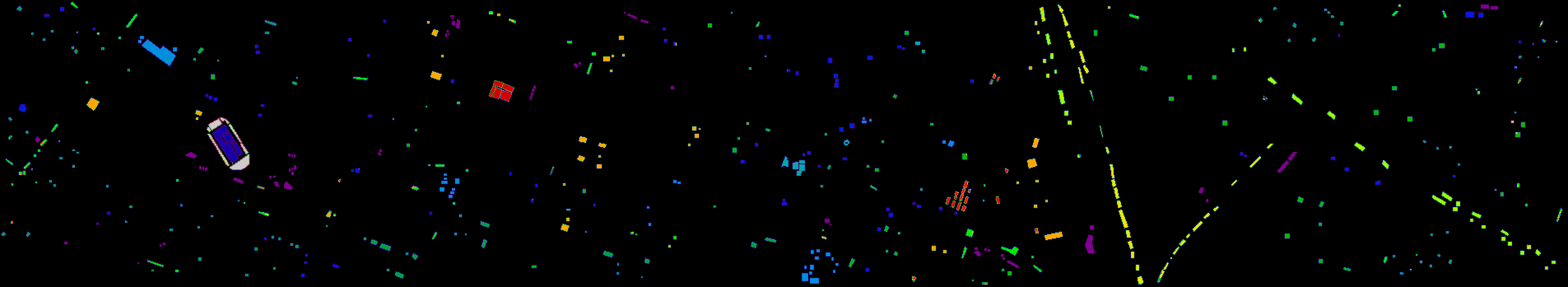}
            \caption{3DIN}
        \end{subfigure}
        \begin{subfigure}{0.48\textwidth}
            \centering
            \includegraphics[width=0.99\textwidth]{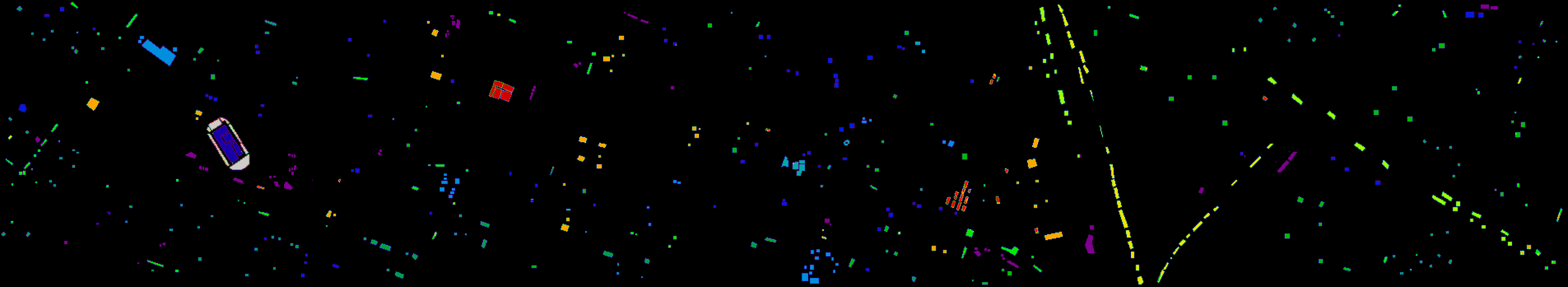}
            \caption{HybIN}
        \end{subfigure}
        \begin{subfigure}{0.48\textwidth}
            \centering
            \includegraphics[width=0.99\textwidth]{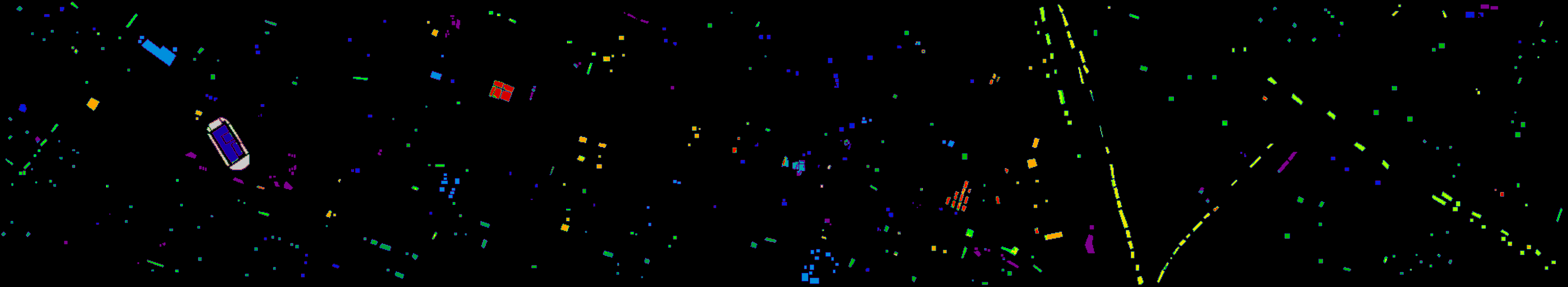}
            \caption{Attention Graph CNN}
        \end{subfigure}
        \begin{subfigure}{0.48\textwidth}
            \centering
            \includegraphics[width=0.99\textwidth]{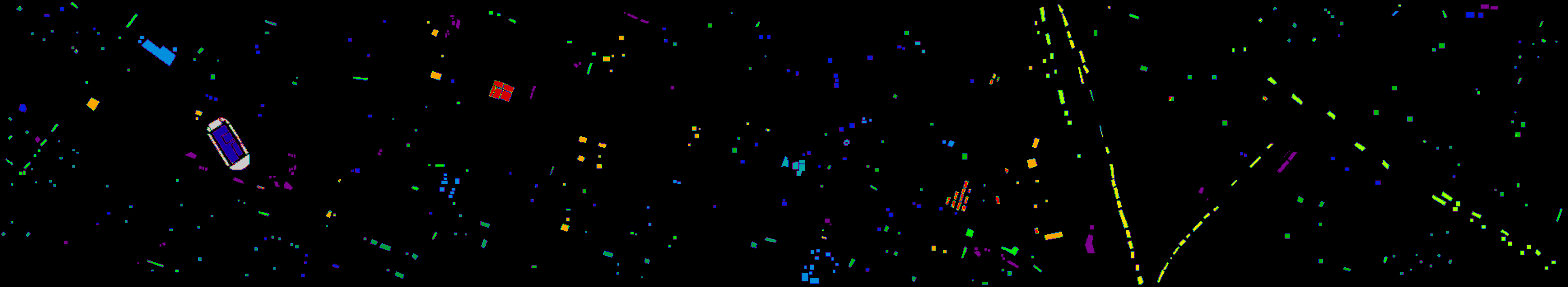}
            \caption{Hybrid ViT}
        \end{subfigure}
        \begin{subfigure}{0.48\textwidth}
            \centering
            \includegraphics[width=0.99\textwidth]{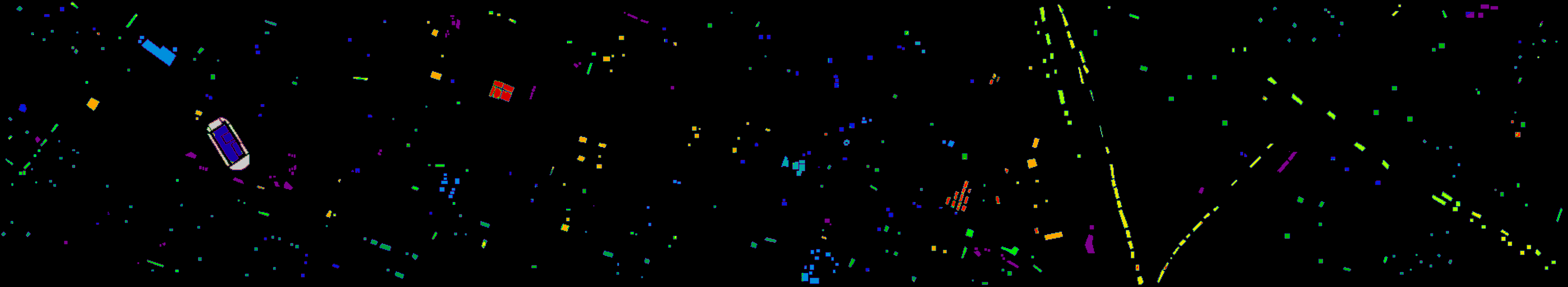}
            \caption{Hir\_ViT}
        \end{subfigure}
        \begin{subfigure}{0.48\textwidth}
            \centering
            \includegraphics[width=0.99\textwidth]{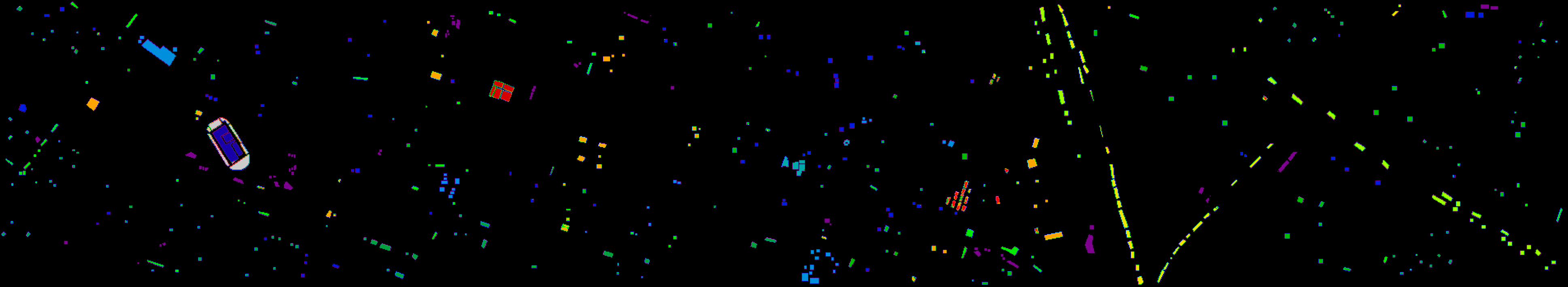}
            \caption{SST}
        \end{subfigure}
        \begin{subfigure}{0.48\textwidth}
            \centering
            \includegraphics[width=0.99\textwidth]{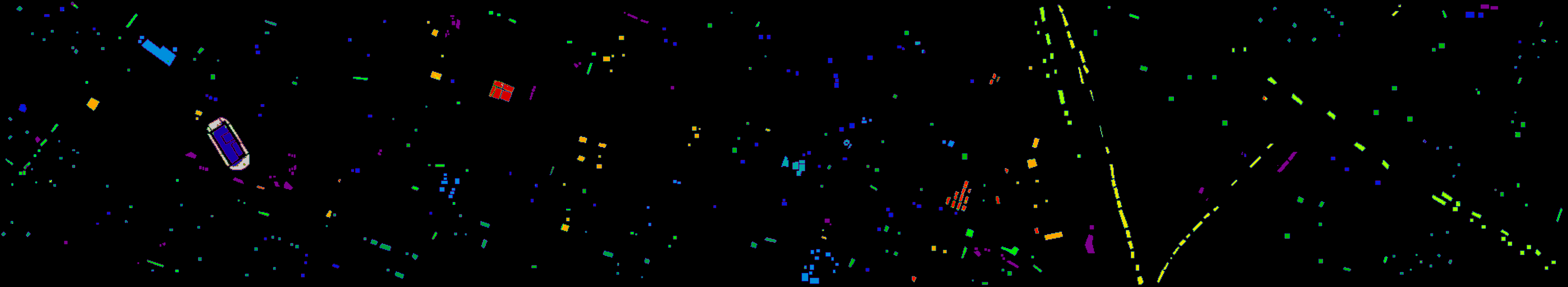}
            \caption{GraphMamba}
        \end{subfigure}
    \caption{The predicted ground truth maps for the UH dataset are presented for various state-of-the-art methods along with the GraphMamba.}
    \label{fig:UH_results}
\end{figure*}

\section{Comparative Results and Discussion}
\label{Results}

The comparative analysis, presented in Table \ref{comperative} and Figures \ref{fig:HC_results}, \ref{fig:PU_results}, \ref{fig:SA_results}, \ref{fig:UH_results}, and \ref{fig:PC_results}, evaluates various models for HSI classification across multiple datasets. The analysis highlights key performance metrics such as overall accuracy (OA), average accuracy (AA), Kappa coefficient ($\kappa$), and the number of trainable parameters. The models compared include CNN-based models (2D CNN, 3D CNN, and HybCNN) and transformer-based models (Hybrid-ViT, Hir-ViT, and SSViT).

As shown in Table \ref{comperative}, GraphMamba consistently has the fewest parameters across all datasets compared to other models, significantly reducing complexity. For example, on the PU dataset, GraphMamba has only 143,498 parameters, far fewer than 2DCNN (1.1M) or Hybrid-ViT (835K), demonstrating its lightweight architecture. Despite the lower parameter count, GraphMamba achieves the highest accuracy on most datasets, including PU, SA, and UH, where it surpasses other models in OA, AA, and Kappa scores.

On the PU dataset, GraphMamba achieves an impressive 99.51\% OA and 99.36\% Kappa, outperforming all other models, including Hybrid CNNs, CNN-based models, and Transformer-based models. A similar performance trend is observed across other datasets such as SA and UH, where GraphMamba consistently surpasses existing models in accuracy. For instance, on the SA dataset, GraphMamba achieves 99.12\% OA, outperforming SSViT (97.69\%) and Hybrid-ViT (97.56\%). Additionally, GraphMamba demonstrates superior model reliability with a consistent Kappa above 99\% across several datasets, highlighting its effectiveness in reducing classification errors and enhancing spatial-spectral feature learning.

On the PC dataset, although GraphMamba does not secure the top spot for OA, it still performs competitively, achieving a Kappa of 98.92\% and an OA of 99.23\%. This demonstrates its ability to generalize well across various datasets and maintain strong performance even in challenging cases. On the HC dataset, GraphMamba achieves an OA of 97.36\% and an AA of 94.24\%, outperforming models such as 2DCNN, 3DCNN, and Transformer variants, further reinforcing its strength in datasets with complex spectral-spatial relationships.

GraphMamba introduces an advanced architecture that leverages graph-based token interactions and hybrid state-space modeling to enhance classification performance. Its lightweight nature, evidenced by its minimal parameter count, makes it highly efficient for real-world applications. Furthermore, its robust performance across multiple datasets—surpassing conventional CNN and Transformer models in accuracy—demonstrates its effectiveness in capturing spectral-spatial relationships in HSIs. The strength of the GraphMamba model lies in its ability to maintain high classification accuracy while operating with fewer parameters and potentially less favorable computational environments. This demonstrates the model’s adaptability and its potential for deployment in resource-constrained scenarios. In summary, GraphMamba is a highly promising model for HSI classification, offering an optimal balance between efficiency and accuracy. Its performance across multiple datasets highlights the effectiveness of its graph-based approach, enabling it to outperform state-of-the-art CNN and Transformer models while utilizing significantly fewer computational resources.

In terms of visual outcomes, as shown in Figures \ref{fig:HC_results}, \ref{fig:PU_results}, \ref{fig:SA_results}, \ref{fig:PC_results}, and \ref{fig:UH_results}, GraphMamba consistently produces superior predicted ground truth maps across all datasets, outperforming state-of-the-art models in various aspects. For instance, in the PC and PU datasets, GraphMamba demonstrates remarkable spatial coherence and class separability—key factors for precise HSI classification—while utilizing significantly fewer parameters. This level of visual and quantitative consistency is also evident in the HC, SA, and UH datasets, where GraphMamba generates maps that are comparable to, and often surpass, those produced by more parameter-heavy models like 3DCNN, 3DIN, and Hir-ViT. A key advantage of GraphMamba is its ability to deliver these high-quality results while minimizing model size and computational complexity, making it highly suitable for real-world applications in resource-constrained environments, such as UAV-based HSI systems. The model's superior balance of accuracy, efficiency, and simplicity establishes it as a leading solution for HSI classification tasks, consistently outperforming its counterparts across various contexts.

\section{Computational Complexity}
\label{Complexity}

The computational complexity of GraphMamba is analyzed by examining key components such as token generation, graph construction, cross-attention, feature fusion, and the state-space model. The token generation layer involves two convolutional operations: a spatial convolution with a kernel size of $3 \times 3$ and spectral convolutions with a kernel size of $1 \times 1$. The complexity of the convolutional layer is given by $O(HWC \cdot K^2 \cdot F)$, where $H$, $W$, and $C$ represent the height, width, and channels of the input, $K$ is the kernel size, and $F$ is the number of features. For the spatial convolution, this results in $O(HWC \cdot F \cdot 9)$ operations per layer, while the spectral convolution, with a smaller kernel size, results in $O(HWC \cdot F)$ operations. Tokenization, which involves dense operations on flattened spatial and spectral tokens, introduces an additional $O(HW \cdot F^2)$ complexity for each tokenization step.

The token graph construction involves computing an adjacency matrix using inner products, which has a complexity of $O(T^2 \cdot F)$, where $T$ is the number of tokens and $F$ is the output dimensionality. Selecting prioritized tokens and constructing the graph involves sorting operations with a complexity of $O(T \log T)$. The graph-based feature processing is handled by multiplying the adjacency matrix with the token matrix, resulting in $O(T^2 \cdot F)$ operations. The cross-attention mechanism requires matrix multiplications between the query, key, and value matrices, with a complexity of $O(T_Q \cdot T_K \cdot F)$, where $T_K$ is the number of key and value tokens, $T_Q$ is the number of query tokens, and $F$ is the feature dimensionality. The attention weights are computed and used to aggregate values, which also have a complexity of $O(T_Q \cdot T_K \cdot F)$. The fusion layer combines the outputs from the graph and attention mechanisms through dense operations, concatenations, and reshaping, leading to a complexity of $O(T \cdot F^2)$, where $T$ is the token count and $F$ is the feature dimension. Finally, the state-space model, implemented via a GRU, has a complexity of $O(T \cdot F \cdot S)$, where $T$ is the number of tokens, $F$ is the feature dimension, and $S$ is the state size.

\section{Conclusions and Future Research Directions}
\label{Con}
HSI (Hyperspectral Image) classification plays a pivotal role in various high-impact domains such as environmental monitoring and urban planning. Although traditional methods and deep learning techniques have made significant strides, they often struggle to capture the complex spatial-spectral relationships inherent in hyperspectral data while managing the associated computational complexity. In this research, we introduce GraphMamba, a groundbreaking model designed to overcome these challenges. By combining spectral-spatial token generation, graph-based token prioritization, cross-attention mechanisms, and state-space modeling, GraphMamba effectively captures intricate spatial-spectral dependencies and dynamically prioritizes essential features. This novel integration enables superior classification accuracy and scalability, especially in large-scale and complex datasets. Extensive experimental results demonstrate that GraphMamba not only outperforms current state-of-the-art models in both accuracy and efficiency but also sets a new benchmark for computational efficiency in hyperspectral image classification. Looking ahead, further optimization of the model's computational efficiency, particularly in graph-based operations and attention mechanisms, is essential for enabling real-time applications and handling even larger datasets. Additionally, enhancing the model's robustness in noisy, incomplete, or less-controlled data environments will open new avenues for practical deployment. Future research may explore innovative approaches, such as integrating meta-learning and self-supervised techniques, to push the boundaries of feature extraction and classification, ensuring that GraphMamba remains at the forefront of HSI classification advancements.

\bibliographystyle{IEEEtran}
\bibliography{IEEEabrv,Sam}
\end{document}